\documentclass{article}
\usepackage{geometry}               
\usepackage{microtype}
\usepackage{graphicx}
\usepackage{subcaption}
\usepackage{booktabs} 
\usepackage{tikz}
\usetikzlibrary{plotmarks,calc}

\usepackage{natbib}
\usepackage{xurl}
\usepackage{hyperref}

\usepackage{amsmath}
\DeclareMathOperator*{\argmax}{arg\,max}
\DeclareMathOperator*{\argmin}{arg\,min}
\usepackage{amssymb}
\usepackage{mathtools}
\usepackage{amsthm}

\usepackage{thmtools}
\usepackage{thm-restate}
\usepackage{comment}

\usepackage[capitalize,noabbrev]{cleveref}

\theoremstyle{plain}
\newtheorem{theorem}{Theorem}[section]
\newtheorem{proposition}[theorem]{Proposition}
\newtheorem{lemma}[theorem]{Lemma}
\newtheorem{corollary}[theorem]{Corollary}
\theoremstyle{definition}
\newtheorem{definition}[theorem]{Definition}

\theoremstyle{remark}

\begin{document}

\title{Sharpness-Aware Minimization  Can Hallucinate Minimizers}
\author{
  Chanwoong Park\thanks{Department of Electrical and Computer Engineering, Seoul National University.} \and
  Uijeong Jang\thanks{Department of Mathematics, University of California, Los Angeles.} \and
  Ernest K.\ Ryu\footnotemark[2] \and
  Insoon Yang\footnotemark[1]
}
\date{}
\maketitle










\begin{abstract}
Sharpness-Aware Minimization (SAM) is widely used to seek flatter minima---often linked to better generalization.
In its standard implementation, SAM updates the current iterate using the loss gradient evaluated at a point perturbed by distance $\rho$ along the normalized gradient direction. We show that, for some choices of $\rho$, SAM can stall at points where this shifted (perturbed-point) gradient vanishes despite a nonzero original gradient, and therefore, they are not stationary points of the original loss. 
We call these points \emph{hallucinated minimizers}, prove their existence under simple nonconvex landscape conditions (e.g., the presence of a local minimizer and a local maximizer), and establish sufficient conditions for local convergence of the SAM iterates to them. We corroborate this failure mode in neural network training and observe that it aligns with SAM's performance degradation often seen at large $\rho$. Finally, as a practical safeguard, we find that a short initial SGD warm-start before enabling SAM mitigates this failure mode and reduces sensitivity to the choice of $\rho$.
\end{abstract}

\begin{figure*}[t]
    \centering
    \begin{subfigure}[t]{0.32\textwidth}
        \centering
        \includegraphics[width=\linewidth]{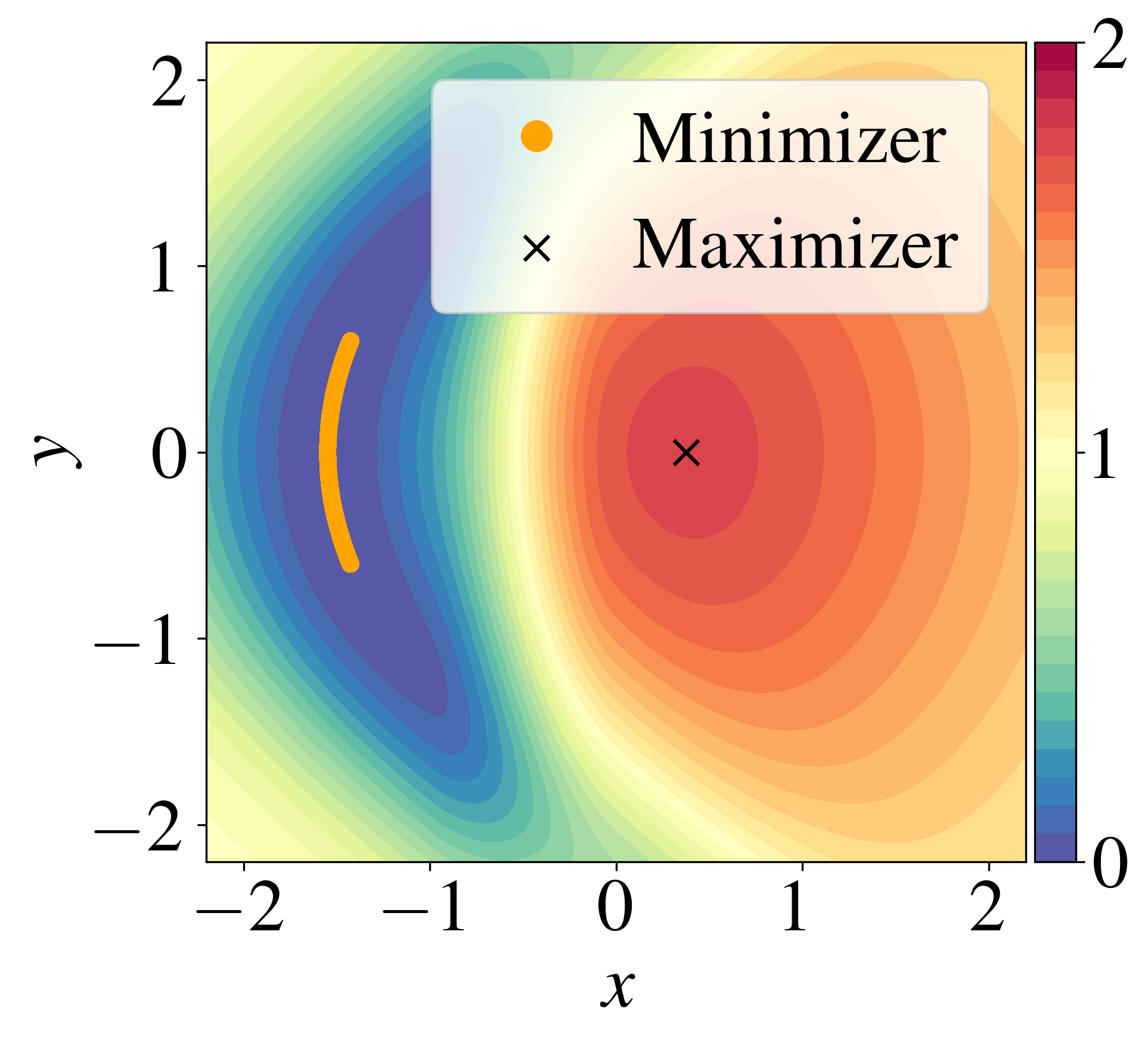}
        \caption{Landscape of \( f \)}
    \end{subfigure}
    \hfill
    \begin{subfigure}[t]{0.322\textwidth}
        \centering
        \includegraphics[width=\linewidth]{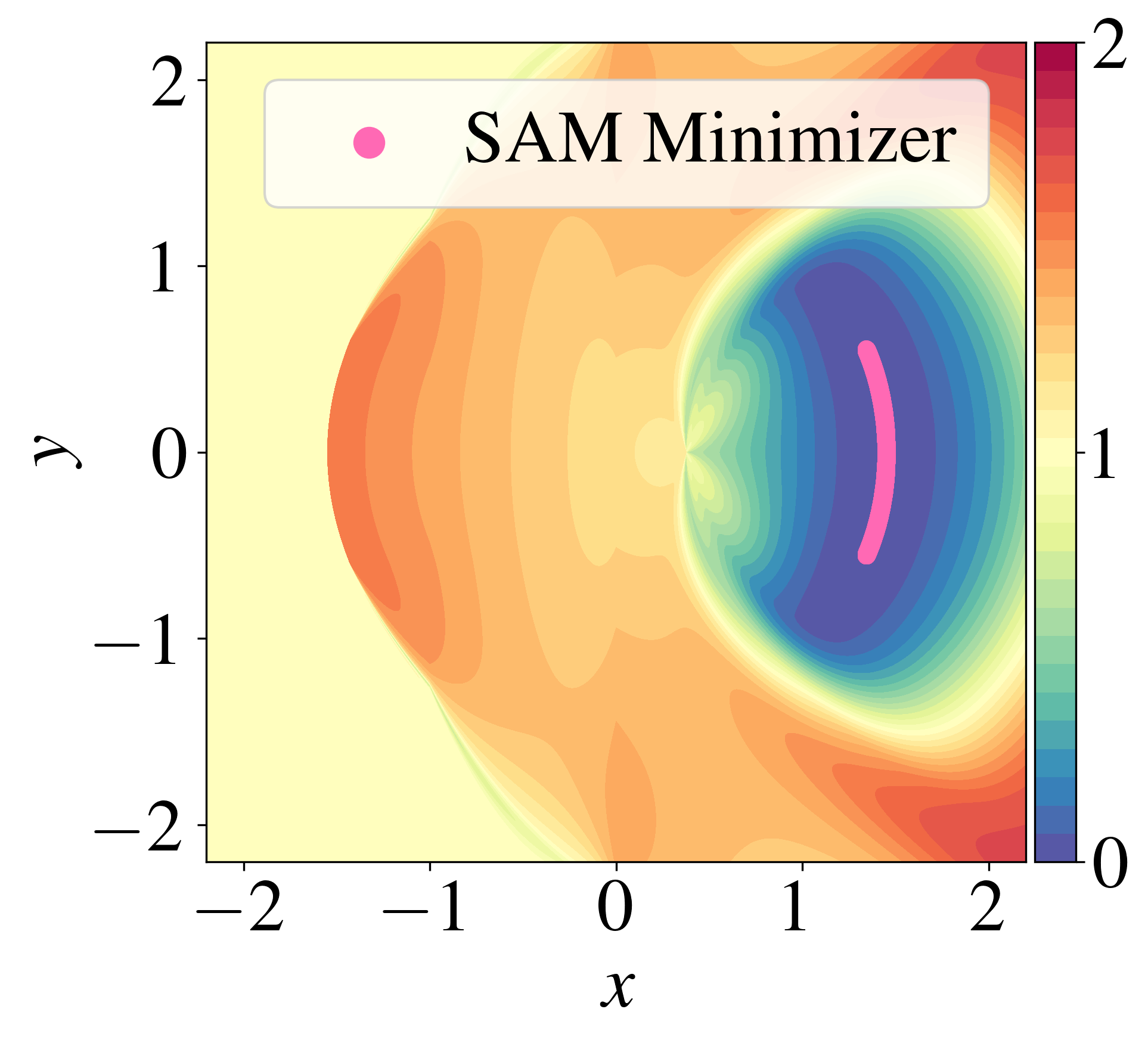}
        \caption{Landscape of \( f^{\mathrm{SAM}}\)}
    \end{subfigure}
    \hfill
    \begin{subfigure}[t]{0.299\textwidth}
        \centering
        \includegraphics[width=\linewidth]{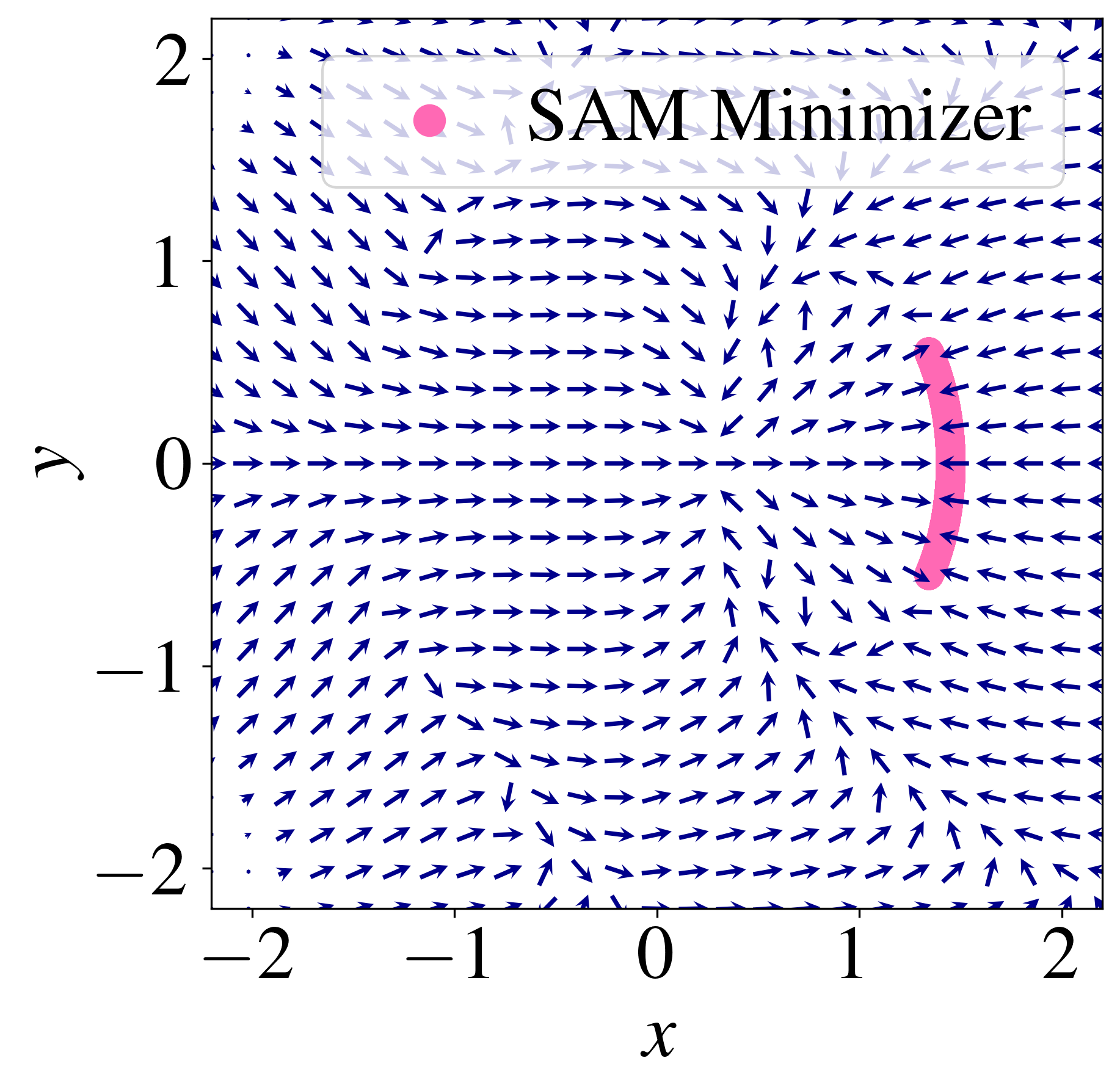}
        \caption{SAM gradient field}
    \end{subfigure}

    \caption{Illustrative example of hallucinated minimizers. See Appendix~\ref{app:exp1} for details. (a) Smooth function $f$ with a minimizer set and an isolated maximizer. 
    (b) $f^{\mathrm{SAM}}=f(x+\rho \, u(x))$; its minimizers  (labeled as ``SAM minimizers'' in the plot) do not correspond to minimizers or stationary points of $f$ and are therefore hallucinated.
    (c) Vector field of the SAM gradient $\nabla f(x^+)$; the hallucinated minimizers are attractors of the SAM iteration.}
    \label{fig:toy}
\end{figure*}

\section{Introduction}

Sharpness-Aware Minimization (SAM) is widely used to bias training toward flatter regions of the loss landscape~\citep{foret2021sharpnessaware},
often correlated with improved generalization~\citep{neyshabur2017exploring, Jiang2020Fantastic}.
For a differentiable objective $f:\mathbb{R}^d\to\mathbb{R}$, SAM is commonly motivated by the min--max formulation
\[
\min_x \max_{\|\epsilon\|\le \rho} f(x+\epsilon),
\]
and the inner maximization is typically approximated by a first-order perturbation along the normalized gradient direction.
This leads to the surrogate (``SAM'') loss
\[
f^{\mathrm{SAM}}(x) :=f\!\left(x+\rho\,u(x)\right),
\qquad
u(x)=\frac{\nabla f(x)}{\|\nabla f(x)\|},
\]
where $\rho>0$ is the perturbation radius controlling the strength of the sharpness regularization.

\paragraph{A practical mismatch in SAM updates.}
A central subtlety---often glossed over in high-level descriptions of SAM---is that the \emph{standard} implementation does not follow
the \emph{surrogate gradient} $\nabla f^{\mathrm{SAM}}(x)$.
Instead, it updates using the \emph{shifted gradient} evaluated at the perturbed point:
\[
x_{k+1}=x_k-\eta_k\,\nabla f(x_k^{+}), \qquad
x^{+}:=x+\rho\,u(x).
\]
Consequently, the practical SAM dynamics can become stationary whenever $\nabla f(x^{+})=0$,
even if $\nabla f(x)\neq 0$.
This mismatch becomes especially relevant in the \emph{large-$\rho$} regime: 
in practice, $\rho$ is tuned by sweeps or grid search,
and prior work reports performance degradation and instability at larger radii on standard benchmarks~\citep{andriushchenko2022towards}.
On the theory side, most convergence guarantees assume $\rho$ is below a threshold under various regularity conditions,
leaving the large-$\rho$ regime comparatively less understood~\citep{si2024practical, khanh2024fundamental, oikonomou2025sharpnessaware}.

\paragraph{Hallucinated minimizers.}
We show that the shifted-gradient dynamics can admit a qualitatively distinct failure mode:
SAM can stall at points where  $\nabla f(x^{+})= 0$
despite $\nabla f(x)\neq 0$.
We call such points \emph{hallucinated minimizers} if, in addition, they are local minimizers of the surrogate loss $f^{\mathrm{SAM}}$.
Since $\nabla f(x)\neq 0$, hallucinated minimizers  are not stationary points   of the original objective $f$ (Figure~\ref{fig:toy}).
This failure mode is inherently nonconvex (it cannot occur for convex $f$), and becomes
particularly relevant in the large-$\rho$ regime,  where the empirical performance of SAM is often observed to degrade~\citep{andriushchenko2022towards}.

\paragraph{Our results.}
We provide a geometric and dynamical characterization of hallucinated minimizers.
Specifically, we prove that 
hallucinated minimizers exist for a nontrivial interval of radii, including values in a large-$\rho$ regime
(Theorem~\ref{main:thm1})
under simple nonconvex landscape conditions
(e.g., the coexistence of a local minimizer and a local maximizer).
We extend this existence result from isolated maximizers to bounded \emph{local maximizer sets}
under a real-analyticity assumption (Theorem~\ref{thm:exist}).
We further show that hallucinated minimizers are not merely an algebraic possibility:
an isolated connected component can be \emph{locally attracting} for the discrete-time SAM dynamics
(Theorem~\ref{thm:attractor}).
Finally, we show that hallucinated minimizers can be \emph{non-isolated} and inherit the dimension of minimizer manifolds
via local invertibility of the perturbation map (Theorem~\ref{thm:manifold}).

Empirically, we provide evidence consistent with this failure mode on neural network objectives using diagnostics that compare shifted vs.\ unshifted gradients and losses:
in smooth full-batch settings we directly verify convergence to points with
nearly vanishing shifted gradient but nonzero original gradient,
and we observe similar mismatch signatures in larger networks beyond the smooth/analytic setting.
Finally, as a practical safeguard, we report that a short initial SGD warm-start before enabling SAM mitigates this failure mode
and reduces sensitivity to the choice of $\rho$.
Our formal results concern deterministic (full-batch) dynamics; stochastic experiments are included only as empirical observations.

\paragraph{Contributions.} 
\begin{itemize}\setlength{\itemsep}{0pt}
\item We identify and formalize \emph{hallucinated minimizers}: points where the shifted gradient vanishes, $\nabla f(x^+)=0$, even though the original gradient is nonzero, $\nabla f(x)\neq 0$, and $x$ is a local minimizer of $f^{\mathrm{SAM}}(x)=f(x^+)$.

\item We prove existence of hallucinated minimizers for a nontrivial interval of radii (including a large-$\rho$ regime) under mild nonconvex landscape conditions, and extend the result to bounded local maximizer sets under real-analyticity.

\item We provide sufficient conditions under which an isolated hallucinated-minimizer component is locally attracting for the discrete-time SAM dynamics.

\item We show that hallucinated minimizers can form smooth manifolds inheriting the dimension of minimizer manifolds via local invertibility of the perturbation map.

\item We provide empirical diagnostics and observations in neural network training that are consistent with the proposed mechanism.

\end{itemize}

\subsection{Related work}

\paragraph{SAM and practical variants.}
SAM was introduced as a method to regularize sharpness by adversarially perturbing parameters during training~\citep{foret2021sharpnessaware}.
A large body of follow-up work proposes variants that modify the perturbation geometry or the update rule, e.g.,
scale-invariant perturbations (ASAM)~\citep{kwon2021asam}, Fisher-geometry perturbations~\citep{kim2022fishersam},
incorporation of momentum into the perturbation~\citep{li2024enhancing}
or modified updates such as orthogonal decomposition (GSAM)~\citep{zhuang2022gsam}.
Other works reduce SAM's two-backward-pass overhead via approximation, partial perturbations, or scheduling (e.g.,~\citep{liu2022towards, jiang2023an, du2022efficient, du2022sharpness, mueller2023normalization}).
In this paper, we focus on the \emph{standard}  SAM update---the shifted-gradient step used in common implementations---and analyze its behavior in the large-$\rho$ regime.

\paragraph{Theoretical analyses of SAM.}
Existing theory has examined SAM from multiple perspectives, including sharpness notions and generalization interpretations~\citep{wen2022how, mollenhoff2023bayes, chen2023does, wei2023sharpness}.
Another line studies SAM's dynamics and stability, e.g., via continuous-time and SDE approximations~\citep{compagnoni2023sde},
quadratic dynamics and surrogate interpretations~\citep{bartlett2023dynamics, si2024practical},
and stability phenomena related to normalization and edge-of-stability effects~\citep{dai2023crucial, long2024sharpness}.
Several works provide convergence guarantees for SAM and related methods under various assumptions,
typically requiring a small-radius condition or a regularity condition that effectively bounds the perturbation effect
(e.g., convex/strongly convex regimes, inexact gradient descent frameworks, or PL-type conditions)~\citep{andriushchenko2022towards, si2024practical, khanh2024fundamental, oikonomou2025sharpnessaware}.
In contrast, our results characterize a distinct phenomenon that can arise at \emph{large} $\rho$:
the standard SAM dynamics can stall at points that are not stationary (and hence not locally optimal) for $f$.

\paragraph{Differences from prior work: hallucinated minimizers vs.\ saddle trapping.}
Prior work has reported that SAM can become trapped near saddle points~\citep{kaddour2022flat},
and theoretical explanations have been developed
via \emph{continuous-time} SAM models~\citep{kim2023stability, compagnoni2023sde}.
Hallucinated minimizers represent a \emph{qualitatively different} failure mode:
saddle trapping corresponds to $\nabla f(x)=0$ (stationarity of the original loss),
whereas hallucinated minimizers satisfy $\nabla f(x)\neq 0$ but $\nabla f(x^{+})=0$,
so they are stationary \emph{only} for the SAM's shifted-gradient update.
Moreover, our analysis provides $(i)$ geometric existence mechanisms tied to nonconvex landscapes and large $\rho$,
$(ii)$ sufficient conditions for local attraction under the \emph{discrete-time} SAM dynamics,
and $(iii)$ manifold structure results via local invertibility of the perturbation map.

\paragraph{Switching schedules.}
Switching schedules for SAM have been explored in prior work, often motivated by computational overhead or late-phase effects~\citep{andriushchenko2022towards, zhou2025sharpnessaware}.
Complementary to these perspectives, we report that a short \emph{early} SGD warm-start before enabling SAM can mitigate
the hallucinated-minimizer failure mode identified in this paper.
Our goal is not to propose a new training recipe, but to provide a simple and broadly compatible safeguard aligned with our mechanism.

\subsection{Notation and terminology}
\label{sec:notation}

We use $\|\cdot\|$ for the Euclidean norm (and its induced operator norm for matrices), write $\mathrm{Sym}(A):=(A+A^\top)/2$, and let $\lambda_{\min}(A)$ denote the smallest eigenvalue of a symmetric matrix $A$. A function $f\colon \mathbb{R}^d\to\mathbb{R}$ is called \emph{real-analytic} if its Taylor series at any point $x_0$ converges to $f$ on a neighborhood of $x_0$. For $\alpha\in\mathbb{R}$, the $\alpha$-superlevel set of $f$ is defined as $\{x:\,f(x)\ge \alpha\}$. For a set $C\subset\mathbb{R}^d$, we write $\partial C$ for its boundary, and we say that $C$ is \emph{connected} if it cannot be expressed as the union of two disjoint, nonempty open sets. The distance from a point $x\in\mathbb{R}^d$ to a nonempty set $C\subseteq\mathbb{R}^d$ is $d(x,C):=\inf_{y\in C}\|x-y\|$; if $C$ is closed, the infimum is attained, and hence $d(x,C)=\min_{y\in C}\|x-y\|$. For $\delta>0$, $B_\delta(x)$ is the open ball centered at $x$ with radius $\delta$ and $\mathcal{N}_\delta(C)=\{x:\,d(x,C)\le \delta\}$ is
the (closed) $\delta$-neighborhood of $C$.

\section{Existence of hallucinated minimizers}\label{sec:2}

In this section, under simple nonconvex landscape assumptions,
we establish the existence of \emph{hallucinated minimizers}:
points where the (full-batch) standard SAM update can become stationary even though the original gradient is nonzero.

\paragraph{Why a mismatch is possible.}
Recall that SAM is commonly motivated by the surrogate objective
\[
f^{\mathrm{SAM}}(x) := f\!\left(x+\rho\,u(x)\right),
\quad
u(x):=\frac{\nabla f(x)}{\|\nabla f(x)\|}
\]
for $\nabla f(x)\neq 0$.
However, the standard SAM implementation does \emph{not} follow the \emph{surrogate gradient} $\nabla f^{\mathrm{SAM}}(x)$.
Instead, it uses the shifted gradient $\nabla f(x+\rho u(x))$, which we refer to as the \emph{SAM gradient}.
For brevity, we write
\[
x^{+}:=x+\rho\,u(x).
\]
Then the (full-batch) SAM iteration takes the form
\begin{equation}\label{eq:sam-update-sec3}
x_{k+1}=x_k-\eta_k\,\nabla f(x_k^{+}),
\end{equation}
so the dynamics can stall whenever $\nabla f(x^{+})=0$, regardless of whether $\nabla f(x)=0$.

\begin{definition}[Hallucinated minimizers]
Fix $\rho>0$.
A point $x\in\mathbb{R}^d$ is said to be a \emph{hallucinated minimizer} at radius $\rho$ if
\[
\nabla f(x^+)=0
\quad\text{and}\quad
\nabla f(x)\neq 0,
\]
and $x$ is a local minimizer of $f^{\mathrm{SAM}}(x)=f(x^+)$.
\end{definition}

\paragraph{A useful relation between $\nabla f^{\mathrm{SAM}}(x)$ and the SAM gradient $\nabla f(x^{+})$.}
Assume in addition that $f$ is twice continuously differentiable on a neighborhood of $x$ and that $\nabla f(x)\neq 0$.
Then $u$ is differentiable at $x$ and the chain rule gives
\[
\nabla f^{\mathrm{SAM}}(x)=\bigl(I+\rho\,\nabla u(x)\bigr)^\top \nabla f(x^{+}).
\]
If $I+\rho\nabla u(x)$ is nonsingular (equivalently, the perturbation map $x\mapsto x^{+}$ is locally invertible at $x$), then $\nabla f^{\mathrm{SAM}}(x)=0$ implies $\nabla f(x^{+})=0$.
In particular, at such \emph{nondegenerate} points, any local minimizer of $f^{\mathrm{SAM}}$ with $\nabla f(x)\neq 0$
automatically satisfies $\nabla f(x^{+})=0$.
For brevity, when nondegeneracy is understood, we will therefore sometimes describe hallucinated minimizers simply as
\emph{local minimizers of $f^{\mathrm{SAM}}$ with $\nabla f(x)\neq 0$}.
This nondegeneracy condition is used only to justify this shorthand; our existence results do not rely on it.

\paragraph{Convexity rules out hallucinated minimizers.}
Hallucinated minimizers cannot arise when $f$ is convex.
Indeed, Proposition~\ref{thm:hmcv} in  Appendix~\ref{appendix2} shows that if $f$ is convex and $\nabla f(x)\neq 0$, then $\nabla f(x^{+})\neq 0$.

\begin{figure}[htbp]
  \centering
\begin{tikzpicture}[x=1cm,y=1cm]

  \coordinate (xs)  at (1.5,0.34);    
  \coordinate (xh)  at (6.32,3.03);   
  \coordinate (xst) at (4.9,1.5);     
  \coordinate (p1)  at (3.7,3.0);
  \coordinate (p2)  at (5.0,2.9);
  \coordinate (p4)  at (6.2,3.2);
  \coordinate (p6)  at (6.3,1.5);
  \coordinate (p7)  at (5.1,0.8);
  \coordinate (p8)  at (3.8,1.4);

 \path 
   coordinate (s1) at ($(p1)!0!(xst)$)
    coordinate (s2)  at ($(p2)!0!(xst)$)
    coordinate (s4) at ($(p4)!0!(xst)$)
     coordinate (s6)  at ($(p6)!0!(xst)$)
     coordinate (s7) at ($(p7)!0!(xst)$)
    coordinate (s8) at ($(p8)!0!(xst)$) ;

 \path 
   coordinate (q1) at ($(p1)!0.3!(xst)$)
    coordinate (q2) at ($(p2)!0.3!(xst)$)
    coordinate (q4) at($(p4)!0.34!(xst)$)
     coordinate (q6) at($(p6)!0.4!(xst)$)
     coordinate (q7) at ($(p7)!0.4!(xst)$)
    coordinate (q8) at ($(p8)!0.45!(xst)$);

     \path 
   coordinate (r1) at ($(p1)!-0.2!(xst)$)
    coordinate (r2) at  ($(p2)!-0.2!(xst)$)
    coordinate (r4) at   ($(p4)!-0.21!(xst)$)
     coordinate (r6) at  ($(p6)!-0.18!(xst)$)
     coordinate (r7) at  ($(p7)!-0.22!(xst)$)
    coordinate (r8) at  ($(p8)!-0.3!(xst)$);


  \draw[black,very thick] plot [smooth cycle, tension=0.8] coordinates {
    (p1) (p2) (p4) (p6) (p7) (p8)
  };
  
      \fill[gray!30] plot [smooth cycle, tension=0.8] coordinates {
    (s1) (s2) (s4) (s6) (s7) (s8)
  };


  \node[black, above] at ($(p1)+(0.63,-0.7)$) {\footnotesize{$  C_{\varepsilon}$}};

\draw let \p1=($(xh)-(xs)$),        
          \n1={veclen(\x1,\y1)},     
          \n2={atan2(\y1,\x1)}       
    in [black, line width=1pt]
       (xs) ++({\n1*cos(\n2-30)},{\n1*sin(\n2-30)})
       arc[start angle={\n2-30}, end angle={\n2+05}, radius=\n1];

    
  \draw[dashed,black] (xs) -- (xh);

  \draw let \p1=($(xh)-(xs)$) in
        [->,black!60!black,line width=0.5pt,shorten >=0pt,shorten <=1pt, densely dashed]
        (xs) -- ++({veclen(\x1,\y1)},0) node[midway,below]{$\rho=\|x_h-x_\star\|$};

  \draw[->,black!50!black,line width=0.5pt]
    (xh) -- ($(xh)!0.18!(xs)$) node[below, xshift=12pt] {$\nabla f(x_h)$};

  \fill[black]           (xs)  circle (1pt) node[below left]  {$x_\star$};
  
\draw plot[mark=*, mark size=0.8pt] coordinates {(xh)}; \node[above right] at (xh) {$x_h$};

\draw plot[mark=*, mark size=0.8pt] coordinates {(xst)};
    \node[above , xshift=7pt] at (xst) {$x^\bullet$};

\node[fill=white, align=left, anchor=west, text width=5cm]
    at (8,2) {
      \renewcommand{\baselinestretch}{1.2}\selectfont
      $x_\star$: global minimizer \\
      $x^\bullet$: local maximizer \\
      $C_\varepsilon$: superlevel set near $x^\bullet$ \\
      $x_h$: farthest from $x_\star$ on $C_\varepsilon$ \\
      $\nabla f(x_h)$ points toward $x_\star$\\[0pt]
    };
\end{tikzpicture}
\vspace{-0.1in}
  \caption{
Illustration of the proof for Theorem~\ref{main:thm1}. 
The point \( x_h \) is the farthest from \( x_\star \) among the points in \(C_{\varepsilon} \).
By the method of Lagrange multipliers, its gradient \( \nabla f(x_h) \) points exactly toward \( x_\star \).
}
  \label{fig:thm1}
\end{figure}

\subsection{Simplified existence proof with isolated maximizers}

We now present a geometric construction that produces a hallucinated minimizer.
For clarity, we begin with the case where $f$ admits an isolated local maximizer, namely, a point $x^\bullet$ with an open neighborhood $U$ such that for all $x \in U\setminus\{x^\bullet\}$,
\[
f(x) < f(x^\bullet)
\quad\text{and}\quad
\nabla f(x)\neq 0.
\]
Under this assumption, the following theorem guarantees the existence of a hallucinated minimizer in a large-$\rho$ regime.

\begin{restatable}{theorem}{main}\label{main:thm1}
Let $f\colon\mathbb{R}^d\rightarrow\mathbb{R}$ be continuously differentiable.
Assume $f$ has a global minimizer $x_\star$ (not necessarily unique) and an isolated local maximizer $x^\bullet$.
Then, there exists a nontrivial interval of radii $\rho$ for which hallucinated minimizers exist, and this interval contains values with $\rho \geq \|x_\star-x^\bullet\|$.
\end{restatable}

\begin{proof}[Sketch of proof]
We provide a brief sketch of the argument, with full details deferred to Appendix~\ref{app:thm1pf}. Figure~\ref{fig:thm1} illustrates the key idea.

Let $\varepsilon>0$ and define $C_\varepsilon$ as the $(f(x^\bullet)-\varepsilon)$-superlevel set restricted to a neighborhood of the isolated local maximizer $x^\bullet$.
For sufficiently small $\varepsilon > 0$, the set $C_\varepsilon$ is compact  and satisfies:  
$(i)$ $f(x^\bullet)-\varepsilon \le f(x) \le f(x^\bullet)$ for all $x\in C_\varepsilon$; 
$(ii)$ $\nabla f$ does not vanish on $C_\varepsilon\setminus\{x^\bullet\}$; and  
$(iii)$ $f(x) = f(x^\bullet)-\varepsilon$ for all $x\in\partial C_\varepsilon$.

Next, consider $g(x)=\|x-x_\star\|^2$ and choose
\[
x_h\in \argmax_{x\in C_\varepsilon}g(x).
\]
Then $x_h\in\partial C_\varepsilon$.
Moreover, since $\nabla f(x_h)\neq 0$ and $\partial C_\varepsilon$ locally coincides with the level set
$\{x:\ f(x)=f(x^\bullet)-\varepsilon\}$ near $x_h$, 
there exists a neighborhood $V$ of $x_h$ such that
\[
\Sigma:=\{x\in V:\ f(x)=f(x^\bullet)-\varepsilon\}
\]
is an embedded $C^1$ hypersurface and $\Sigma = V\cap \partial C_\varepsilon$.

Since $x_h$ maximizes $g$ over $C_\varepsilon$ and lies on $\partial C_\varepsilon$, it is also a local maximizer of $g$ over $\Sigma$.
By the method of Lagrange multipliers, there exists $\lambda\in\mathbb{R}$ such that
\[
\nabla g(x_h)+\lambda \nabla f(x_h)=0,
\]
equivalently,
$2(x_\star-x_h)=\lambda \nabla f(x_h)$.
Furthermore, one can show that $\lambda>0$ (see Lemma~\ref{lem:positive} in Appendix~\ref{app:thm1pf}).

Let $\rho:=\|x_\star-x_h\|$.
Taking norms gives $\lambda=\frac{2\rho}{\|\nabla f(x_h)\|}$, and hence
\[
x_\star = x_h + \rho \frac{\nabla f(x_h)}{\|\nabla f(x_h)\|}=x_h^{+}.
\]
Therefore, $\nabla f(x_h^{+})=\nabla f(x_\star)=0$, while $\nabla f(x_h)\neq 0$ by construction.
Moreover, $f^{\mathrm{SAM}}(x_h)=f(x_h^{+})=f(x_\star) = \min f$.
Thus, $x_h$ is a (global) minimizer of $f^{\mathrm{SAM}}$ and is therefore a hallucinated minimizer.
Finally, since $x^\bullet\in C_\varepsilon$ and $x_h$ maximizes $\|x-x_\star\|$ over $C_\varepsilon$, 
we have
$\rho=\|x_\star-x_h\|\ge \|x_\star-x^\bullet\|$.

As $\varepsilon$ varies while satisfying $(i)$--$(iii)$, the induced radius
$\rho(\varepsilon):=\max_{x\in C_\varepsilon}\|x-x_\star\|$
depends continuously on $\varepsilon$ and is strictly increasing (Lemma~\ref{lem:rho-mono} in Appendix~\ref{app:thm1pf}).
This yields an interval of radii $\rho$ for which hallucinated minimizers exist.
\end{proof}

Importantly, the existence of a hallucinated minimizer also holds when $x_\star$ is a \emph{local} minimizer,
provided that $f$ has a locally Lipschitz gradient; see Appendix~\ref{app:thm1ext} for details.

\paragraph{When hallucinated minimizers arise.}
The proof of Theorem~\ref{main:thm1} reveals conditions under which hallucinated minimizers are likely to arise. Figure~\ref{fig:thm1} illustrates the core mechanism: the construction enforces $x_\star =  x_h + \rho \frac{\nabla f(x_h)}{\|\nabla f(x_h)\|} = x_h^+$. This means that at $x_h$, the gradient points directly toward $x_\star$.
Near a minimizer, gradients typically point away from the minimizer, so such an alignment is hard to obtain locally.
The presence of a nearby maximizer provides a region where gradients can align in the required direction, which explains why nonconvexity is essential.
The construction is also inherently a large-$\rho$ phenomenon: it yields radii satisfying $\rho=\|x_\star-x_h\|\ge \|x_\star-x^\bullet\|$, and the hallucinated minimizer $x_h$ lies near the maximizer region (inside $C_\varepsilon$).

\subsection{Existence with local maximizer sets}
\label{ss:general-existence}

Theorem~\ref{main:thm1} assumes that the local maximizer is isolated.
In many modern objectives, however, symmetries and flat directions can produce \emph{non-isolated} maximizers
(e.g., a plateau or a manifold of maximizers).
We therefore extend the existence result to the case of a \emph{local maximizer set}, i.e., a connected set that maximizes $f$ within a neighborhood.

\begin{definition}[Local maximizer set]
Let $f\colon \mathbb{R}^d\rightarrow\mathbb{R}$ be continuous.
A nonempty connected set $X\subseteq \mathbb{R}^d$ is a \emph{local maximizer set} of $f$ if there exists $\delta>0$ such that
\[
X=\argmax_{y\in \mathcal{N}_\delta (X)}f(y).
\]
In this case, $f$ is constant on $X$, and we denote this common value by $f(X)$.
\end{definition}

The following theorem shows that hallucinated minimizers can still arise when the maximizers are non-isolated.

\begin{restatable}{theorem}{maintwo}\label{thm:exist}
Let $f\colon\mathbb{R}^d\rightarrow\mathbb{R}$ be real-analytic.
Assume $f$ has a global minimizer $x_\star$ (not necessarily unique) and a bounded local maximizer set $X$.
Then, there exists a nontrivial interval of radii $\rho$ for which hallucinated minimizers exist, and this interval contains values with $\rho \geq d(x_\star, X)$.
\end{restatable}

The proof of Theorem~\ref{thm:exist} is provided in Appendix~\ref{app:thm2pf}.

\paragraph{What changes relative to the isolated case.}
When $X$ is not isolated, critical points may in principle accumulate near $X$,
which could invalidate the ``regular level set'' argument needed at the boundary $\partial C_\varepsilon$.
Real-analyticity rules out such pathological accumulation:
via the \L{}ojasiewicz inequality,
we show that any critical point sufficiently close to $X$ with function value near $f(X)$ must lie in $X$ itself  (Lemma~\ref{lem:nocritical} in Appendix~\ref{app:thm2pf}).
This ensures $\nabla f\neq 0$ on $\partial C_\varepsilon$, enabling the same Lagrange-multiplier construction as in Theorem~\ref{main:thm1}.
As in the isolated case, varying $\varepsilon$ yields a strictly increasing, continuous radius map $\rho(\varepsilon)=\max_{x\in C_\varepsilon}\|x-x_\star\|$, hence an interval of admissible radii.

\paragraph{On real-analyticity of $f$.}
The real-analyticity assumption is used to invoke the {\L}ojasiewicz inequality and exclude pathological
accumulations of critical points near $X$.
For finite-dataset objectives, this assumption holds whenever both the loss and the network mapping are built
from real-analytic primitives (e.g., \texttt{Tanh}/sigmoid/softplus/GELU activations with standard smooth losses);
see Appendix~\ref{app:NN} for a self-contained sufficient condition.
In particular, it applies directly to our full-batch MNIST experiment with \texttt{Tanh} activations
(Section~\ref{sub:exp1}).
While this excludes nonsmooth architectures such as ReLU and max-pooling, we empirically observe hallucinated minimizers even in ReLU-based networks (Section~\ref{sec:expnn}).

\section{Dynamical and geometric properties of hallucinated minimizers}
\label{sec:3}

In this section, we establish  a dynamical property of the SAM iterates as well as a finer geometric property of hallucinated minimizers.

\subsection{Hallucinated minimizers can be attractors}
\label{sec:attractor}

Section~\ref{sec:2} established that hallucinated minimizers can exist for suitable radii.
We now show that this phenomenon is not merely an algebraic possibility:
an isolated connected component of hallucinated minimizers can be \emph{locally attracting} for the
 (full-batch) SAM update.

\begin{restatable}{theorem}{attractor}\label{thm:attractor}
Suppose $f\colon\mathbb{R}^d\rightarrow\mathbb{R}$ is real-analytic, and let
$H\subset \mathbb{R}^d$ be a bounded, connected set of hallucinated minimizers (for $f$ at radius $\rho$).
Assume there exists $\delta >0$ such that the $\delta$-neighborhood of $H$
contains no minimizers of $f^{\mathrm{SAM}}$ other than those already in $H$.
Assume further that every $x_h \in H$ satisfies
\[
1+\rho\,\lambda_{\min}\!\bigl(\mathrm{Sym}(\nabla u(x_h))\bigr) > 0.
\]
If the initialization $x_0$ is chosen sufficiently close to $H$, then there exists a sufficiently small fixed
step size $\eta_k=\eta>0$ such that the SAM iterates \eqref{eq:sam-update-sec3} satisfy $d(x_k,H)\to 0$.
\end{restatable}

We defer the proof to Appendix~\ref{app:thm4pf}.

\paragraph{Interpreting the stability condition.}
Unlike gradient descent on $f^{\mathrm{SAM}}$, the update \eqref{eq:sam-update-sec3} does not a priori guarantee
descent of the surrogate loss $f^{\mathrm{SAM}}$.
The condition
$1+\rho\,\lambda_{\min}(\mathrm{Sym}(\nabla u(x_h)))>0$ ensures a local \emph{acute-angle} relation between
the surrogate gradient $\nabla f^{\mathrm{SAM}}(x)$ and the SAM gradient $\nabla f (x^+)$.
Indeed, for any $x$ with $\nabla f(x)\neq 0$ for which $u$ is differentiable, we have
$\nabla f^{\mathrm{SAM}}(x)=\bigl(I+\rho\,\nabla u(x)\bigr)^\top \nabla f(x^{+})$.
Therefore, with $g:=\nabla f(x^+)$,
\[
\langle \nabla f^{\mathrm{SAM}}(x),\, g\rangle
= g^\top\!\bigl(I+\rho\,\mathrm{Sym}(\nabla u(x))\bigr)g.
\]
Hence, on a neighborhood where
$I+\rho\,\mathrm{Sym}(\nabla u(x))\succeq \gamma I$ for some $\gamma>0$,
we have $\langle \nabla f^{\mathrm{SAM}}(x), g\rangle \ge \gamma\|g\|^2$.
Combining this with $L$-smoothness of $f^{\mathrm{SAM}}$ on that neighborhood yields the descent estimate
\[
f^{\mathrm{SAM}}(x-\eta g)
\le f^{\mathrm{SAM}}(x)
-\eta\Bigl(\gamma-\tfrac{L\eta}{2}\Bigr)\|g\|^2,
\]
so sufficiently small steps decrease $f^{\mathrm{SAM}}$ unless $g=0$.
The proof of Theorem~\ref{thm:attractor} formalizes this argument near $H$ and uses the assumption that $H$
is an isolated minimizer component of $f^{\mathrm{SAM}}$ to conclude $d(x_k,H)\to 0$.

\paragraph{A simple sufficient condition.}
Directly verifying the eigenvalue condition in Theorem~\ref{thm:attractor} can be difficult in high dimensions, so we record a convenient sufficient condition.
At any $x$ with $\nabla f(x)\neq 0$ and $f\in C^2$ in a neighborhood of $x$, we have
\[
\nabla u(x)
=\frac{1}{\|\nabla f(x)\|}\bigl(I-u(x)u(x)^\top\bigr)\nabla^2 f(x),
\]
and thus $\|\nabla u(x)\|\le \|\nabla^2 f(x)\|/\|\nabla f(x)\|$.
Using $\lambda_{\min}(\mathrm{Sym}(A))\ge -\|A\|$,
the stability condition
$1+\rho\,\lambda_{\min}(\mathrm{Sym}(\nabla u(x_h)))>0$
is implied by
\[
\rho\,\|\nabla^2 f(x_h)\|<\|\nabla f(x_h)\|.
\]
This ratio-type condition is meaningful at hallucinated minimizers since $\|\nabla f(x_h)\|>0$ by definition.

\subsection{Hallucinated minimizers can form manifolds}
\label{ss:hm-manifold}

We next refine the geometry of hallucinated minimizers.
In the toy example of Figure~\ref{fig:toy}, they are not isolated and instead form a curve-like set.
The following result shows that this is not accidental:
whenever the original objective has a manifold of global minimizers, the set of hallucinated minimizers
can inherit the \emph{same manifold dimension}.

\begin{restatable}{theorem}{manifold}\label{thm:manifold}
Assume $f:\mathbb{R}^d\to\mathbb{R}$ is $C^2$.
Let $\mathcal{M}\subseteq \argmin f$ be a nonempty embedded $m$-dimensional $C^1$ manifold,
and fix $x_\star\in \mathcal{M}$.
Suppose there exists a point $x_h\in\mathbb{R}^d$ such that
\[
x_h^{+}=x_\star
\quad\text{and}\quad
\nabla f(x_h)\neq 0.
\]
If in addition $I+\rho\nabla u(x_h)$ is nonsingular, then there exist neighborhoods
$V$ of $x_h$ and $U$ of $x_\star$ such that the set
\[
\mathcal{H}:=\{x\in V:\ x^{+}\in \mathcal{M}\cap U\}
\]
is an embedded $m$-dimensional $C^1$ manifold.
Moreover, every $x\in\mathcal{H}$ is a hallucinated minimizer at radius $\rho$.
\end{restatable}

Its proof is provided in Appendix~\ref{app:thm3pf}.

\paragraph{Interpretation.}
The nonsingularity of $I+\rho\nabla u(x_h)$ exactly means that the perturbation map
$\Phi_\rho(x):=x+\rho u(x)$ is locally invertible at $x_h$ (inverse function theorem).
Thus, near $x_h$, the preimage $\Phi_\rho^{-1}(\mathcal{M})$ is a smooth manifold of the same dimension as $\mathcal{M}$.

\begin{figure}[t!]
    \centering
    \begin{subfigure}[t]{0.3\textwidth}
        \centering
        \includegraphics[width=\linewidth]{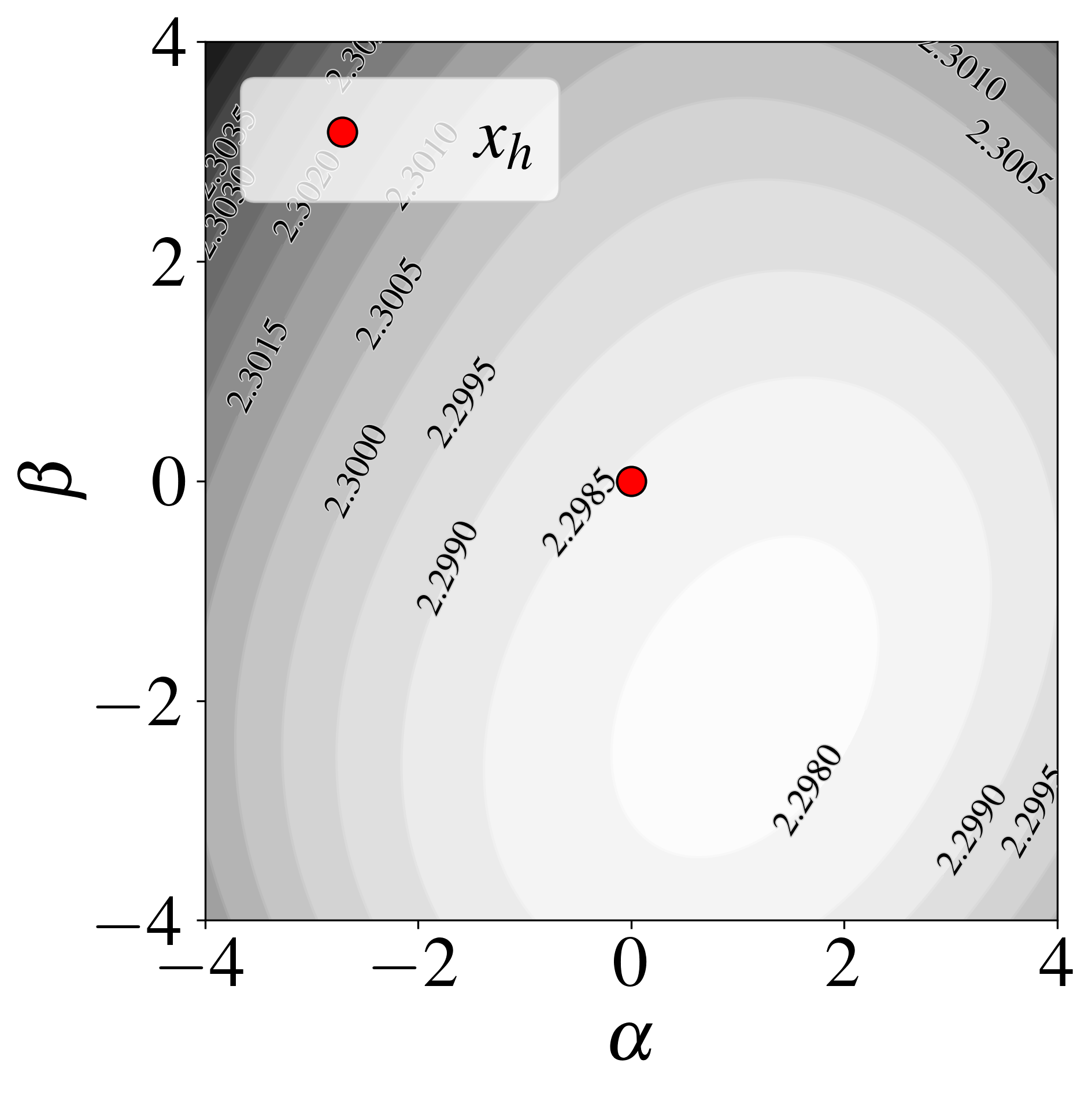}
        \caption{Original loss $f$ around $x_h$}
        \label{fig:HM0}
    \end{subfigure}
    \hfill
    \begin{subfigure}[t]{0.3\textwidth}
        \centering
        \includegraphics[width=\linewidth]{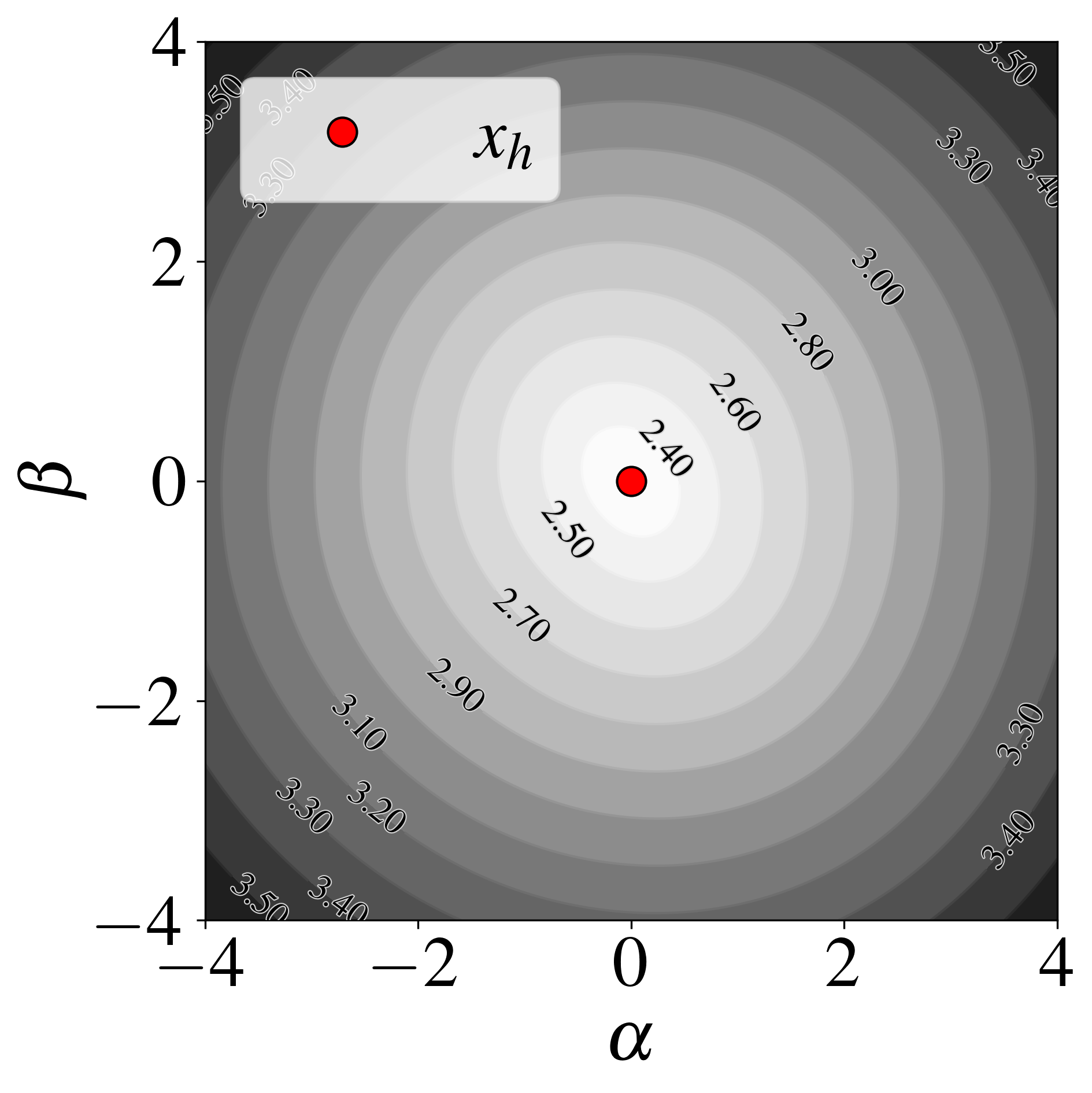}
        \caption{SAM loss $f^{\mathrm{SAM}}$ around $x_h$}
        \label{fig:HM1}
    \end{subfigure}
    \hfill
    \begin{subfigure}[t]{0.32\textwidth}
        \centering
        \includegraphics[width=\linewidth]{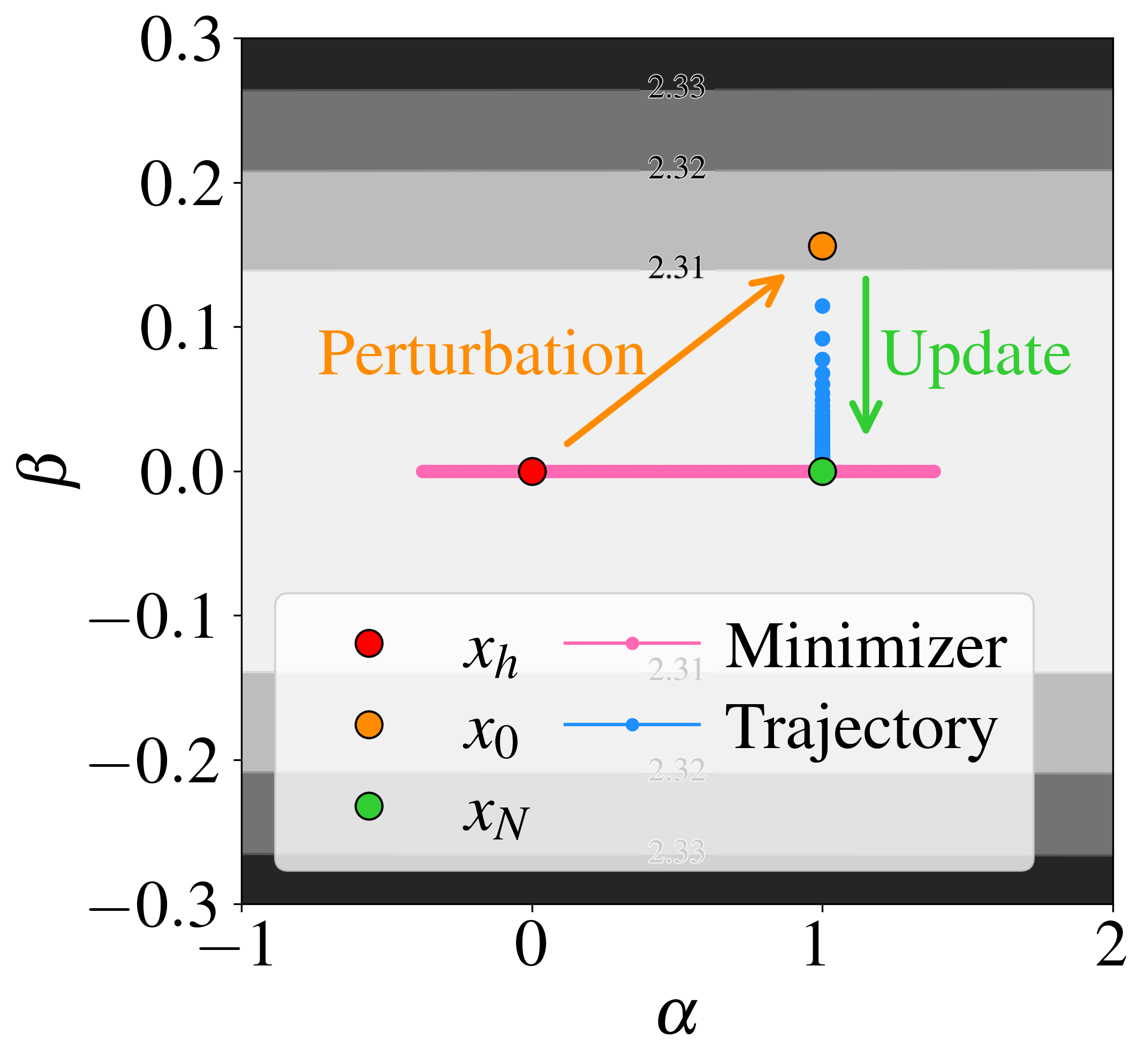}
        \caption{SAM iterates converge back to the hallucinated minimizer set.}
        \label{fig:HM2}
    \end{subfigure}

    \caption{
    Visualizations of $f$ and $f^{\mathrm{SAM}}$ around the hallucinated minimizer $x_h$.
    Plots  (a) and (b) are taken on a two-dimensional plane defined by $x_h$ and two random directions. These show that $x_h$ is not stationary for $f$, while it appears as a minimizer of $f^{\mathrm{SAM}}$ on the same plane.
    Plot (c) depicts $f^{\mathrm{SAM}}$ on the two-dimensional plane containing $x_h$, $x_0$, and $x_N$, where $x_0$ is a small perturbation of $x_h$ and $x_N$ is obtained after $N=1000$ SAM steps from $x_0$. The pink horizontal line segment indicates the set of hallucinated minimizers, showing that the SAM trajectory converges back to this set.    }
    \label{fig:HMexist}
\end{figure}

\paragraph{Connection to our existence results.}
Under the assumptions of Theorem~\ref{thm:exist}, we may choose $x_\star$ to lie on any manifold component
$\mathcal{M}\subseteq \argmin f$ (when such a component exists).
The construction in Section~\ref{sec:2} then produces an $x_h$ with $x_h^{+}=x_\star$,
and Theorem~\ref{thm:manifold} explains why the resulting hallucinated minimizers can appear as
curve-/surface-like sets in practice (e.g., Figure~\ref{fig:toy} and Figure~\ref{fig:HMexist}).

\section{Experiments}\label{sec:expnn}

We empirically validate the mechanisms identified in Sections~\ref{sec:2}--\ref{sec:3} on neural network objectives.
The implementation of our experiments is available through a repository at \url{https://anonymous.4open.science/r/SAM-can-hallucinate-minimizers-4B82/}.

\vspace{-0.1in}

\paragraph{Diagnostics.}
To diagnose convergence to hallucinated minimizers along a trajectory $\{x_k\}$, we monitor two pointwise mismatch quantities
$\texttt{GradRatio}(x):=\frac{\|\nabla f(x^{+})\|}{\|\nabla f(x)\|+\epsilon}$ and
$\texttt{LossGap}(x):= f(x^{+})-f(x)$,
with $\epsilon>0$ a small numerical stabilizer.
At a hallucinated minimizer $x_h$, we expect $\|\nabla f(x_h^{+})\|\approx 0$ while $\|\nabla f(x_h)\|>0$, hence $\texttt{GradRatio}(x_h)\ll 1$.
Moreover, since $x_h^{+}$ lies near a (local) minimizer of $f$, the ``ascent'' perturbation can paradoxically \emph{decrease} the loss, yielding $\texttt{LossGap}(x_h)\ll 0$.
We emphasize that in neural networks these diagnostics are \emph{evidence consistent with} hallucinated minimizers, rather than a formal certification.

\begin{figure*}[t]
  \centering

  \begin{minipage}[t]{0.245\textwidth}
    \centering \textbf{$\rho = 1.0$}
  \end{minipage}
  \begin{minipage}[t]{0.245\textwidth}
    \centering \textbf{$\rho = 1.3$}
  \end{minipage}
  \begin{minipage}[t]{0.245\textwidth}
    \centering \textbf{$\rho = 1.6$}
  \end{minipage}
  \begin{minipage}[t]{0.245\textwidth}
    \centering \textbf{$\rho = 1.9$}
  \end{minipage}

  \vspace{1mm}

  \begin{subfigure}[t]{\textwidth}
    \centering
    \begin{minipage}[t]{0.245\textwidth}
      \includegraphics[width=\linewidth]{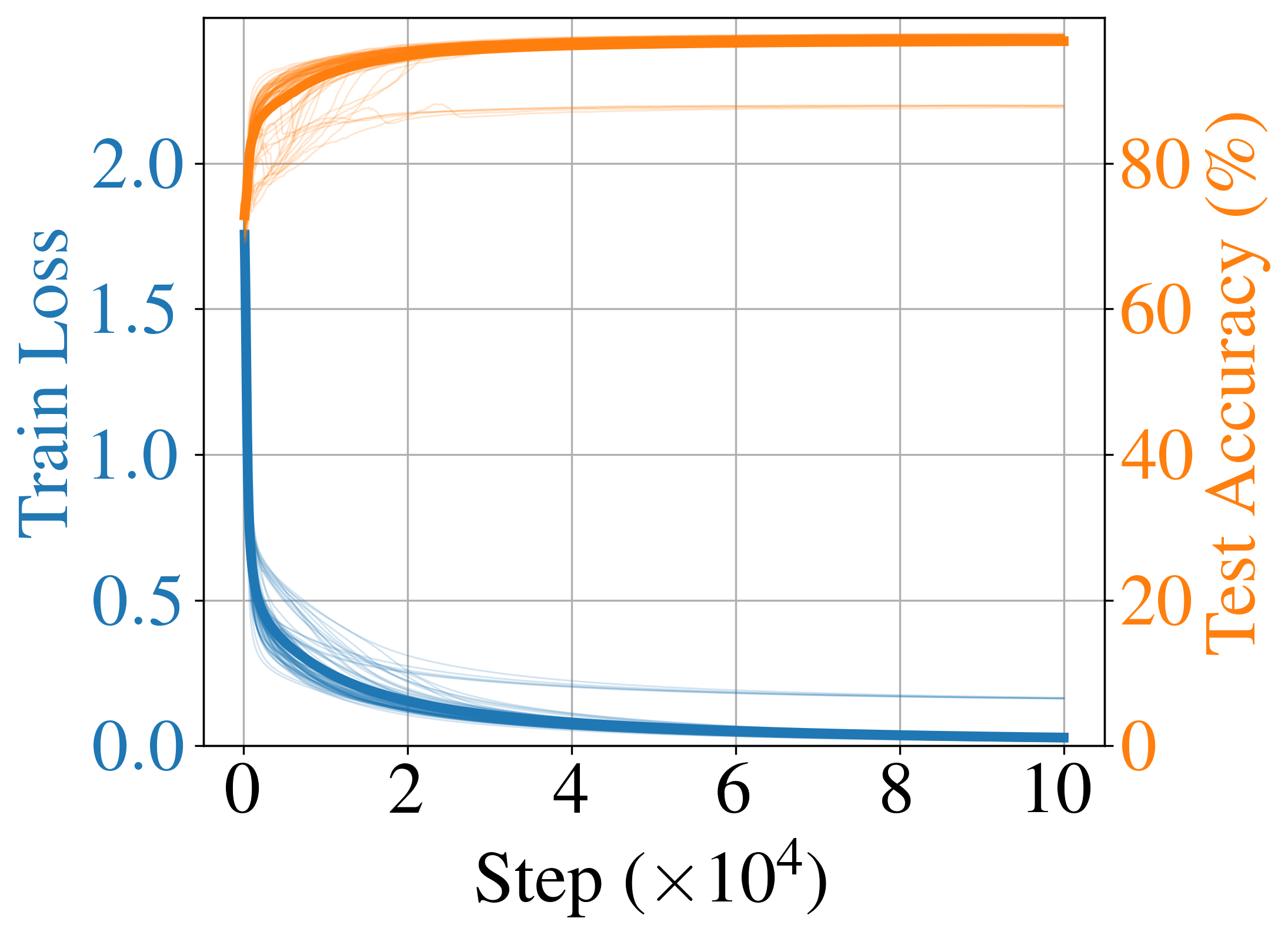}
    \end{minipage}
    \begin{minipage}[t]{0.245\textwidth}
      \includegraphics[width=\linewidth]{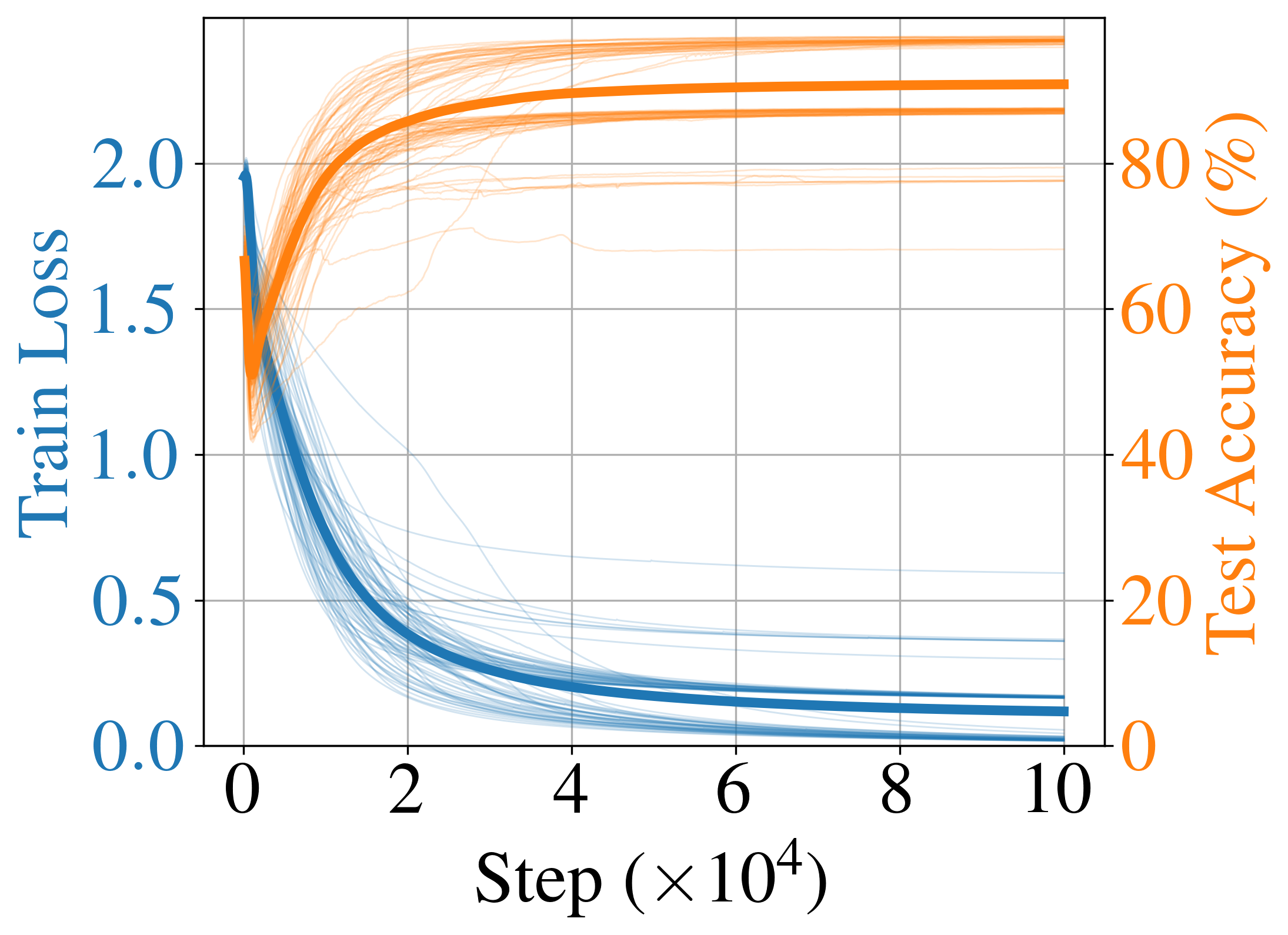}
    \end{minipage}
    \begin{minipage}[t]{0.245\textwidth}
      \includegraphics[width=\linewidth]{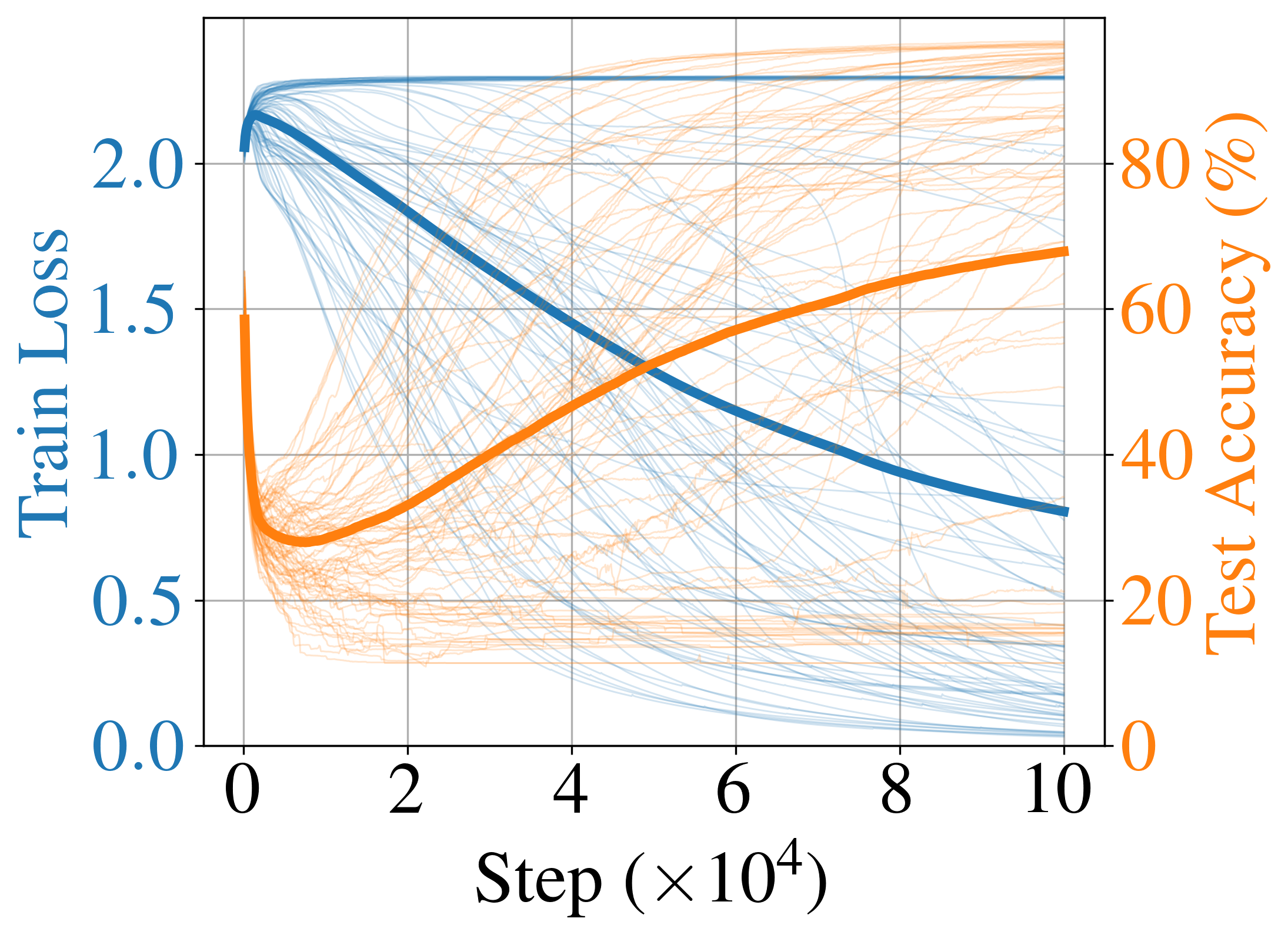}
    \end{minipage}
    \begin{minipage}[t]{0.245\textwidth}
      \includegraphics[width=\linewidth]{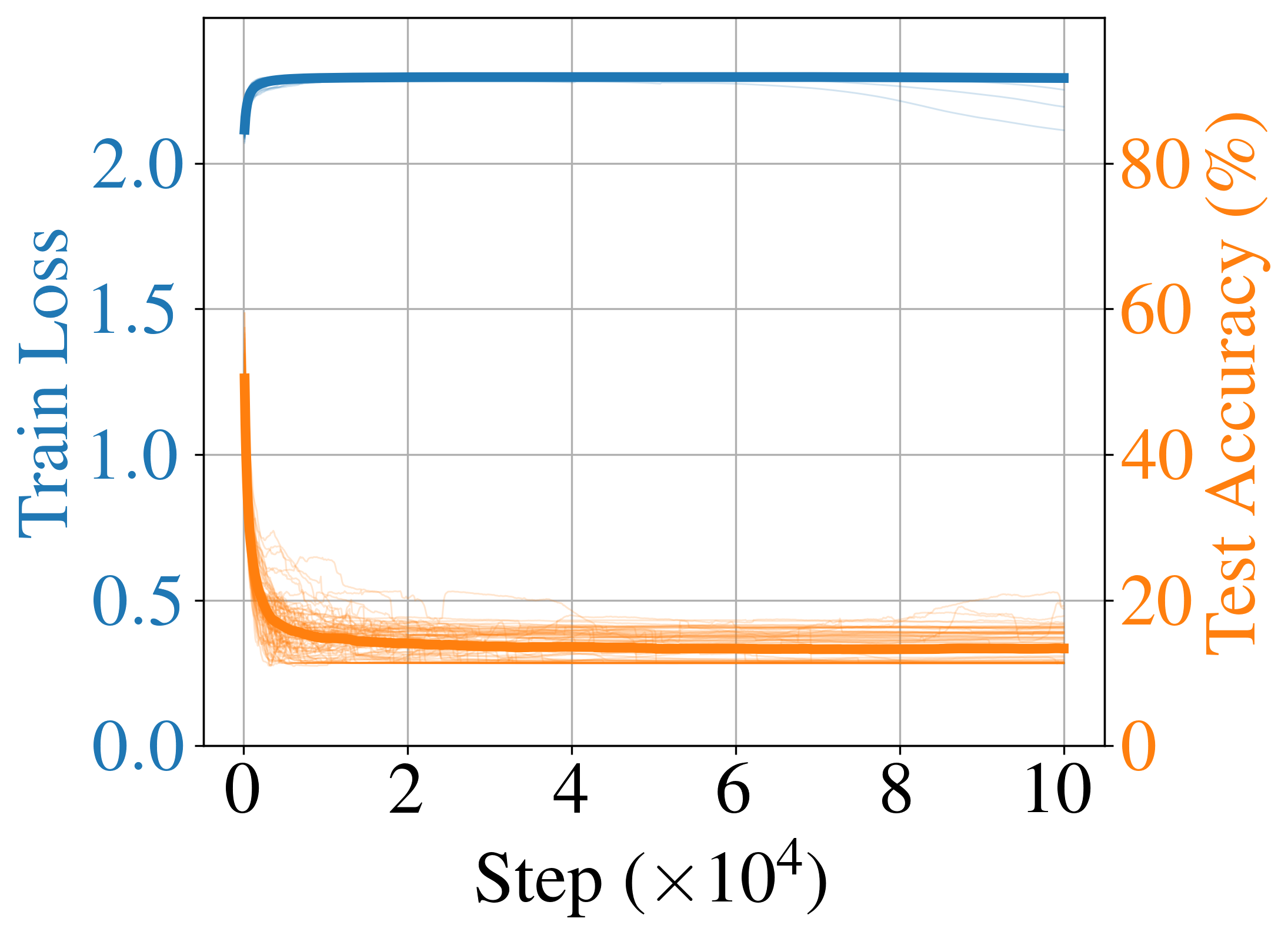}
    \end{minipage}
    \subcaption{SAM-only}
  \end{subfigure}

  \vspace{4mm}

  \begin{subfigure}[t]{\textwidth}
    \centering
    \begin{minipage}[t]{0.243\textwidth}
      \includegraphics[width=\linewidth]{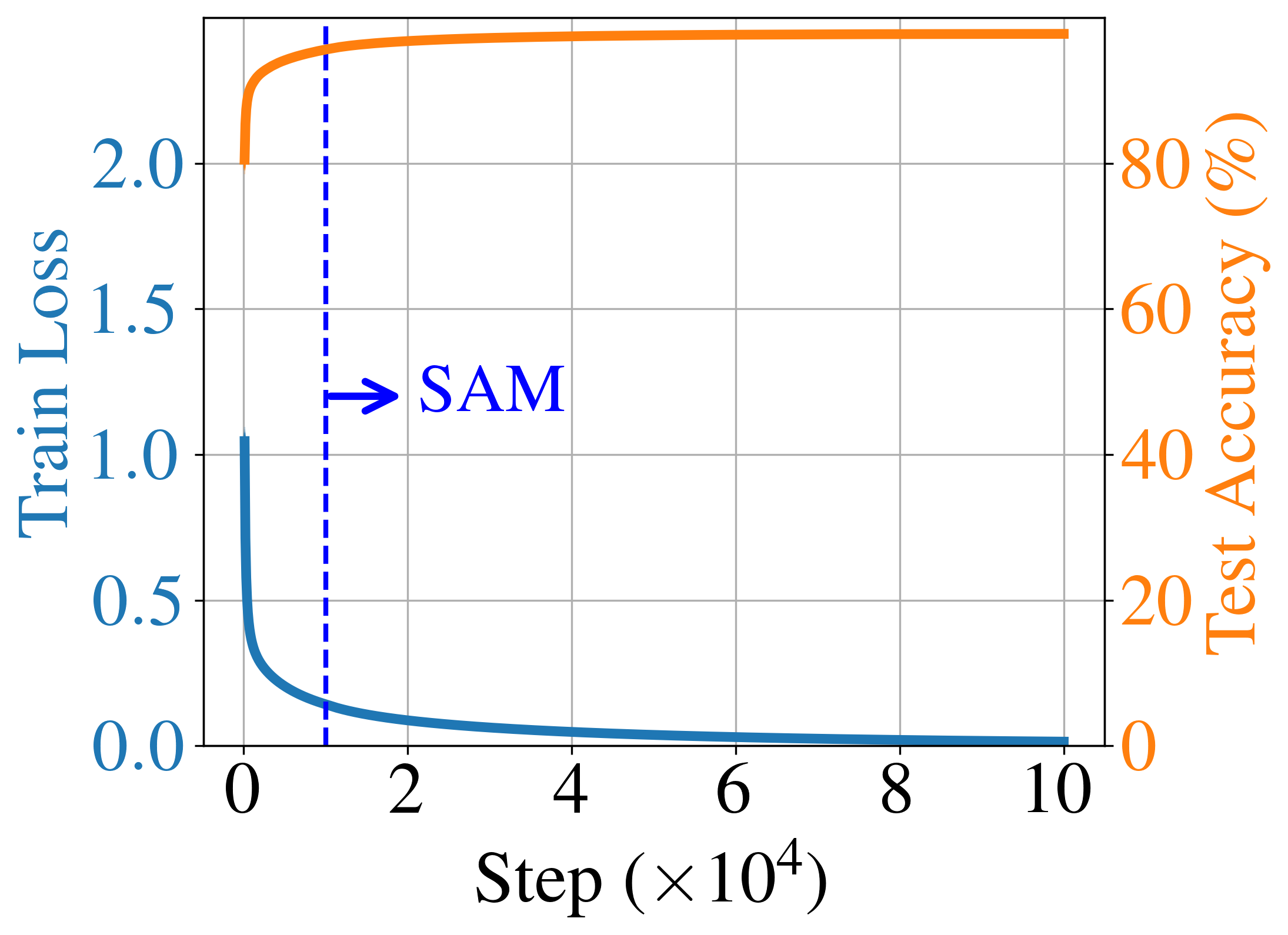}
    \end{minipage}
    \begin{minipage}[t]{0.243\textwidth}
      \includegraphics[width=\linewidth]{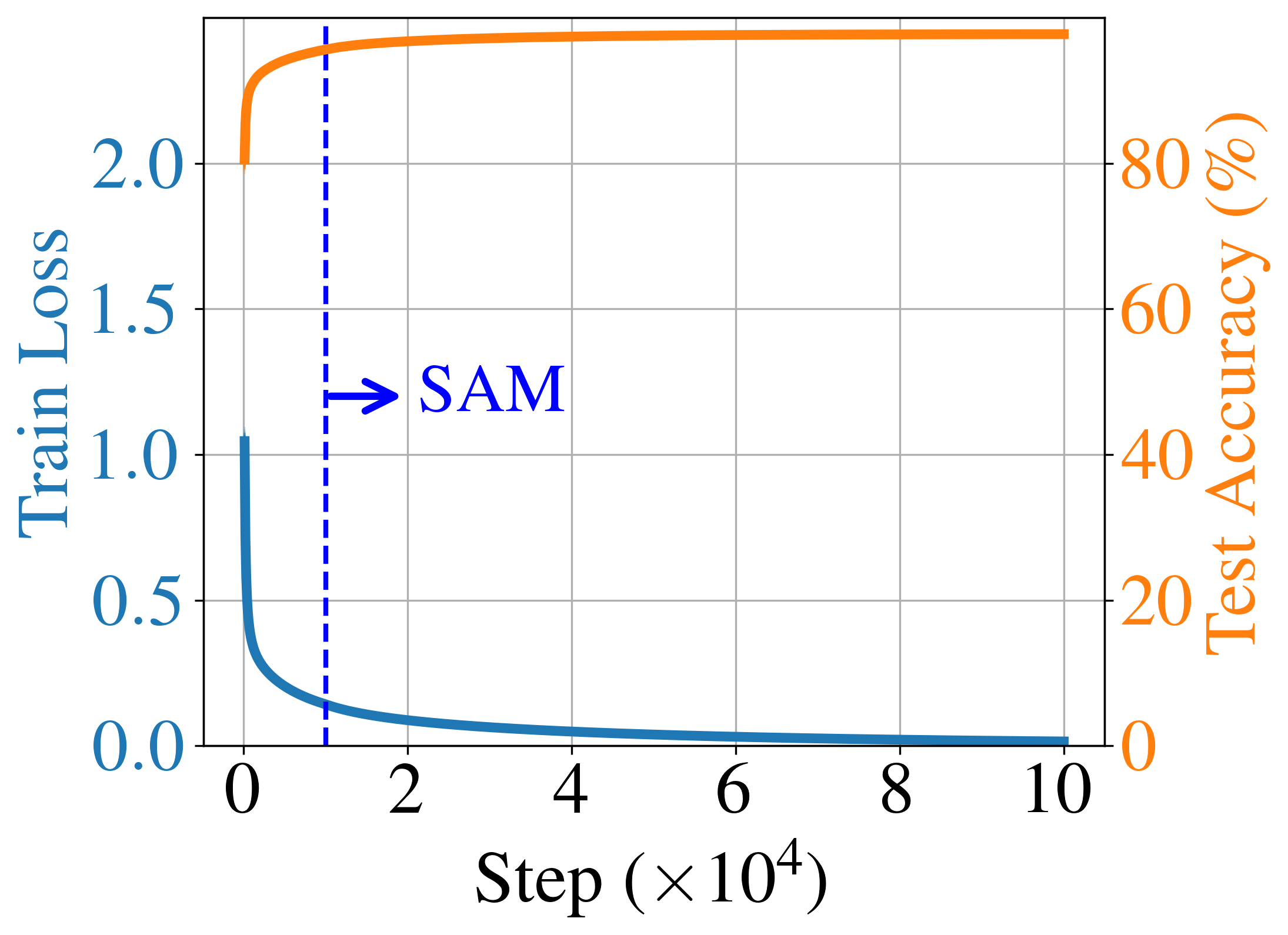}
    \end{minipage}
    \begin{minipage}[t]{0.243\textwidth}
      \includegraphics[width=\linewidth]{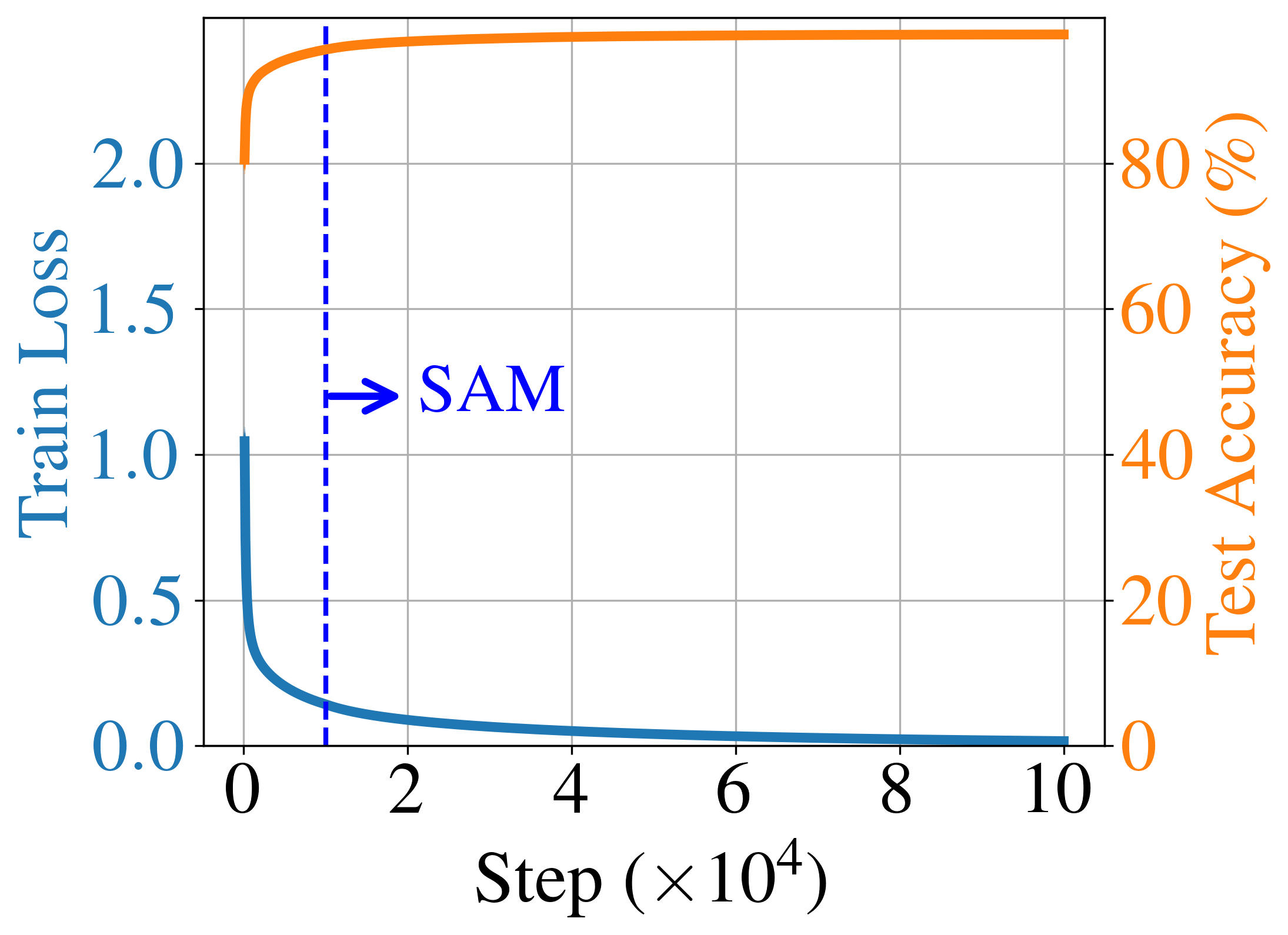}
    \end{minipage}
    \begin{minipage}[t]{0.243\textwidth}
      \includegraphics[width=\linewidth]{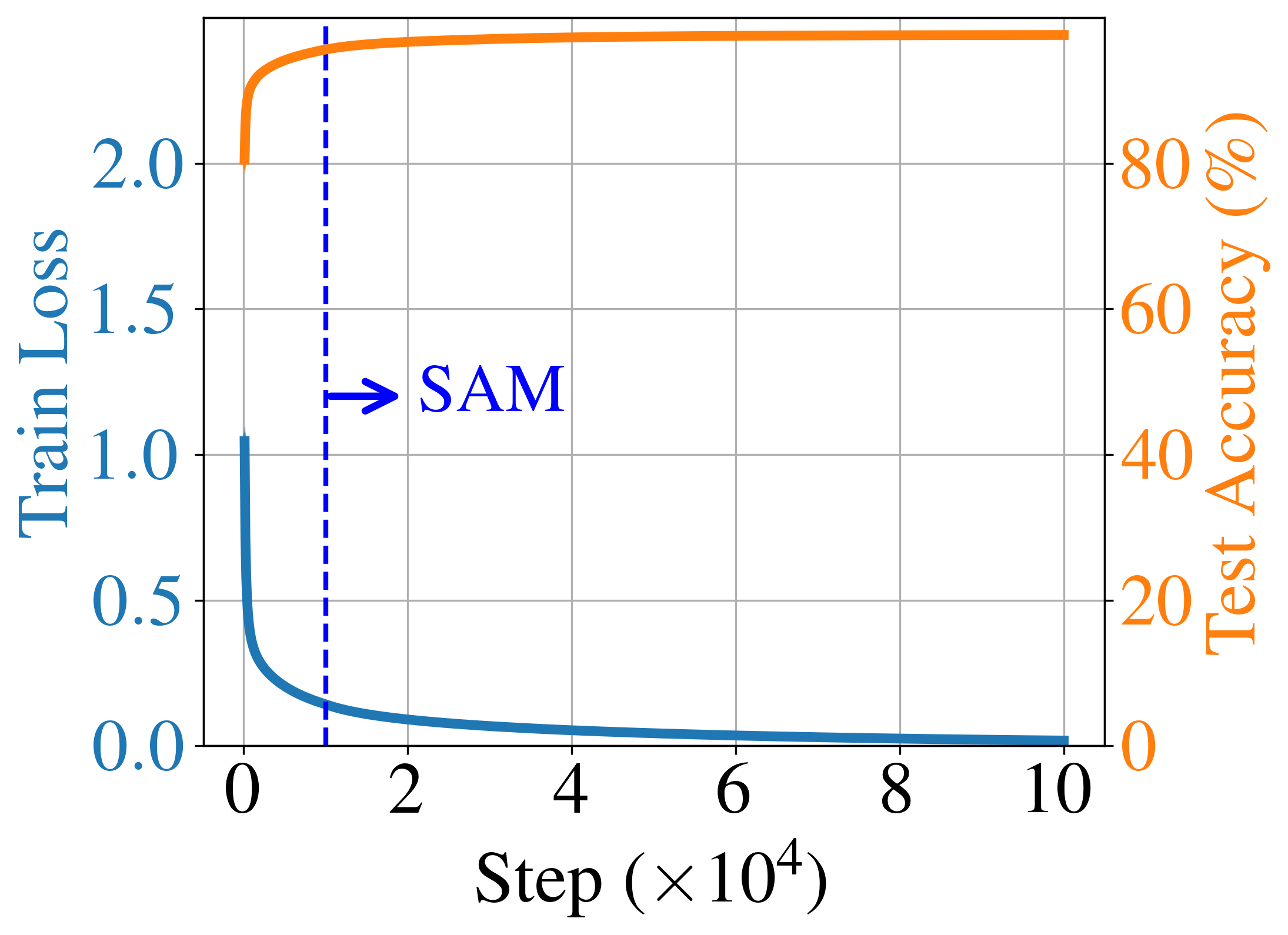}
    \end{minipage}
    \subcaption{GD $\rightarrow$ SAM switching}
    \label{fig:det_samb}
  \end{subfigure}

  \caption{
Full-batch MNIST (2-layer \texttt{Tanh} network): training loss and test accuracy for SAM-only (top) and GD$\rightarrow$SAM switching (bottom) across perturbation radii.
Thin curves are individual seeds (80 total) and bold curves are means.
At larger $\rho$, SAM-only frequently plateaus at positive loss and suffers accuracy collapse, while switching remains stable across $\rho$.
  }
  \label{fig:det_sam}
  \vspace{-0.1in}
\end{figure*}

\subsection{SAM can converge to hallucinated minimizers}
\label{sub:exp1}

We first provide a direct verification in a smooth, full-batch setting, and then consider larger networks and stochastic settings. 

\paragraph{Direct verification in a smooth full-batch setting.}
We train a two-layer network with \texttt{Tanh} activations on MNIST using full-batch SAM.
This setting yields a smooth (indeed, real-analytic) objective, aligning with the assumptions underlying our existence and attractor results.
Implementation details are in Appendix~\ref{app:full_mnist}.
With perturbation radius $\rho=1.8$, we run training for a long horizon until the trajectory numerically stabilizes near a point $x_h$.
At this point, the shifted gradient is nearly zero while the original gradient is not:
$\|\nabla f(x_h^+)\| = 4.8\times 10^{-9}$, and $\|\nabla f(x_h)\| = 6.27\times 10^{-2}$.
Thus, the practical SAM update becomes (nearly) stationary at $x_h$ even though $x_h$ is not a critical point of $f$.

To further probe the local geometry, we visualize two-dimensional loss landscapes around $x_h$ following \citet{li2018visualizing}.
Specifically, we consider planes of the form $x(\alpha,\beta)=x_h+\alpha d_1+\beta d_2$ where $d_1,d_2$ are random orthogonal directions of matched norm.
Figure~\ref{fig:HMexist} shows that $x_h$ is \emph{not} a minimizer of $f$ on this plane, while it appears as a local minimizer of the SAM loss $f^{\mathrm{SAM}}(x)=f(x^{+})$ on the same slice.
This is consistent with $x_h$ behaving as a hallucinated minimizer.

\paragraph{Local attraction in practice.}
To test whether the observed point is locally attracting (rather than an artifact of a single run),
we perturb $x_h$ by a random displacement of size $0.1$ to form $x_0$, run $N=1000$ additional SAM steps,
and examine the resulting point $x_N$.
Figure~\ref{fig:HM2} shows that the perturbed initialization returns to the same near-minimizer region of $f^{\mathrm{SAM}}$ around $x_h$ within this local slice,
suggesting robust local attraction, consistent with Theorem~\ref{thm:attractor}.
Moreover, the near-minimizer region is visibly non-isolated on the slice, qualitatively aligning with the manifold mechanism in Theorem~\ref{thm:manifold} (see also Appendix~\ref{app:full_mnist}).

\paragraph{How common is the failure mode? (full-batch MNIST)}
We next sweep $\rho\in\{1.0,1.3,1.6,1.9\}$ and run 80 independent seeds for 100{,}000 iterations.
Figure~\ref{fig:det_sam} reports training loss and test accuracy.
For $\rho=1.0$, trajectories consistently reach near-zero loss.
As $\rho$ increases, a substantial fraction of runs plateau at positive training loss and exhibit severe accuracy degradation,
matching the trajectory-level signature of convergence to a spurious stationary region for the practical SAM update.
Notably, at the same $\rho$, different seeds can yield either near-zero loss or a plateau,
consistent with the basin-of-attraction viewpoint in Theorem~\ref{thm:attractor}.

\begin{table}[t]
\centering
\caption{SAM with full-batch gradients on CIFAR-10/100 (ResNet-18): metrics at the final iterate $x_T$ across perturbation radii $\rho$. Here, $x_T^+=x_T+\rho\,u(x_T)$; $f(x_T^+)$ is the SAM loss; and $\nabla f(x_T^+)$ is the SAM gradient.}
\label{tab:final_metrics_resnet18}
\resizebox{0.85\textwidth}{!}{%
\begin{tabular}{ccccccccc}
\toprule
& \multicolumn{4}{c}{CIFAR-10} & \multicolumn{4}{c}{CIFAR-100} \\
\cmidrule(lr){2-5}\cmidrule(lr){6-9}
$\rho$
& $f(x_T)$ & $f(x_T^+)$ & $\lVert\nabla f(x_T)\rVert$ & $\lVert\nabla f(x_T^+)\rVert$
& $f(x_T)$ & $f(x_T^+)$ & $\lVert\nabla f(x_T)\rVert$ & $\lVert\nabla f(x_T^+)\rVert$ \\
\midrule
0.0 & 0.001 & 0.001 & 0.006 & 0.006 & 0.009 & 0.009 & 0.058 & 0.058 \\
0.5 & 0.081 & 0.001 & 3.190 & 0.015 & 0.050 & 0.009 & 0.944 & 0.100 \\
1.0 & 0.120 & 0.001 & 3.131 & 0.015 & 0.132 & 0.011 & 4.034 & 0.134 \\
1.5 & 0.773 & 0.001 & 11.932 & 0.011 & 0.419 & 0.012 & 8.119 & 0.136 \\
2.0 & 0.219 & 0.002 & 4.033 & 0.021 & 0.828 & 0.014 & 13.722 & 0.123 \\
\bottomrule
\end{tabular}%
}
\end{table}

\begin{figure}[t] 
    \centering
    \includegraphics[width=0.6\linewidth]{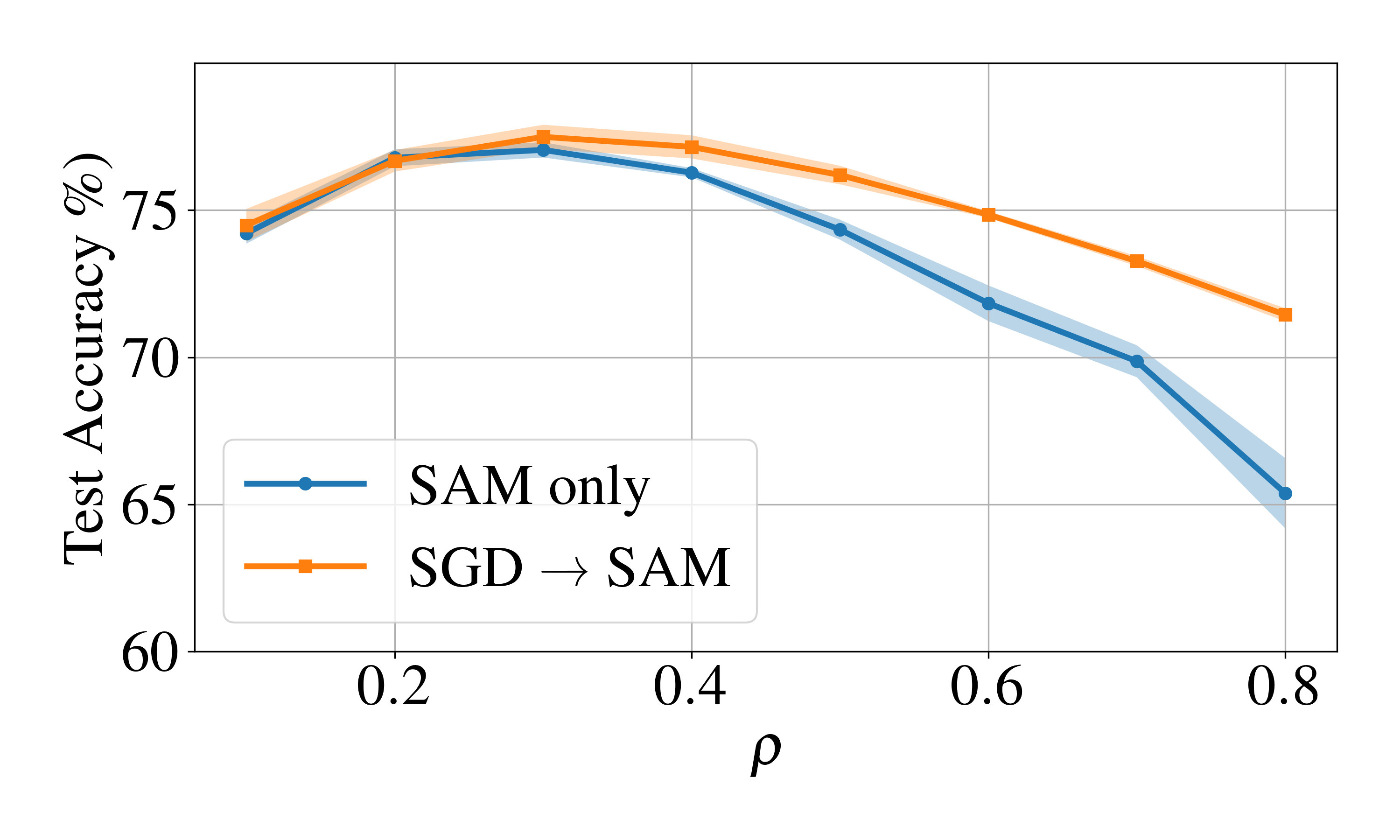}
    \caption{
    Final test accuracy for SAM-only and the switching strategy on CIFAR-100 with ResNet-18 using stochastic gradients. Each curve shows the mean (bold) and standard deviation (shaded area) over 5 seeds, evaluated at perturbation radii $\rho = 0.1, 0.2, \ldots, 0.8$. Both methods achieve peak accuracy at $\rho = 0.3$, with 77.05\% for SAM-only and 77.49\% for the switching strategy.
    }
    \label{fig:sto_sam_acc}
\end{figure}

\paragraph{Larger networks: full-batch ResNet-18 on CIFAR.}
To test whether the same mismatch arises beyond small smooth networks,
we run full-batch SAM on ResNet-18 for CIFAR-10/100 across $\rho\in\{0.0,0.5,1.0,1.5,2.0\}$ (Appendix~\ref{app:full_cifar}).
Although ReLU networks fall outside our smooth/analytic assumptions, the same \emph{shifted-vs-original} mismatch signatures are visible.
Table~\ref{tab:final_metrics_resnet18} reports final-step metrics and shows that for $\rho>0$,
the SAM loss $f(x_T^{+})$ and the shifted gradient norm $\|\nabla f(x_T^{+})\|$ can be orders of magnitude smaller than their unshifted counterparts.
The corresponding trajectories (Appendix~\ref{app:full_cifar}) plateau at strictly positive $f(x_T)$ while $f(x_T^{+})$ remains small, consistent with the hallucinated-minimizer mechanism.

\paragraph{Stochastic training on CIFAR-100.}
Finally, we examine the practical behavior under standard mini-batch training for CIFAR-100 with ResNet-18.
Figure~\ref{fig:sto_sam_acc} shows that SAM-only training becomes increasingly unstable as $\rho$ grows, while a short SGD warm-start (next subsection) improves stability and reduces sensitivity to $\rho$.
Since our theory is deterministic/full-batch, 
we make no formal claims in the stochastic regime;
we report these results as an empirical \emph{observation} that the large-$\rho$ degradation can also arise in practice.
Additional details are provided in Appendix~\ref{app:sto_SAM}.

\subsection{A short SGD warm-start as a practical safeguard}
\label{sub:exp3}

A trivial way to reduce the risk of hallucinated minimizers is to keep $\rho$ very small, but this can weaken the sharpness-regularization effect that motivates SAM.
Instead, we consider a minimal and widely compatible \emph{SGD warm-start} schedule:
we run the \emph{base optimizer} (SGD) for a short initial phase,
and then enable SAM for the remainder of training.
We use a fixed warm-start fraction $10\%$ in all experiments and do not tune it.
Warm-start schedules and related heuristics have been considered in prior work (e.g., \citet{andriushchenko2022towards});
our focus here is not to propose a new training recipe, but to show that a short warm-start is sufficient to mitigate the hallucinated-minimizer failure mode identified in Sections~\ref{sec:2}--\ref{sec:3}.

\paragraph{Empirical effect.}
In the full-batch MNIST setting, Figure~\ref{fig:det_sam} shows that SAM-only training frequently plateaus at positive loss for larger radii,
whereas the SGD warm-start reliably avoids these plateaus and drives training loss to (near-)zero across the same $\rho$ sweep.
In the stochastic CIFAR-100 setting, Figure~\ref{fig:sto_sam_acc} similarly shows improved stability and reduced sensitivity to $\rho$.

\paragraph{Connection to our theory.}
The existence proof in Theorem~\ref{main:thm1} constructs hallucinated minimizers via geometry near local maximizer regions,
and Theorem~\ref{thm:attractor} shows that such points can have basins of attraction under the practical SAM dynamics.
Early in training, iterates typically lie in high-loss regions, where encountering these basins is more likely.
A brief SGD warm-start moves the trajectory away from high-loss regions before the SAM perturbation is enabled,
reducing the chance of entering a hallucinated-minimizer basin without requiring $\rho$ to be small.

\section{Conclusion}
We identified a previously under-emphasized mismatch in standard SAM: 
the practical update uses the shifted gradient $\nabla f(x^{+})$, rather than the surrogate gradient $\nabla f^{\mathrm{SAM}}(x)$.
This mismatch enables a distinct large-$\rho$ failure mode in nonconvex landscapes:
SAM can stall at \emph{hallucinated minimizers}---local minimizers of $f^{\mathrm{SAM}}$ that are not stationary points of the original objective.
We established geometric existence mechanisms (including extensions to local maximizer sets), 
showed that isolated hallucinated-minimizer components can be locally attracting under the discrete-time SAM iteration, and characterized how hallucinated minimizers can inherit manifold structure via local invertibility of the perturbation map.
Empirically, our diagnostics on neural network objectives exhibit mismatch signatures consistent with this mechanism, and we report that a short initial SGD warm-start can serve as a simple safeguard that improves stability and reduces sensitivity to $\rho$.

An important direction for future work is to develop analogous guarantees for stochastic mini-batch training as well as for nonsmooth architectures. 
It would also be valuable to analyze whether and how SAM variants that modify the ascent step suppress or amplify hallucinated minimizers.



\appendix

\section{Omitted details for Section~\ref{sec:2}}\label{appendix2}

We begin by showing that the hallucinated minimizers cannot exist when the function is convex.

\begin{proposition}\label{thm:hmcv}
Let \(f:\mathbb{R}^d\to\mathbb{R}\) be convex and differentiable, and fix \(\rho>0\).
Then for any \(x\in\mathbb{R}^d\) with \(\nabla f(x)\neq 0\), we have \(\nabla f(x^+)\neq 0\),
where \(x^+:=x+\rho \frac{\nabla f(x)}{\|\nabla f(x)\|}\).
\end{proposition}

\begin{proof}
Suppose \( \nabla f(x) \ne 0 \) and \( \nabla f(x^+) = 0 \). Since \( f \) is convex, \( x^+ \) must be a global minimizer. However, convexity also implies
\begin{equation*}
f(x^+) \ge f(x) + \left\langle \nabla f(x), \rho\frac{\nabla f(x)}{\|\nabla f(x)\|} \right\rangle = f(x) + \rho \| \nabla f(x) \| > f(x),
\end{equation*}
which contradicts the optimality of \( x^+ \).
\end{proof}

\subsection{Full proof of Theorem~\ref{main:thm1}} \label{app:thm1pf}

We now provide the full proof of the existence theorem.

\main*

\begin{proof}
Let $x^\bullet$ be an isolated strict local maximizer with $f(x^\bullet)=M$.
Choose a closed ball $C\subseteq U$ centered at $x^\bullet$ such that
$f(x) < f(x^\bullet)$ for all $x\in C\setminus\{x^\bullet\}$,
and $x^\bullet$ is the only critical point of $f$ in $C$. 
Define $m:=\max_{\partial C} f(x)<M$ and consider the preimage
$f^{-1}([M-\varepsilon,M])$ where $0<\varepsilon<M-m$.  
Let $C_\varepsilon$ denote the connected component of this preimage containing $x^\bullet\in f^{-1}([M-\varepsilon,M])$.

By Lemma~\ref{lem:bounded},
 $C_\varepsilon \subseteq \mathrm{int}\,\,C$, and hence $C_\varepsilon$ is compact. 
 Moreover, by Lemma~\ref{lem:level}, the function value on the boundary satisfies $f(x) = M - \varepsilon$
for all 
 $x \in \partial C_\varepsilon$.

Consider the squared distance function $g(x)=\|x-x_\star\|^2$, and let
\[
x_h\in\mathrm{argmax}_{ C_\varepsilon}g(x), \quad x_h \ne x_\star, \quad \textup{ and} \quad \rho:=\|x_h-x_\star\|.
\]
Then, $x_h$ must be on the boundary of $C_\varepsilon$, and thus  $f(x_h)=M-\varepsilon$. 
Furthermore, since $x^\bullet$ is the only critical point of $f$ in $C_\varepsilon$ and $x_h\neq x^\bullet$,
we have $\nabla f(x_h)\neq 0$.
Consequently, there exists an open neighborhood $V$ of $x_h$ such that $\Sigma:=\{x \in V : f(x) = M - \varepsilon\}$ is an embedded $C^1$ hypersurface near $x_h$. By shrinking $V$ if necessary, we may assume $V\cap\partial C_\varepsilon = \Sigma$. 
Thus, maximizing 
$g(x)$ over $C_\varepsilon$ is locally equivalent to maximizing 
$g(x)$ over the hypersurface $\Sigma=\{x\in V:\ f(x)=M-\varepsilon\}$ near $x_h$.
Since $\nabla f(x_h)\neq 0$, $\Sigma$ is a regular level set and hence an embedded $C^1$ hypersurface.
Because $x_h$ maximizes $g$ over $C_\varepsilon$ and lies on $\partial C_\varepsilon$, it is also a local maximizer of $g$ over $\Sigma$.
By the method of Lagrange multipliers, there exists $\lambda\in\mathbb{R}$ such that
\[
\nabla g(x_h) + \lambda \nabla f(x_h) = 0.
\]
Moreover, $\lambda > 0$ by Lemma~\ref{lem:positive}.
It follows that
\[
2(x_\star-x_h)=\lambda \nabla f(x_h).
\]
Taking norms of both sides yields $\lambda=\frac{2\rho}{\|\nabla f(x_h)\|}$. Therefore,
\[
x_\star = x_h + \rho \frac{\nabla f(x_h)}{\|\nabla f(x_h)\|},
\]
which implies that $x_h$ is a hallucinated minimizer. This shows that for each $\varepsilon\in(0,M-m)$ there is a corresponding radius $\rho>0$ such that $x_h$ is a hallucinated minimizer.
Since $\rho(\varepsilon)$ is strictly increasing and continuous in $\varepsilon$ by Lemma~\ref{lem:rho-mono}, there exists an interval of radii $\rho$ as claimed.
\end{proof}

We now prove the three lemmas used in the proof of Theorem~\ref{main:thm1}.

\begin{lemma}\label{lem:bounded}
The set $C_\varepsilon$ from Theorem~\ref{main:thm1} is contained in $\mathrm{int}\,C$. Hence, it is compact.  
\end{lemma}
\begin{proof}
Suppose $x\in C_\varepsilon$. If $x \in \partial C$, then $f(x)\ge M-\varepsilon>m$, contradicting the definition of $m$ as the maximum value of $f$ on $\partial C$. 

If instead $x\in \mathrm{ext}\,C=\mathbb{R}^d\setminus C$, then $\mathrm{int}\,C$ and $\mathbb{R}^d\setminus C$ are two  nonempty disjoint open sets that separate $C_\varepsilon$, contradicting the fact that $C_\varepsilon$ is connected. Therefore, $x$ must lie in $\mathrm{int}\,C$, and $C_\varepsilon \subseteq \mathrm{int}\,C$. 
\end{proof}

The following lemma shows that every point on $\partial C_\varepsilon$ in Theorem~\ref{main:thm1} lies on the same level set. 

\begin{lemma}\label{lem:level}
Any point $x \in \partial C_\varepsilon$ from Theorem~\ref{main:thm1} satisfies $f(x)=M-\varepsilon$. 
\end{lemma}
\begin{proof}
   Take $x\in\partial C_\varepsilon$. 
If $f(x)>M-\varepsilon$, continuity of $f$ implies that there exists $r>0$ such that $f(y)>M-\varepsilon$ for all $y$ in an open ball $B_r(x)$ centered at $x$ with radius $r$. 
Since $B_r(x)\subset f^{-1}([M-\varepsilon,M])$ is connected and intersects $C_\varepsilon$,
maximality of the connected component implies $B_r(x)\subset C_\varepsilon$,
so $x$ would be an interior point.
This contradicts the fact that 
$x$ is a boundary point of $C_\varepsilon$. 

If $f(x)<M-\varepsilon$, this directly contradicts the fact that $x \in f^{-1}([M-\varepsilon, M])$. Therefore, $f(x)=M-\varepsilon$.
\end{proof}

The next lemma establishes that  $\lambda>0$ in Theorem~\ref{main:thm1}.

\begin{lemma}\label{lem:positive}
    In the proof of Theorem~\ref{main:thm1}, the vectors $x_\star-x_h$ and $\nabla f(x_h)$ point in the same direction. Equivalently, $\lambda>0$.
\end{lemma}
\begin{proof}
    Let $V$ be an open neighborhood of $x_h$. By possibly shrinking $V$, we may assume the local superlevel set $\{x\in V : f(x)\ge M-\varepsilon\}$ is contained in $C_\varepsilon$.  Then, $x_h$ is a maximizer of $g$ over the feasible region $\{x\in V : f(x)\ge M-\varepsilon\}$. 
    
    Take the direction $d:=\nabla f(x_h)$, for which $\langle \nabla f(x_h),d\rangle=\|\nabla f(x_h)\|^2>0$.
    Let $\gamma\colon(-\varepsilon,\varepsilon)\to\mathbb{R}^d$ be a smooth curve with $\gamma(0)=x_h$ and $\gamma'(0)=d$.
Then, $f(\gamma(t))\ge M-\varepsilon$ for sufficiently small  $t>0$. Moreover,
    \[
    \frac{d}{dt}f(\gamma(t))\Bigg|_{t=0}=\langle \nabla f(x_h), d\rangle  \ge 0,
    \]
    so $\gamma(t)$ remains in the feasible set for small $t>0$. 
    Since $x_h$ maximizes $g$, it follows that
    \[
    \lim_{t\downarrow 0}\frac{d}{dt}g(\gamma(t))= \lim_{t\downarrow 0}\frac{d}{dt}\|\gamma(t)-x_\star\|^2=\lim_{t\downarrow 0}2\langle \gamma(t)-x_\star , \gamma'(t) \rangle =2\langle x_h-x_\star, d\rangle \le 0.
    \]
    Since $d=\nabla f(x_h)$,
    \[
    2\langle x_h-x_\star, \nabla f(x_h)\rangle = -\lambda \|\nabla f(x_h)\|^2 \le 0. 
    \]
    Since $\lambda \ne 0$, 
    this inequality implies
 $\lambda > 0$.
\end{proof}

\begin{lemma}\label{lem:rho-mono}
In the proof of Theorem~\ref{main:thm1}, define
\[
\rho(\varepsilon):=\max_{x\in C_\varepsilon}\|x-x_\star\|.
\]
Then $\rho(\varepsilon)$ is continuous and strictly increasing for $0<\varepsilon<M-m$.
\end{lemma}

\begin{proof}
Since $C_\varepsilon$ is compact (Lemma~\ref{lem:bounded}), the maximum defining $\rho(\varepsilon)$ is attained.
Moreover, $\|x-x_\star\|$ cannot attain its maximum in $\mathrm{int}(C_\varepsilon)$.
Therefore
\[
\rho(\varepsilon)=\max_{x\in \partial C_\varepsilon}\|x-x_\star\|.
\]

\paragraph{Strict monotonicity.}
Fix $0<\varepsilon_1<\varepsilon_2<M-m$.
By construction of $C_\varepsilon$ in the proof of Theorem~\ref{main:thm1},
$C_{\varepsilon_1}\subset C_{\varepsilon_2}$.
Let $x_1\in\partial C_{\varepsilon_1}$ attain $\rho(\varepsilon_1)$.
By Lemma~\ref{lem:level}, $f\equiv M-\varepsilon$ on $\partial C_\varepsilon$, so
$\partial C_{\varepsilon_1}\cap \partial C_{\varepsilon_2}=\varnothing$.
Thus $x_1\in \mathrm{int}(C_{\varepsilon_2})$, and there exists $\delta>0$ such that
$B(x_1,\delta)\subset C_{\varepsilon_2}$.
Let $v:=(x_1-x_\star)/\|x_1-x_\star\|$ and choose any $t\in(0,\delta)$.
Then $x_1+t v\in C_{\varepsilon_2}$ and
\[
\rho(\varepsilon_2)\ge \|x_1+t v-x_\star\|
\ge\|x_1-x_\star\|+t
>\|x_1-x_\star\|
=\rho(\varepsilon_1).
\]
Thus $\rho(\varepsilon)$ is strictly increasing on $(0,M-m)$.

\paragraph{Continuity.}
Fix $\varepsilon_0\in(0,M-m)$.
By construction, $x^\bullet\in \mathrm{int}(C_{\varepsilon_0})$, and $x^\bullet$ is the only critical point of $f$ in $C$.
Hence $\nabla f(x)\neq 0$ for all $x\in \partial C_{\varepsilon_0}$.
In particular, the level value $M-\varepsilon_0$ is regular on $\partial C_{\varepsilon_0}$,
so
$\partial C_{\varepsilon}$ forms a $C^1$ hypersurface that varies continuously with $\varepsilon$ in a neighborhood of $\varepsilon_0$
(by the implicit function theorem, applied locally and combined via a finite cover of the compact set $\partial C_{\varepsilon_0}$).

Since $C_{\varepsilon}$ is compact for all $\varepsilon\in(0,M-m)$ and
\[
\rho(\varepsilon)=\max_{x\in C_{\varepsilon}}\|x-x_\star\|
=\max_{x\in \partial C_{\varepsilon}}\|x-x_\star\|,
\]
standard maximum-continuity arguments for continuous functions over continuously varying compact sets yield that
$\rho(\varepsilon)$ is continuous at $\varepsilon_0$.
As $\varepsilon_0$ was arbitrary, $\rho$ is continuous on $(0,M-m)$.
\end{proof}

\subsection{Extending Theorem~\ref{main:thm1} to local minimizers} \label{app:thm1ext}

We now extend Theorem~\ref{main:thm1} by relaxing  the  assumption that $x_\star$ is a global minimizer. In fact, a hallucinated minimizer can still exist when $x_\star$ is only a local minimizer. To show this, we first establish the following lemma.

\begin{lemma}\label{lem:local}
    Suppose $\|\nabla f(x)-\nabla f(y)\|\le L \|x-y\|$ for some $L>0$, and that $\nabla f(x)\ne 0$ and $\nabla f(y)\ne 0$. Then, we have 
    \[
    \|y + \rho \,u(y) - x - \, \rho \,u(x)\| \leq \Big(1 + \frac{2\rho L}{\|\nabla f(x)\|}\Big)\|y - x\|.
    \]
\end{lemma}
\begin{proof}
It follows from the triangle inequality that
\begin{equation*}
\begin{split}
    &\|y + \rho \, u(y) - x - \rho \, u(x)\| \\
    &\leq \|y - x\| + \rho \Big\|\frac{\nabla f(y)}{\|\nabla f(y)\|} - \frac{\nabla f(x)}{\|\nabla f(x)\|}\Big\| \\
    &\leq \|y - x\| + \rho \Big\|\frac{\nabla f(y)}{\|\nabla f(y)\|} - \frac{\nabla f(y)}{\|\nabla f(x)\|}\Big\| + \rho \Big\|\frac{\nabla f(y)}{\|\nabla f(x)\|} - \frac{\nabla f(x)}{\|\nabla f(x)\|}\Big\| \\
    &\leq \|y - x\| + \rho \frac{\|\nabla f(x) - \nabla f(y)\|}{\|\nabla f(x)\|} + \rho \frac{\|\nabla f(y) - \nabla f(x)\|}{\|\nabla f(x)\|} \\
    &\leq \|y - x\| \Big(1 + \frac{2\rho L}{\|\nabla f(x)\|}\Big).
\end{split}
\end{equation*}
\end{proof}

Finally, we obtain the following corollary. 

\begin{corollary}\label{cor:local}
Suppose $f$ has a locally Lipschitz gradient.
Then Theorem~\ref{main:thm1} remains valid even when $x_\star$ is a local minimizer of $f$.
\end{corollary}
\begin{proof}
    Recall the last equation of the proof of Theorem~\ref{main:thm1}:
    \[
    x_\star = x_h + \rho \frac{\nabla f(x_h)}{\|\nabla f(x_h)\|}.
    \]
Consider an open ball centered at $x_h$ with radius $r$ chosen sufficiently small so that $\nabla f$ does not vanish on the ball. Let $L>0$ be such that $\| \nabla f( x_h)- \nabla f(y)\| \le L\|x_h - y\|$ for any $y\in B_r(x_h)$. Since $x_\star$ is a local minimizer, there exists $\delta >0$ such that
\[
f(y) \ge f(x_\star) \quad \forall y \mbox{ with } \|y-x_\star\| \le \delta.
\]
 Now consider an open ball centered at $x_h$ with radius 
\[
r^\star := \min\left\{ \frac{\delta}{1+\frac{2\rho L}{\|\nabla f(x_h)\|}}, r \right\}.
\]
Then, for any $y\in B_{r^\star}(x_h)$, we have
\[
\|x_\star-y-\rho \, u(y)\|=\|x_h + \rho \, u(x_h) -  y - \rho \, u(y)\|\le \|x_h-y\|\left(1+\frac{2\rho L}{\|\nabla f(x_h)\|}\right)\le \delta,
\]
where the first inequality follows from Lemma~\ref{lem:local}. Hence, 
\[
f(y+\rho \, u(y))\ge f(x_\star).
\]
This implies $f^{\mathrm{SAM}}(y)\ge f^{\mathrm{SAM}}(x_h)$, so $x_h$ is a local minimizer of $f^{\mathrm{SAM}}$, and therefore a hallucinated minimizer.
\end{proof}

\subsection{Full proof of Theorem~\ref{thm:exist}}\label{app:thm2pf}

\begin{lemma}[\cite{lojasiewicz1963propriete}]\label{lem:Lineq}
If $f\colon\mathbb{R}^d\to\mathbb{R}$ is real-analytic, then for every $p\in \mathbb{R}^d$, there exist an open neighborhood $U$ of $p$, a constant $c>0$, and an exponent $q\in(0,1)$ such that
\[
|f(p)-f(x)|^{q} \le c\|\nabla f(x)\| \quad \text{for all } x\in U.
\]
\end{lemma}

The following lemma shows that, 
under the real-analyticity assumption,
critical points cannot accumulate around the local maximizer set $X$. The argument relies on the {\L}ojasiewicz inequality.

\begin{lemma}\label{lem:nocritical}
   Suppose $f\colon\mathbb{R}^d\to\mathbb{R}$ is real-analytic and $X$ is a bounded local maximizer set of $f$ for some $\delta>0$. Then, there exists $\varepsilon>0$ with the following property: if $x$ is a critical point in the $\delta$-neighborhood of $X$ with $f(x)\ge f(X)-\varepsilon$, then $x\in X$.
\end{lemma}

\begin{proof}
     Define the closed $\delta$-neighborhood of $X$ by $\mathcal{N}_\delta(X):=\{y: d(y,X)\le \delta\}$.
     Since $X$ is a bounded connected set, $\mathcal{N}_\delta(X)$ 
     is compact and connected. 
     Let $S$ denote the set of critical points in $\mathcal{N}_\delta(X)$ that are not in $X$:
    \[
    S=\{ s\in \mathcal{N}_\delta(X): \nabla f(s)=0 \} \setminus X.
    \]
    If $S =  \emptyset$, then the theorem holds for any $\varepsilon>0$, and we are done. Assume instead that $S \neq \emptyset$.
    For each point $x\in X$,  Lemma~\ref{lem:Lineq} guarantees  the existence of an open neighborhood $U_x$, a constant $C_x>0$, and an exponent $q_x\in(0,1)$ such that 
    \[
     |f(x)-f(y)|^{q_x}= |f(X)-f(y)|^{q_x} \le C_x\|\nabla f(y)\|, \, \quad \, y\in U_x.
    \]
    If $y\in S$, then 
    \[
    |f(X)-f(y)|^{q_x} \le C_x\|\nabla f(y)\|=0.
    \]
    This implies $f(y)=f(X)$, which contradicts $X=\argmax_{z\in \mathcal{N}_\delta(X)} f(z)$ because $y\notin X$. 
    Hence,  $S \subseteq C:= \mathcal{N}_\delta(X) \setminus \bigcup_{x\in X} U_x$. (Note that $C \neq \emptyset$ since $S\neq \emptyset$). 
    Since $C$ is compact, define 
    \[
0<\varepsilon^\star:=f(X)-\max_{y\in C}f(y).
    \]
Then, any $\varepsilon\in (0,\varepsilon^\star)$ satisfies the theorem.
\end{proof}

Finally, we use  Lemma~\ref{lem:nocritical} to complete the proof of Theorem~\ref{thm:exist}.

\maintwo*

\begin{proof}
 Define the closed $\delta$-neighborhood of a local maximizer set $X$ as
 $\mathcal{N}_\delta(X):=\{y: d(y,X)\le \delta\}$.
 Since $X$ 
  is a bounded connected set,
  $\mathcal{N}_\delta(X)$ is
  compact and connected. By Lemma~\ref{lem:nocritical}, there exists $\varepsilon_1>0$ such that any critical point in $\mathcal{N}_\delta(X)$ with function value at least $f(X)-\varepsilon_1$ must lie in $X$.
 Next, choose $\varepsilon_2>0$ such that $0<\varepsilon_2<f(X)-\max_{\partial \mathcal{N}_\delta(X)}f(x)$. 
 
 Let $\varepsilon:=\min\{\varepsilon_1, \varepsilon_2\}$, and consider the preimage
$f^{-1}([f(X)-\varepsilon,f(X)])$.  Define $C_\varepsilon$ as the connected component of this preimage that contains $X$.

By the same reasoning as in Lemma~\ref{lem:bounded}, $C_\varepsilon \subseteq \mathrm{int}\,\, \mathcal{N}_\delta(X)$, and hence $C_\varepsilon$ is compact. Moreover, by Lemma~\ref{lem:level}, every point $x \in \partial C_\varepsilon$  satisfies $f(x)=f(X)-\varepsilon$.

Now define $g(x)=\|x-x_\star\|^2$ and let
\[
x_h\in\mathrm{argmax}_{ C_\varepsilon}g(x), \quad x_h\ne x_\star, \quad  \rho:=\|x_h-x_\star\|.
\]
Then $x_h\in\partial C_\varepsilon$ (otherwise we could move slightly in the direction $x_h-x_\star$ while staying in $C_\varepsilon$ and increase $g$).
Hence $f(x_h)=f(X)-\varepsilon<f(X)$, so $x_h\notin X$, and Lemma~\ref{lem:nocritical} implies $\nabla f(x_h)\neq 0$.
Thus, there exists an open neighborhood $V$ of $x_h$ such that $\Sigma:=\{x \in V : f(x) = f(X) - \varepsilon\}$ is an embedded smooth hypersurface near $x_h$. By possibly shrinking $V$, we may assume $V\cap\partial C_\varepsilon = \Sigma$. Therefore, maximizing 
$g(x)$ over $ C_\varepsilon$ is locally equivalent to maximizing 
$g(x)$ over the hypersurface $\Sigma$. 

Then, by the method of Lagrange multipliers, we obtain
\[
\nabla g(x_h) + \lambda \nabla f(x_h) = 0.
\]
Equivalently,
\[
2(x_\star-x_h)=\lambda \nabla f(x_h).
\]
Moreover, $\lambda>0$ follows from the same argument as in Lemma~\ref{lem:positive}.
Taking norms of both sides yields $\lambda=\frac{2\rho}{\|\nabla f(x_h)\|}$. Therefore,
\[
x_\star = x_h + \rho \frac{\nabla f(x_h)}{\|\nabla f(x_h)\|},
\]
which shows that $x_h$ is a hallucinated minimizer. Moreover, just like in the proof of Theorem~\ref{main:thm1}, the above argument  applies for all
\[
0<\varepsilon<\min\{\varepsilon_1,\varepsilon_2\}.
\]
Hence, by Lemma~\ref{lem:rho-mono-set} below, we conclude that the set of radii for which a hallucinated minimizer exists contains a nontrivial interval.
\end{proof}

\begin{lemma}[Monotonicity/continuity of $\rho(\varepsilon)$ for maximizer-set construction]
\label{lem:rho-mono-set}
With $C_\varepsilon$ defined as the connected component of $f^{-1}([f(X)-\varepsilon,f(X)])$ containing $X$,
the map $\rho(\varepsilon):=\max_{x\in C_\varepsilon}\|x-x_\star\|$ is continuous and strictly increasing
for $\varepsilon\in(0,\min\{\varepsilon_1,\varepsilon_2\})$.
\end{lemma}

\begin{proof}
The proof is identical to Lemma~\ref{lem:rho-mono}
and uses only compactness, nestedness
$C_{\varepsilon_1}\subset C_{\varepsilon_2}$ for $\varepsilon_1<\varepsilon_2$, and connectedness.    
\end{proof}

\subsection{When neural networks are real-analytic: sufficient conditions}
\label{app:NN}

Theorem~\ref{thm:exist} assumes that the objective $f$ is real-analytic.
This assumption is used only to invoke the {\L}ojasiewicz inequality (Lemma~\ref{lem:Lineq})
and thereby rule out pathological accumulations of critical points near a local maximizer set.
Here we record a simple sufficient condition showing that real-analyticity holds for a  class of
\emph{smooth} (deterministic) neural network objectives.
In particular, it covers the full-batch MNIST experiment with \texttt{Tanh} activations in Section~\ref{sub:exp1}.

Throughout, we consider a finite dataset $\{(X_i,Y_i)\}_{i=1}^N$ and a deterministic network map
$\theta \mapsto h_\theta(X_i)$
(i.e., we exclude stochastic layers such as dropout from this discussion).
We also assume normalization layers use the standard numerical constant $\varepsilon_{\mathrm{norm}}>0$
to avoid division by zero.

\begin{lemma}[Real-analyticity of common smooth building blocks]
\label{lem:analytic-blocks}
Fix $\varepsilon_{\mathrm{norm}}>0$.
The following mappings are real-analytic on their natural domains:
\begin{itemize}
\item affine/linear maps and convolutions;

\item residual additions and average pooling;

\item softmax and log-softmax;

\item  layer normalization and batch normalization in the standard form
\[
\mathrm{Norm}(z)=\gamma \odot \frac{z-\mu(z)}{\sqrt{\sigma(z)+\varepsilon_{\mathrm{norm}}}}+\beta,
\]
where $\mu(z)$ and $\sigma(z)$ are (per-sample or per-batch) mean/variance functions of $z$.

\end{itemize}
Moreover, any elementwise real-analytic activation $\sigma(\cdot)$ (e.g., \texttt{Tanh}, sigmoid, softplus, GELU, swish)
is real-analytic.
\end{lemma}

\begin{proof}
Affine maps, convolutions, additions, and average pooling are polynomial/linear maps, hence real-analytic.
Softmax and log-softmax are compositions of $\exp$, finite sums, divisions by strictly positive quantities, and $\log$ applied to strictly positive quantities, hence real-analytic.

For normalization, note that $\mu(z)$ and $\sigma(z)$ are polynomials in the entries of $z$.  Since $\sigma(z)\ge 0$ and $\varepsilon_{\mathrm{norm}}>0$, the quantity $\sigma(z)+\varepsilon_{\mathrm{norm}}$ is strictly positive,
and the map $t\mapsto 1/\sqrt{t}$ is real-analytic on $(0,\infty)$.  
Thus $\mathrm{Norm}(z)$ is a composition of real-analytic maps, hence real-analytic.
Finally, elementwise real-analytic activations preserve real-analyticity under composition.
\end{proof}

\begin{lemma}
\label{lem:real-analytic}
Assume that for each fixed input $X$, the network output $\theta\mapsto h_\theta(X)$ is obtained by composing
the building blocks in Lemma~\ref{lem:analytic-blocks} with real-analytic activations.
Let the loss $\ell(\cdot,Y)$ be real-analytic in its first argument for each label $Y$
(e.g., squared loss; cross-entropy combined with softmax).
Then the finite-dataset objective
\[
f(\theta):=\frac{1}{N}\sum^N_{i=1}\ell(h_\theta(X_i), Y_i)
\]
is real-analytic in $\theta$.
\end{lemma}

\begin{proof}
By Lemma~\ref{lem:analytic-blocks} and closure of real-analytic functions under composition,
each map $\theta\mapsto h_\theta(X_i)$ is real-analytic.
Composing with the real-analytic loss $\ell(\cdot,Y_i)$ preserves real-analyticity,
and finite sums preserve real-analyticity. 
Hence $f$ is real-analytic.
\end{proof}

\section{Omitted details for Section~\ref{sec:3}}\label{appendix3}

\subsection{Proof of Theorem~\ref{thm:attractor} and further discussion}\label{app:thm4pf}

In this subsection, we prove  Theorem~\ref{thm:attractor} and then discuss the special case of isolated hallucinated minimizers.

\attractor*
\begin{proof}
 First, we claim that the set $H$ is closed (hence compact). Indeed, if $x_h\in \bar{H}$, then by continuity of $f^{\mathrm{SAM}}$ we have
 $f^{\mathrm{SAM}}(x_h)=f^{\mathrm{SAM}}(H)=\min f^{\mathrm{SAM}}$. 
By the isolation assumption on the minimizer component, this implies $x_h\in H$. Hence $H=\bar{H}$.

 Let $\mathcal{N}_\delta(H)$ denote the closed $\delta$-neighborhood of $H$ from the theorem assumption. Since
\begin{equation*}
1+\rho\lambda_{\min}(\mathrm{Sym}(\nabla u(x_h)))>0
\end{equation*}
and $\|\nabla f(x_h)\|>0$ for all $x_h \in H$, there exists an open neighborhood $W$ of $H$ such that
\begin{equation*}
1+\rho\lambda_{\min}(\mathrm{Sym}(\nabla u(x)))>0
\end{equation*}
and $\|\nabla f(x)\|>0$ for any $x\in W$.
By shrinking $\mathcal{N}_\delta(H)$  if necessary, we may assume $\mathcal{N}_\delta(H)\subseteq W$ and $f^{\mathrm{SAM}}$ is real-analytic on $\mathcal{N}_\delta(H)$.
 
 Applying an argument analogous to Lemma~\ref{lem:nocritical}, with local maximizers replaced by minimizers, we obtain $\varepsilon^\star>0$ such that if $x$
 is a critical point in $\mathcal{N}_\delta(H)$ with $f^{\mathrm{SAM}}(x)\le f^{\mathrm{SAM}}(H) +\varepsilon^\star$, then $x\in H$.

Now consider the closed neighborhood $\mathcal{N}_{\delta/2}(H)$, and set 
\[
m:=\min_{x\in \partial \mathcal{N}_{\delta/2}(H)}f^{\mathrm{SAM}}(x)>f_\star,
\]
where $f_\star=f^{\mathrm{SAM}}(H)$. The strict inequality follows from the construction of $\mathcal{N}_{\delta}(H)$. Choose $\varepsilon > 0$ such that 
\[
0<\varepsilon<m-f^{\mathrm{SAM}}(H)=m-f_\star \quad \textup { and } \quad 0<\varepsilon<\varepsilon^\star.
\]
Let $C_\varepsilon$ be
 the connected component  of the sublevel set
\[(f^{\mathrm{SAM}})^{-1}((-\infty,f_{\star}+\varepsilon])=(f^{\mathrm{SAM}})^{-1}([f_{\star},f_{\star}+\varepsilon])
\]
that 
contains $H$. Then, $C_\varepsilon$ is compact and contains no other {critical points} of $f^{\mathrm{SAM}}$ besides those in $H$. 
The proof of $C_\varepsilon$ being bounded (hence compact) by $\mathcal{N}_{\delta/2}(H)$ is analogous to Lemma~\ref{lem:bounded}. 

Define
\[
C_\rho :=\{x : d(x,C_\varepsilon)\le \rho\}.
\]
Then, $C_\rho$ is also compact. 
Let $L>0$ be any constant such that $f^{\mathrm{SAM}}$ is $L$-smooth on $\mathcal{N}_\delta(H)$, and
set
$M := \max_{x\in C_{\rho}}\|\nabla f(x)\| >0$.
Let
\[ 
\gamma := \min_{x\in \mathcal{N}_\delta(H)}
\Bigl(1+\rho\,\lambda_{\min}(\mathrm{Sym}(\nabla u(x)))\Bigr).
\]
By continuity and compactness, $\gamma>0$.
Consider the SAM update with fixed $\rho>0$ and constant step size $\eta_k=\eta$ chosen such that 
\[
0<\eta<\min \Big\{ \frac{\delta}{2M} , \ \frac{2\gamma}{L} \Big\},
\]
and initialization at $x_0 \in C_\varepsilon$.  
We show by induction that the SAM iterates $\{x_k\}$ remain in $C_{\varepsilon}$. 
The base case $x_0\in C_{\varepsilon}$ is true by assumption. 
Suppose  $x_k\in C_{\varepsilon}$. Then, by definition of the SAM update,
$x_k^+\in C_{\rho}$  and   $x_{k+1}=x_k-\eta\nabla f(x_k^{+})$.
We claim
$x_{k+1} \in \mathcal{N}_\delta(H)$, since
\begin{align*}
d(x_{k+1}, H)&=\inf_{x_h \in H}\|x_{k+1}-x_h\|\\
&\le \inf_{x_h \in H}\|x_k-x_h\|+\|x_{k+1}-x_k\|\\
&\le\frac{\delta}{2}+\eta \|\nabla f(x_k^{+})\| \\
&\le \frac{\delta}{2}+\frac{\delta}{2M}\cdot M\\
&\le \delta.
\end{align*}
By $L$–smoothness of $f^{\mathrm{SAM}}$ on $\mathcal{N}_\delta(H)$,
\begin{align*}
f^{\mathrm{SAM}}(x_{k+1})&\le f^{\mathrm{SAM}}(x_{k})+\langle \nabla f^{\mathrm{SAM}}(x_{k}),\,x_{k+1}-x_{k}\rangle + \frac{L}{2}\|x_{k+1}-x_{k}\|^2\\
&= f^{\mathrm{SAM}}(x_{k})-\eta\langle \nabla f^{\mathrm{SAM}}(x_{k}),\nabla f(x_k^{+})\rangle + \frac{L\eta^2}{2}\|\nabla f(x_k^{+})\|^2.
\end{align*}
Since
\[
\nabla f^{\mathrm{SAM}}(x) =(I + \rho\nabla u(x))^\top \nabla f(x^+),
\]
we obtain
\begin{align*}
f^{\mathrm{SAM}}(x_{k+1})
&\le f^{\mathrm{SAM}}(x_{k})-\eta\langle \nabla f^{\mathrm{SAM}}(x_{k}),\nabla f(x_k^{+})\rangle + \frac{L\eta^2}{2}\|\nabla f(x_k^{+})\|^2\\
&= f^{\mathrm{SAM}}(x_{k})-\eta\rho\langle \nabla u(x_k)\nabla f(x_k^{+}),\nabla f(x_k^{+})\rangle - \eta\|\nabla f(x_k^{+})\|^2+\frac{L\eta^2}{2}\|\nabla f(x_k^{+})\|^2\\
&\le f^{\mathrm{SAM}}(x_{k})-\eta\rho\lambda_{\min}(\mathrm{Sym}(\nabla u(x_k)))\|\nabla f(x_k^{+})\|^2  - \eta\|\nabla f(x_k^{+})\|^2+\frac{L\eta^2}{2}\|\nabla f(x_k^{+})\|^2\\
&\le f^{\mathrm{SAM}}(x_{k})-\eta\left(1+\rho\lambda_{\min}(\mathrm{Sym}(\nabla u(x_k)))-\frac{L\eta}{2}\right)\|\nabla f(x_k^{+})\|^2\\
&\le f^{\mathrm{SAM}}(x_{k})-\eta\left(\gamma-\frac{L\eta}{2}\right)\|\nabla f(x_k^{+})\|^2\\
&< f^{\mathrm{SAM}}(x_{k}),
\end{align*}
where the last inequality follows from $0<\eta < \frac{2\gamma}{L}$. Moreover, since the descent property $f^{\mathrm{SAM}}(x_{k+1}) <f^{\mathrm{SAM}}(x_{k})$ holds when $\eta$ is replaced by $\eta t$ for $t\in[0,1]$, the line segment from $x_k$ to $x_{k+1}$ lies within the sublevel set $(f^{\mathrm{SAM}})^{-1}([f_{\star},f_{\star}+\varepsilon])$. 
Thus, because $x_k\in C_\varepsilon$, we conclude $x_{k+1} \in C_{\varepsilon}$ by the connectedness of $C_\varepsilon$. This completes the induction.

Finally, since $f^{\mathrm{SAM}}(x_k)$ is decreasing and bounded  below, we have $\eta\left(\gamma-\frac{L\eta}{2}\right)\|\nabla f(x_k^{+})\|^2 \rightarrow 0$. Hence,  $\nabla f(x_k^{+})\to 0$. If $x_{\infty}$ is a limit point of $\{x_k\}_{k=0,1,\dots}$, then $\nabla f(x_\infty^{+})=0$. By  construction of $C_\varepsilon$, this implies $x_\infty\in H$. Therefore, $d(x_k,H)\to 0$.
\end{proof}

To discuss point convergence to isolated hallucinated minimizers, 
we now turn to the case where the manifold $\mathcal{M}$ in Theorem~\ref{thm:manifold} reduces to a single isolated point, i.e.,  $0$-dimensional.
In this case, we can show that the corresponding hallucinated minimizer is also isolated. 
This can be viewed as the special case where $H$ in Theorem~\ref{thm:attractor} is a singleton.

\begin{lemma}\label{lem:existisolated}
Let $x_\star$ be a minimizer of a $C^1$ function $f:\mathbb{R}^d\rightarrow \mathbb{R}$ satisfying the assumptions of Theorem~\ref{main:thm1}. Suppose $x_\star$ is an isolated minimizer; that is, there exists an open neighborhood of $x_\star$ in which it is the unique critical point and the unique minimizer. 
Let $x_h$ be a hallucinated minimizer of $f$ constructed in the proof of Theorem~\ref{main:thm1} with $\rho>0$. If $I+\rho\nabla u(x_h)$ is invertible, then there exists an open neighborhood $W$ of $x_h$ such that
\begin{itemize}
    \item $\nabla f$ never vanishes on $W$, and thus $f^{\mathrm{SAM}}$ is well-defined on $W$;
    \item no  point other than $x_h$ satisfies $\nabla f(x^+)=0$ in $W$; and
    \item  $x_h$ is the unique hallucinated minimizer in $W$. 
\end{itemize}
\end{lemma}
Such a point $x_h$ is called an \emph{isolated hallucinated minimizer}. 

\begin{proof}
Since $x_h$ is a hallucinated minimizer of $f$ constructed in the proof of Theorem~\ref{main:thm1}, it satisfies
    \[
    x_\star = x_h + \rho \frac{\nabla f(x_h)}{\|\nabla f(x_h)\|}=x_h+\rho u(x_h)
    \]
     for some perturbation radius $\rho>0$. Let $U_0$ and $V_0$ be  open neighborhoods of $x_\star$ and $x_h$, respectively, as constructed in the proof of Theorem~\ref{thm:manifold}. Also, let $U_1$ be an open neighborhood of $x_\star$ that contains no other minimizers, by  assumption. Define $W:=G(U_0 \cap U_1)$, where $G$ is the $C^1$ mapping constructed in the proof of Theorem~\ref{thm:manifold}. We claim that $W$ is the desired open neighborhood of $x_h$.

     First, $W$ is open since $G$ is a local diffeomorphism at $x_\star$. Moreover, $f^{\mathrm{SAM}}$ is well-defined on $W$ by the construction of $V_0$. Suppose $y\in W$ satisfies $\nabla f(y^{+})=0$. Then, since 
     \[
      x = y + \rho \frac{\nabla f(y)}{\|\nabla f(y)\|}
     \]
     for the unique $x=G^{-1}(y)\in U_1$, it follows that $\nabla f(y^{+})=\nabla f(x)=0$,   contradicting the fact that $x$ is an isolated minimizer. 
     Similarly, if $y$ is a  hallucinated minimizer, then the uniqueness of $x=G^{-1}(y)$ implies $x=x_\star$, and hence $y=x_h$. This proves the claim.
\end{proof}

Then, as a corollary of Theorem~\ref{thm:attractor} and the previous lemma, we obtain the following point convergence result.

\begin{corollary}
    Assume $f$ satisfies the assumptions in Theorem~\ref{main:thm1} and, in addition, $f\in C^2$. Let $x_h$ be an isolated hallucinated minimizer of $f$ for a fixed perturbation radius $\rho>0$, constructed in the proof of Theorem~\ref{main:thm1}. Suppose 
\[
1+\rho\lambda_{\min}(\mathrm{Sym}(\nabla u(x_h)))>0.
\]
Then, the SAM iterates, when initialized sufficiently close to $x_h$, converge to $x_h$ for  sufficiently small fixed step size $\eta_k=\eta$. 
\end{corollary}

The real-analytic property of $f$ is used to apply the {\L}ojasiewicz inequality in order to construct a neighborhood where critical points do not accumulate. However, since Lemma~\ref{lem:existisolated} guarantees the existence of isolated hallucinated minimizers without this assumption, the real-analytic condition is not required here. Hence, the $C^2$ assumption on $f$ is sufficient to ensure the existence of constants $L$ and $\gamma$ as in the proof of Theorem~\ref{thm:attractor}.

\subsection{Proof of Theorem \ref{thm:manifold}} \label{app:thm3pf}

\manifold*
\begin{proof}
Fix $\rho>0$ and define $\Phi_\rho(x):=x+\rho u(x)$ on the open set
\[
\Omega:=\{x\in\mathbb{R}^d:\ \nabla f(x)\neq 0\}.
\]
Since $f\in C^2$, the map $u(x)=\nabla f(x)/\|\nabla f(x)\|$ is $C^1$ on $\Omega$,
and hence $\Phi_\rho$ is $C^1$ on $\Omega$ with Jacobian
\[
D\Phi_\rho(x)=I+\rho \nabla u(x).
\]

By assumption, $x_h\in\Omega$ and $D\Phi_\rho(x_h)=I+\rho\nabla u(x_h)$ is nonsingular.
Therefore, by the inverse function theorem,
there exist neighborhoods
$V$ of $x_h$ and $U$ of $x_\star=\Phi_\rho(x_h)$ such that
\[
\Phi_\rho:V\to U
\]
is a $C^1$ diffeomorphism with inverse $G:=\Phi_\rho^{-1}:U\to V$.

Now set
\[
\mathcal{H}:=G(\mathcal{M}\cap U)=\{x\in V:\ \Phi_\rho(x)\in\mathcal{M}\cap U\}.
\]
Since $\mathcal{M}\cap U$ is an embedded $m$-dimensional manifold and $G$ is a diffeomorphism,
$\mathcal{H}$ is an embedded $m$-dimensional manifold as well.

\begin{figure}[t]
    \centering
    \begin{subfigure}{0.38\linewidth}
        \centering
        \includegraphics[width=\linewidth]{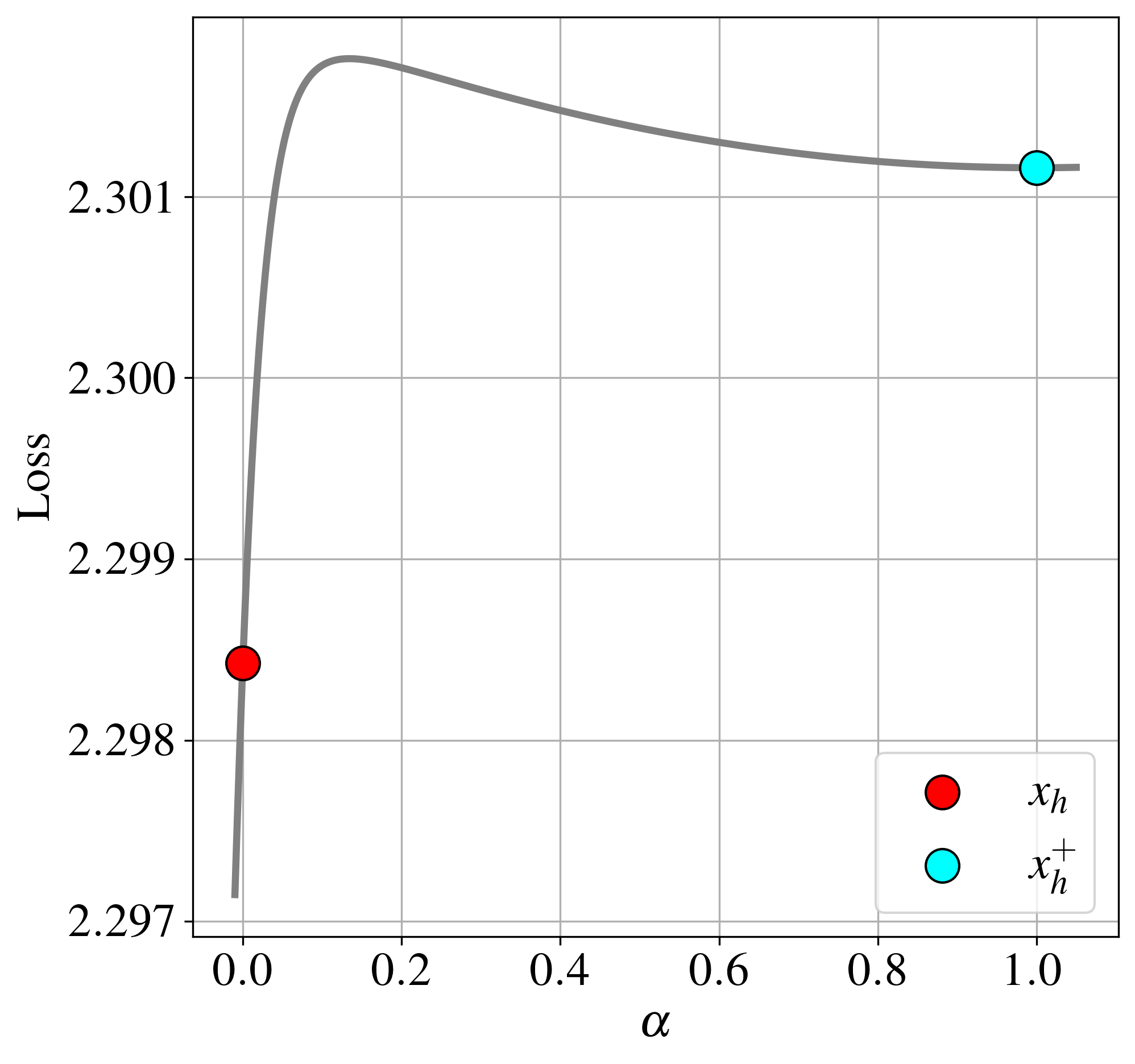}
        \caption{Original loss $f$}
    \end{subfigure}
    \hspace{0.35in}
    \begin{subfigure}{0.36\linewidth}
        \centering
        \includegraphics[width=\linewidth]{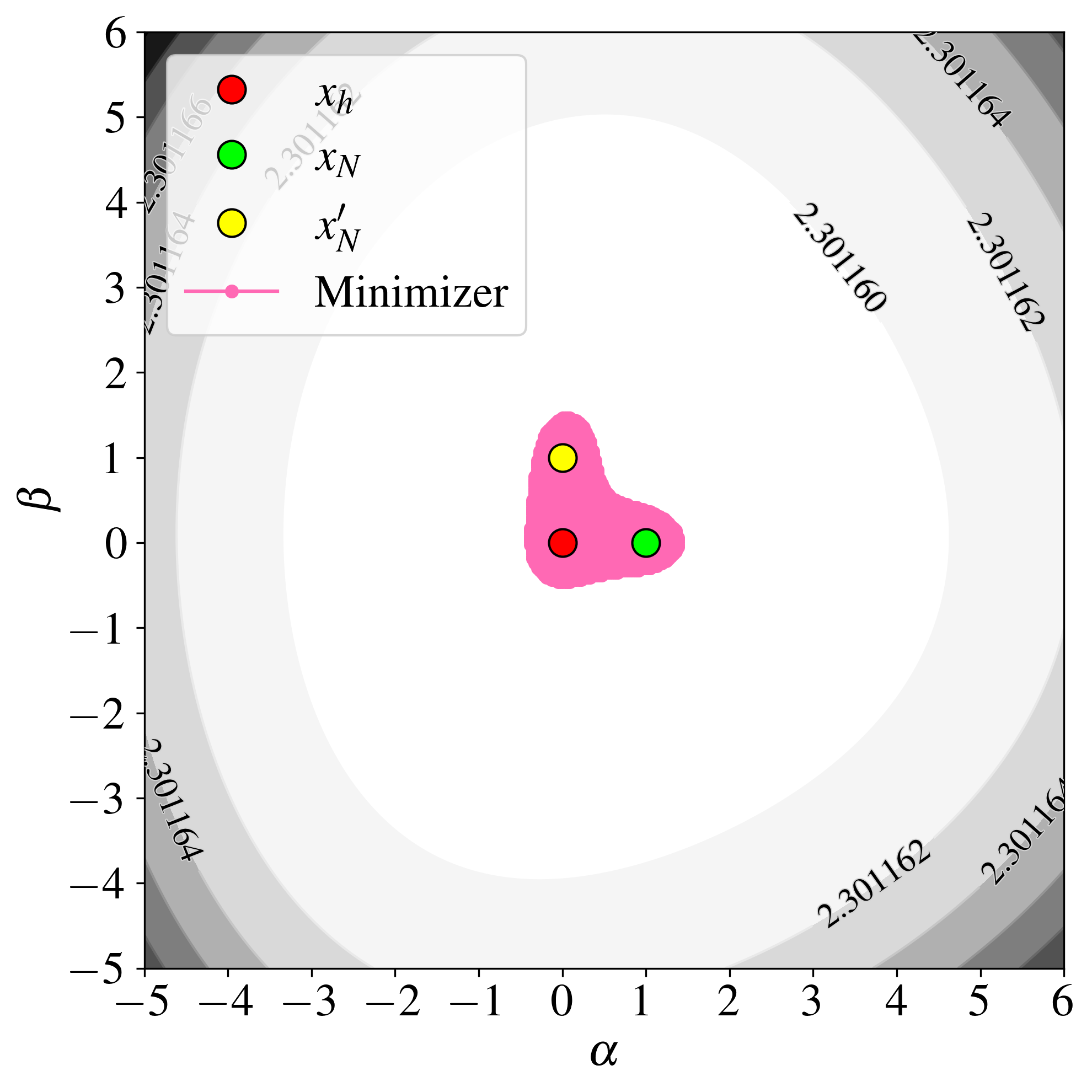}
        \caption{SAM loss $f^{\mathrm{SAM}}$}
    \end{subfigure}
    \caption{Visualizations of the hallucinated minimizer $x_h$: 
(a) original loss $f$ along the line between $x_h$ and $x_h^+ = x_h + \rho \frac{\nabla f(x_h)}{\| \nabla f(x_h) \|}$; 
(b) $f^{\mathrm{SAM}}$ over the affine plane through $x_h$ spanned by the directions $(x_N-x_h)$ and $(x_N'-x_h)$.
}
    \label{fig:HMexist_app}
\end{figure}

\begin{figure}[t]
    \centering
    \begin{minipage}[t]{0.4\textwidth}
        \centering
        \includegraphics[width=\linewidth]{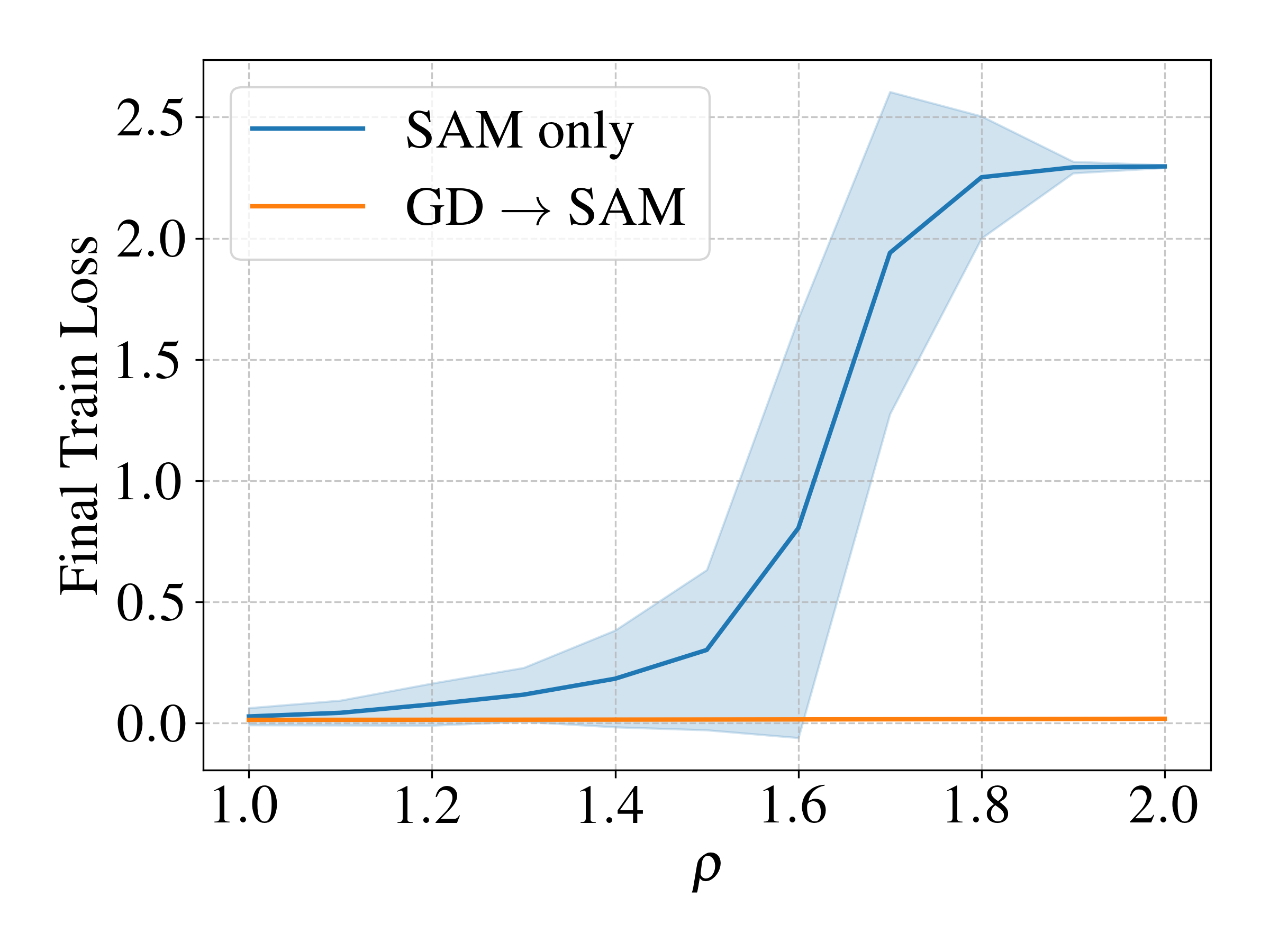}
        \caption*{(a) Final train loss}
    \end{minipage}
    \hspace{0.2in}
    \begin{minipage}[t]{0.4\textwidth}
        \centering
        \includegraphics[width=\linewidth]{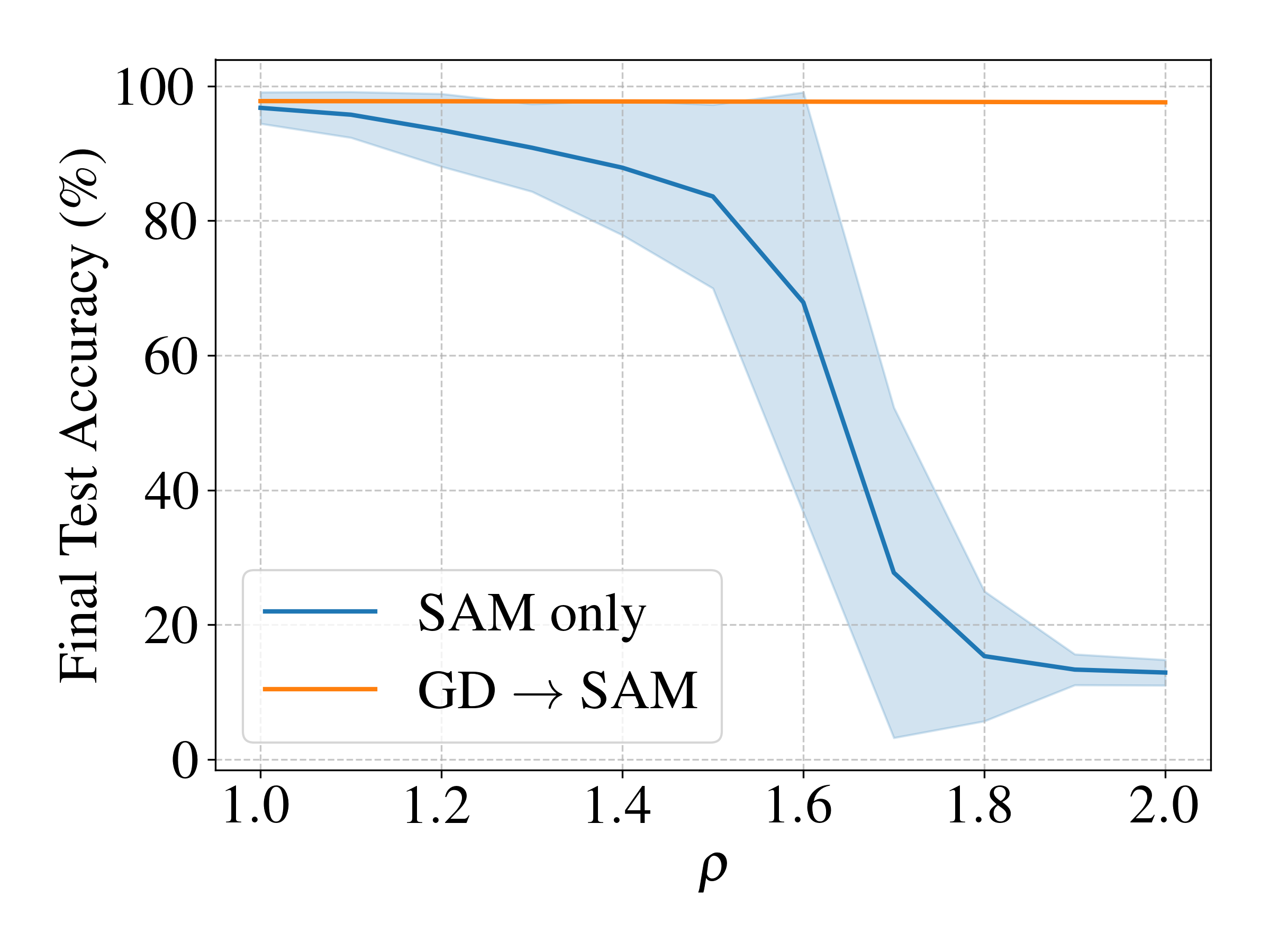}
        \caption*{(b) Final test accuracy}
    \end{minipage}
    \caption{Comparison of SAM-only and the switching strategy across different perturbation radii $\rho$.
Results are obtained using SAM with full-batch gradients over 80 seeds.
Bold lines indicate the mean, and shaded areas represent the standard deviation.}
    \label{fig:loss_acc_vs_rho}
\end{figure}

It remains to show that every $x\in\mathcal{H}$ is a hallucinated minimizer.
Take any $x\in\mathcal{H}$.
By definition, $x^{+}=\Phi_\rho(x)\in\mathcal{M}\subseteq\argmin f$,
so $\nabla f(x^{+})=0$ and $f(x^{+})=\min f$.
Moreover, since $x\in V\subseteq\Omega$, we have $\nabla f(x)\neq 0$.
Finally, $f^{\mathrm{SAM}}(x)=f(x^{+})=\min f$, and hence $x$ is a (global, hence local) minimizer of $f^{\mathrm{SAM}}$ with $\nabla f(x)\neq 0$ and $\nabla f(x^{+})=0$.
Thus, $x$ is a hallucinated minimizer.
\end{proof}

\section{Further details on experiments}\label{app:exp}

\subsection{SAM with full-batch gradients on MNIST}\label{app:full_mnist}

In Section~\ref{sub:exp1}, we train a neural network using SAM with full-batch gradients.  
Specifically, the model is a two-layer network with 128 hidden units and \texttt{Tanh} activations, trained on the MNIST dataset~\citep{lecun1998mnist}.  
The classification task uses cross-entropy loss.  
Training is implemented in PyTorch~\citep{paszke2019pytorch} with a learning rate of $0.01$, momentum $0.9$, and no weight decay. We run $20$ million updates with perturbation radius $\rho = 1.8$ to obtain the convergence point $x_h$, whose loss landscape is shown in Figure~\ref{fig:HMexist}.
In this subsection, we provide additional visualizations to further examine the local properties of $x_h$.

Figure~\ref{fig:HMexist_app}(a) presents a one-dimensional view along the line connecting $x_h$ and $x_h^+ = x_h + \rho \frac{\nabla f(x_h)}{\| \nabla f(x_h) \|} $, parameterized as $x(\alpha) = (1-\alpha)x_h + \alpha x_h^+$.  
The plot shows that $x_h$ is not a minimizer of the original loss and that the surrounding loss landscape differs substantially from that around $x_h^+$. This demonstrates that the phenomenon of SAM converging to a hallucinated minimizer is fundamentally distinct from the case in which a saddle point becomes an attractor, which requires the surrounding quadratic structure to hold~\citep{compagnoni2023sde}.

Figure~\ref{fig:HMexist_app}(b) extends Figure~\ref{fig:HMexist}(c) with an additional visualization of the SAM loss.
We initialize $x_0$ by adding a small random perturbation of magnitude $0.1$ to $x_h$, and then perform $N=1000$ SAM steps, yielding the same $x_N$ reported in Figure~\ref{fig:HMexist}(c).
Applying an independent perturbation followed by the same procedure gives $x_N'$.
We then consider the plane spanned by $x_h$, $x_N$, and $x_N'$, parameterized as $x(\alpha,\beta)=x_h+\alpha d_1+\beta d_2$ with $d_1 = x_N - x_h$ and $d_2$ chosen orthogonal to $d_1$.
On this plane, the visualization shows that the hallucinated minimizers are not confined to a one-dimensional curve but instead extend into a two-dimensional surface-like structure.

In the experiments reported in Figure~\ref{fig:det_sam}, we investigate SAM with full-batch gradients by varying both the perturbation radius and the random seeds.
Under the same experimental setting, Figure~\ref{fig:loss_acc_vs_rho} shows the final training loss and test accuracy at the last step for perturbation radii $\rho = 1.0, 1.1, \ldots, 2.0$, evaluated across 80 seeds.
The results demonstrate that the performance of SAM is highly sensitive to the perturbation radius, whereas the switching strategy maintains stable performance even for larger values of $\rho$.

\begin{figure}[t]
  \centering
  \begin{minipage}[t]{0.48\linewidth}
    \centering
    \includegraphics[width=\linewidth]{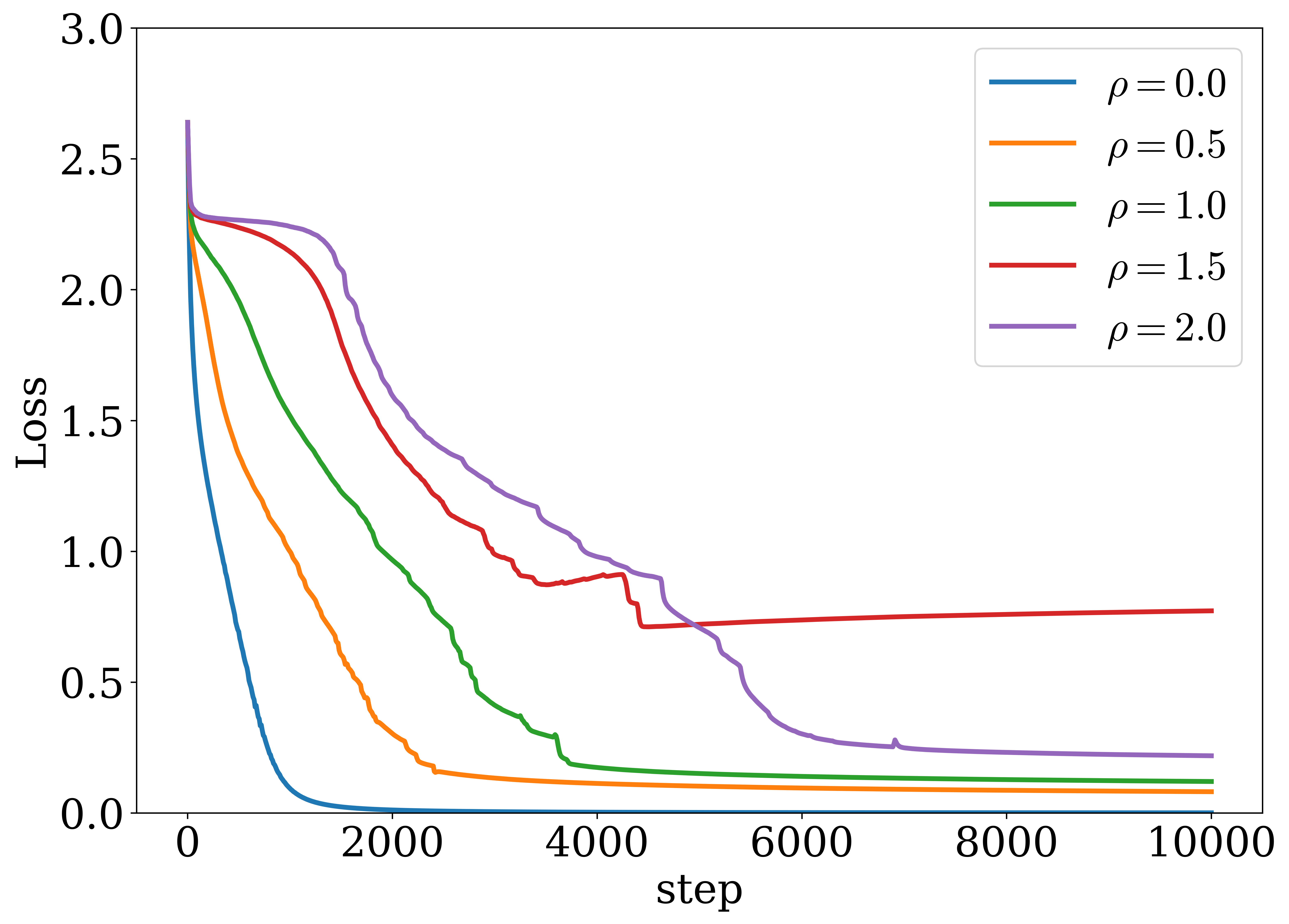}
  \end{minipage}\hfill
  \begin{minipage}[t]{0.48\linewidth}
    \centering
    \includegraphics[width=\linewidth]{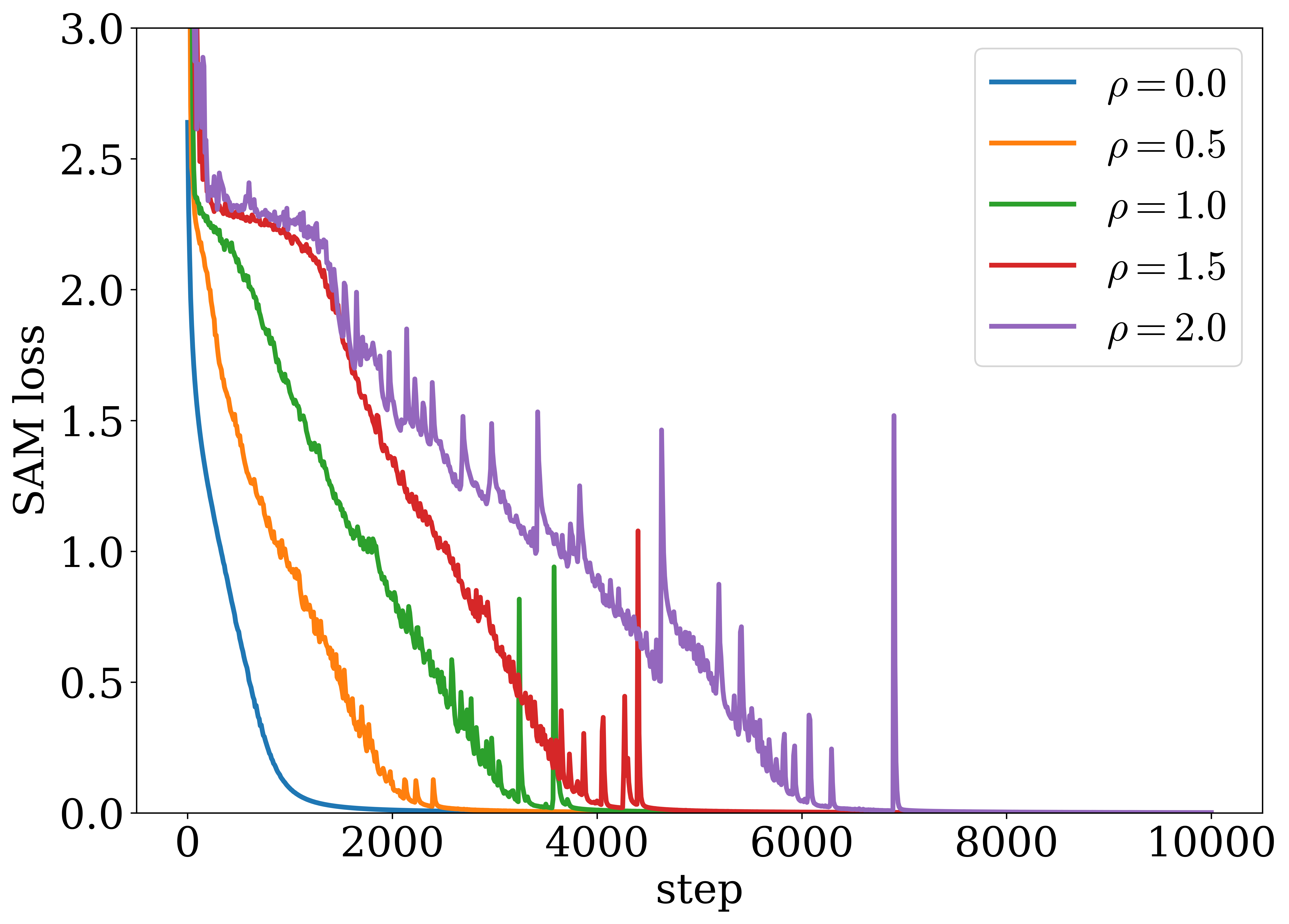}
  \end{minipage}
  \caption{Full-batch SAM on CIFAR-10 with ResNet-18: training loss $f(x)$ (left) and SAM loss $f^{\mathrm{SAM}}(x)$ (right) across perturbation radii $\rho$.}
  \label{fig:fullbatch_sam_loss_c10}
\end{figure}

\begin{figure}[t]
  \centering
  \begin{minipage}[t]{0.48\linewidth}
    \centering
    \includegraphics[width=\linewidth]{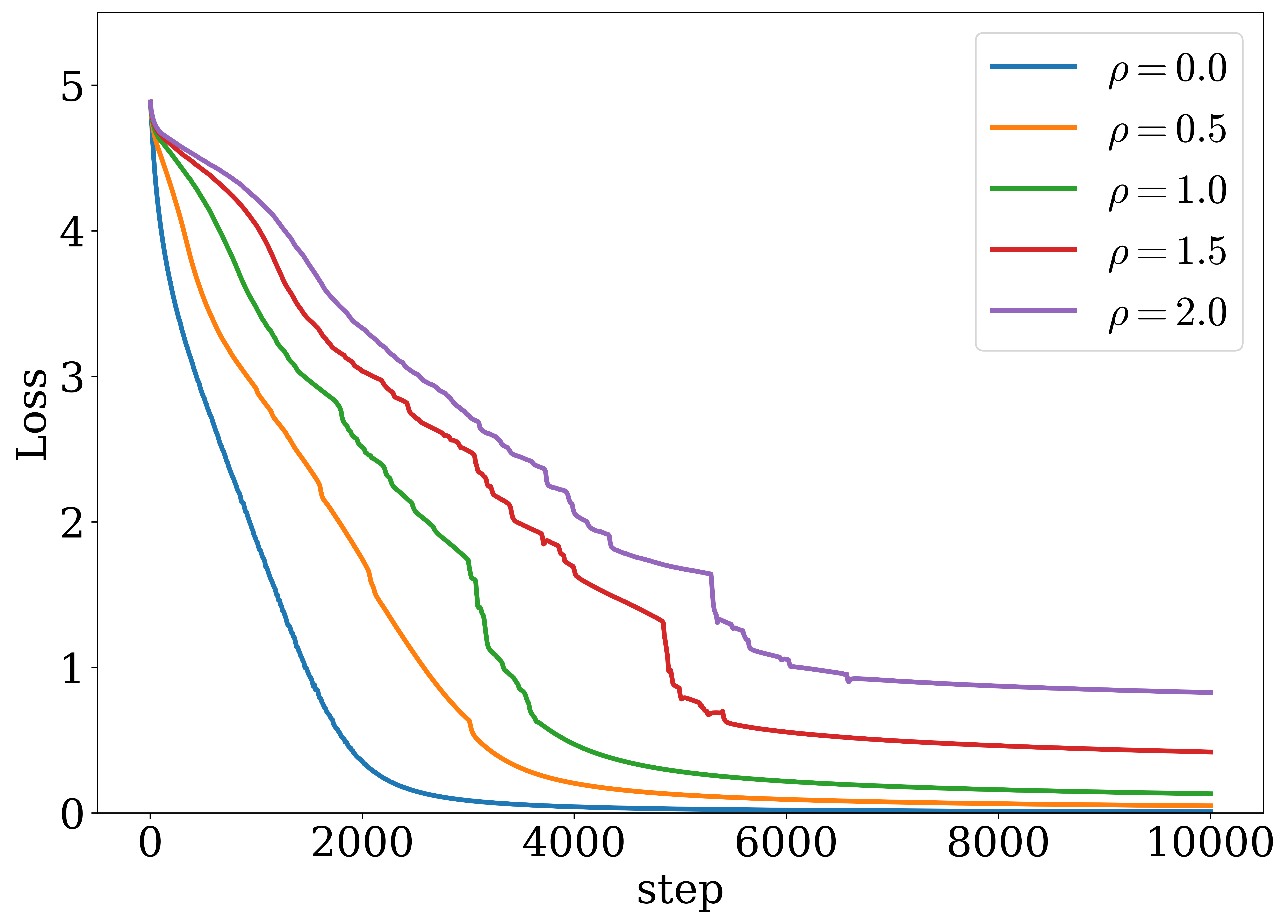}
  \end{minipage}\hfill
  \begin{minipage}[t]{0.48\linewidth}
    \centering
    \includegraphics[width=\linewidth]{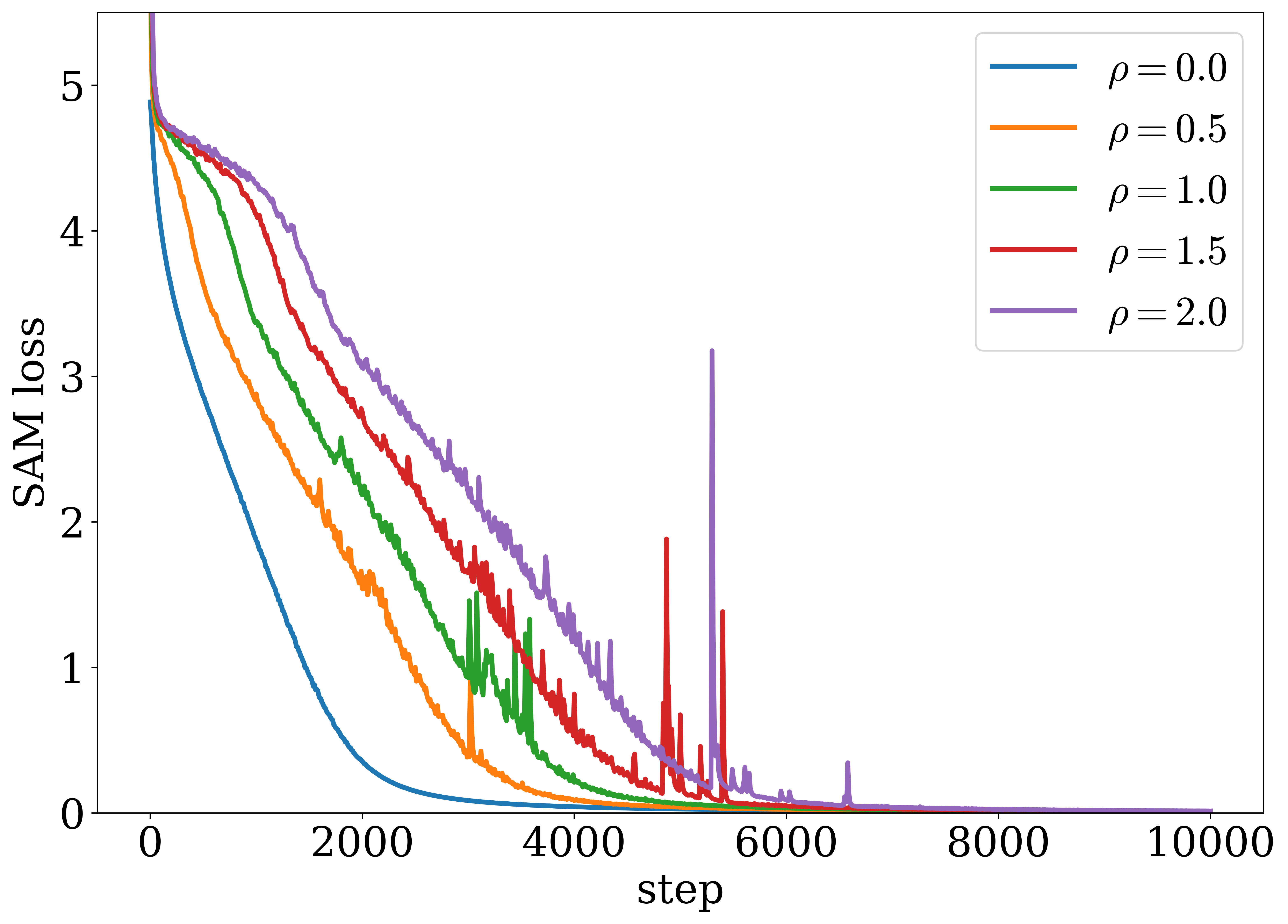}
  \end{minipage}
  \caption{Full-batch SAM on CIFAR-100 with ResNet-18: training loss $f(x)$ (left) and SAM loss $f^{\mathrm{SAM}}(x)$ (right) across perturbation radii $\rho$.}
  \label{fig:fullbatch_sam_loss_c100}
\end{figure}

\begin{figure}[t]
  \centering

  \begin{minipage}[t]{0.23\textwidth}
    \centering \textbf{$\rho = 0.1$}
  \end{minipage}
  \begin{minipage}[t]{0.23\textwidth}
    \centering \textbf{$\rho = 0.4$}
  \end{minipage}
  \begin{minipage}[t]{0.23\textwidth}
    \centering \textbf{$\rho = 0.7$}
  \end{minipage}
  \begin{minipage}[t]{0.23\textwidth}
    \centering \textbf{$\rho = 1.0$}
  \end{minipage}

  \vspace{1mm}

  \begin{subfigure}[t]{\textwidth}
    \centering
    \begin{minipage}[t]{0.23\textwidth}
      \includegraphics[width=\linewidth]{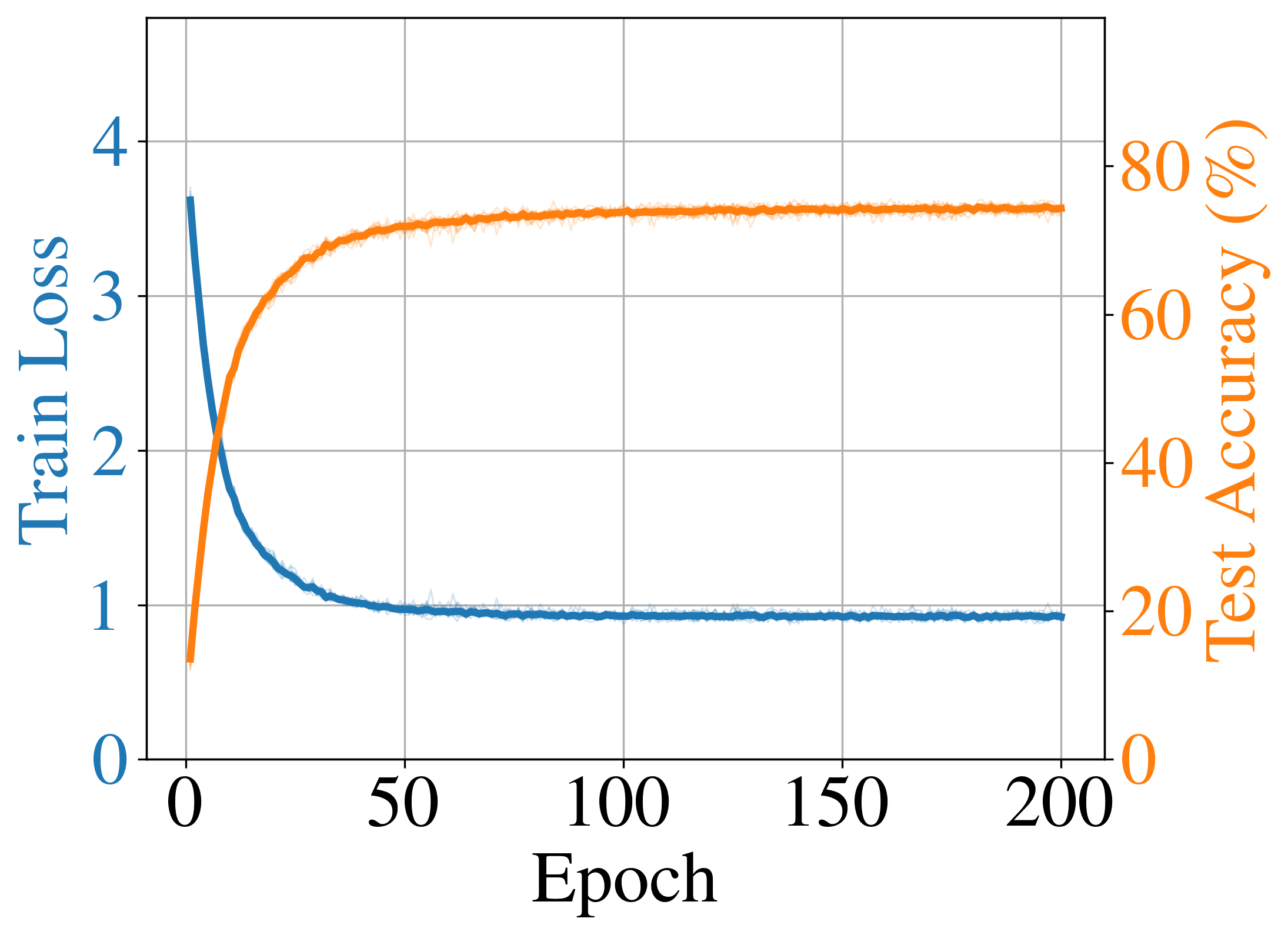}
    \end{minipage}
    \begin{minipage}[t]{0.23\textwidth}
      \includegraphics[width=\linewidth]{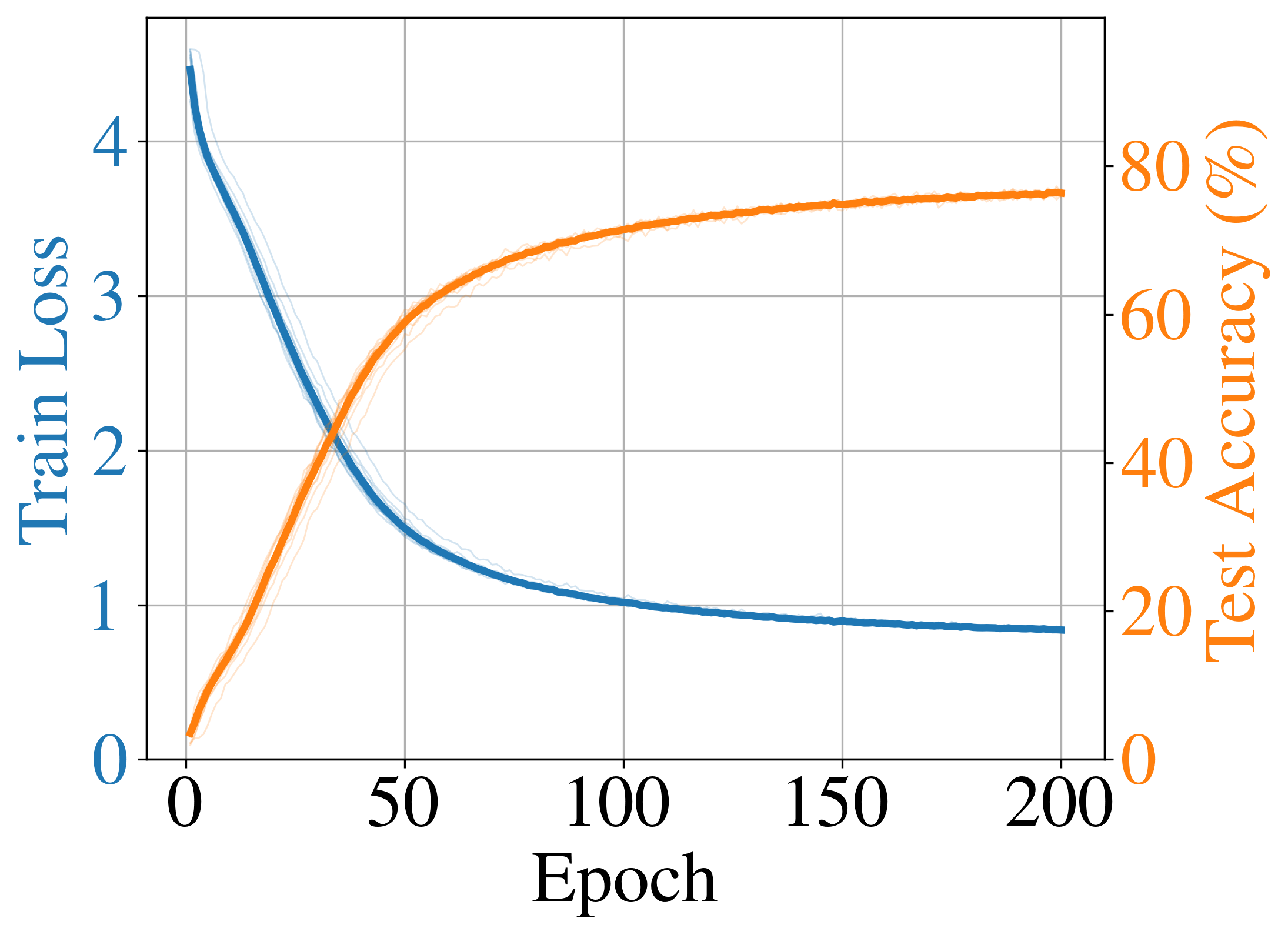}
    \end{minipage}
    \begin{minipage}[t]{0.23\textwidth}
      \includegraphics[width=\linewidth]{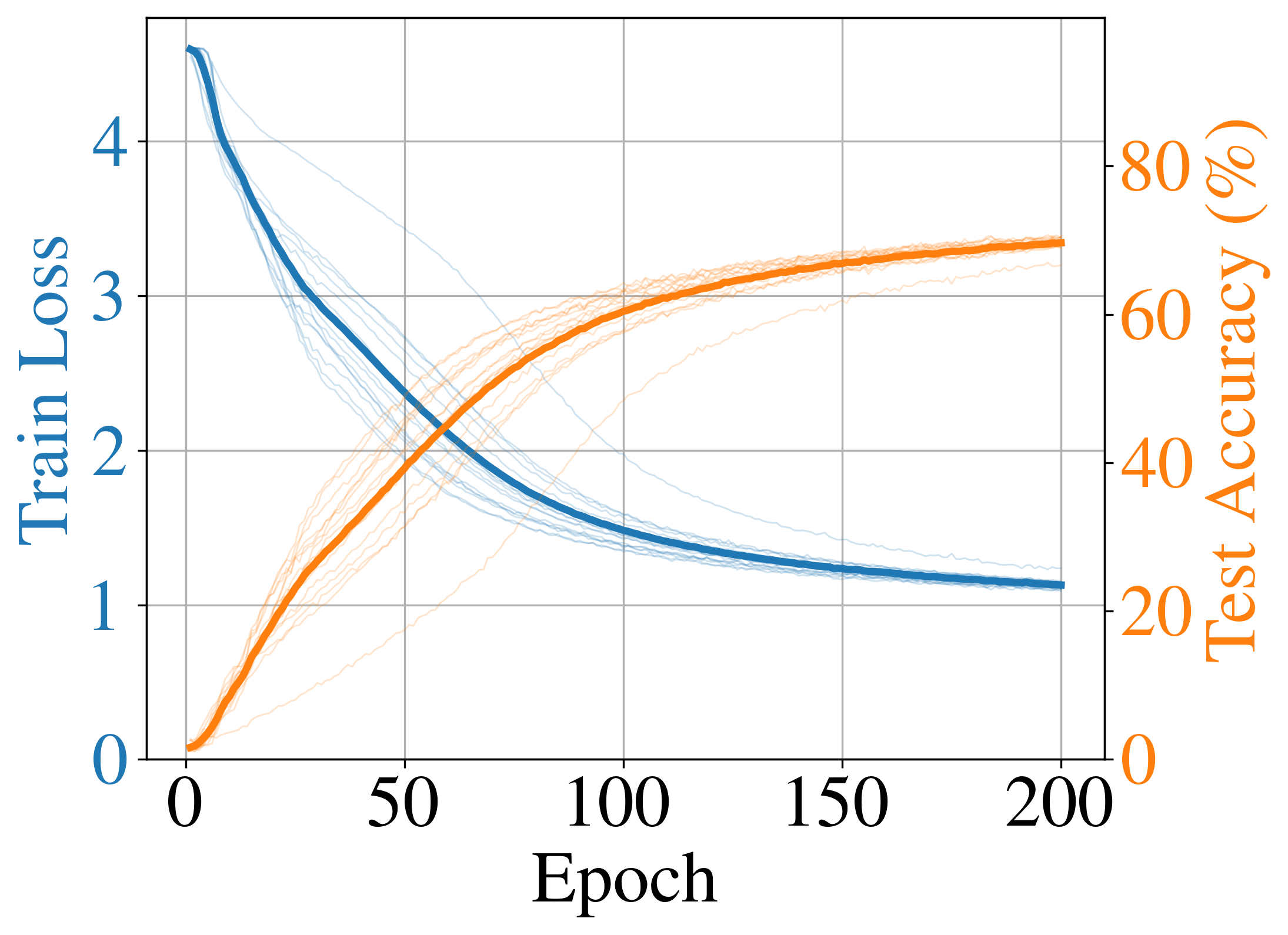}
    \end{minipage}
    \begin{minipage}[t]{0.23\textwidth}
      \includegraphics[width=\linewidth]{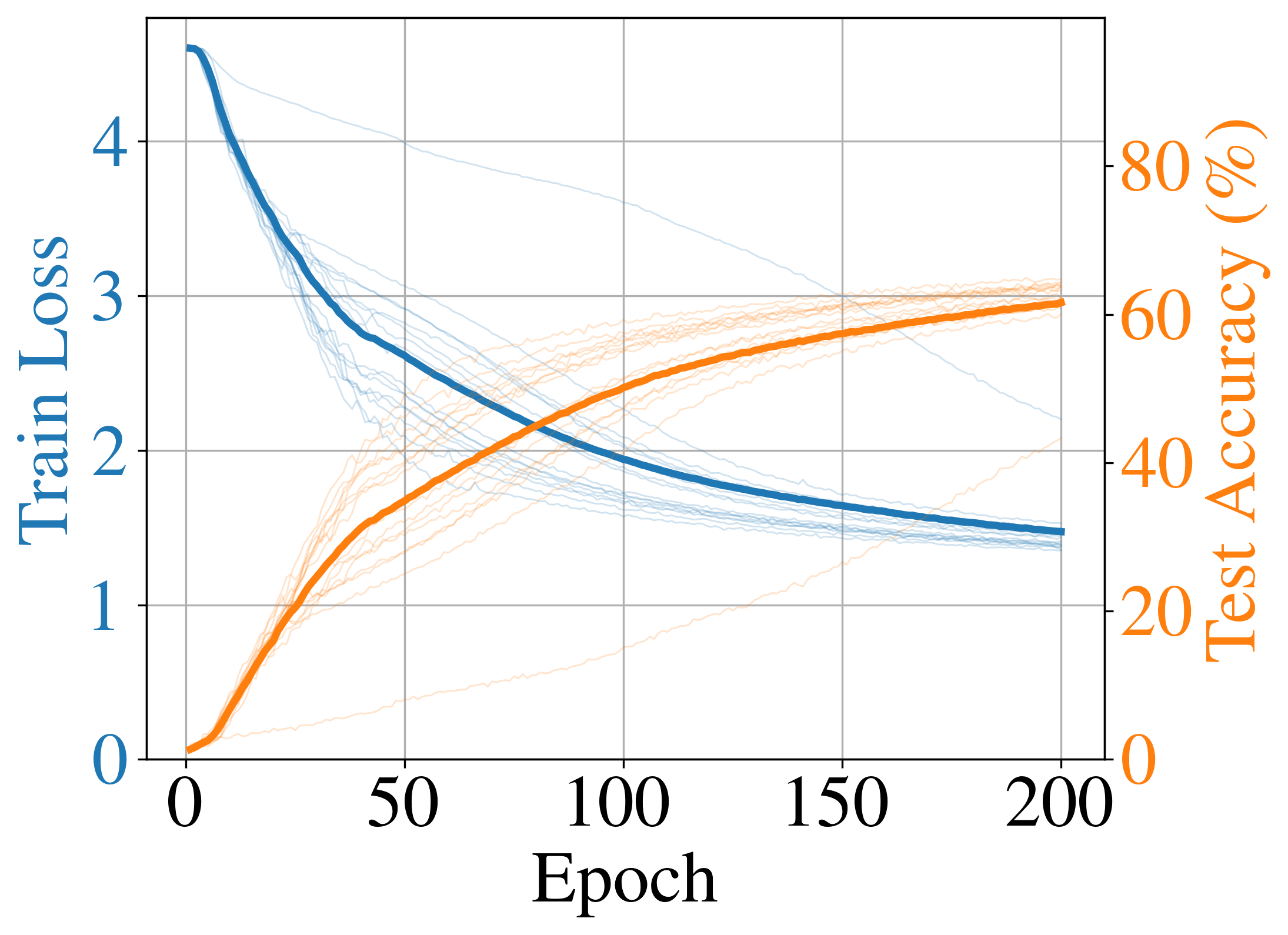}
    \end{minipage}
    \subcaption{SAM-only}
  \end{subfigure}

  \vspace{4mm}

  \begin{subfigure}[t]{\textwidth}
    \centering
    \begin{minipage}[t]{0.23\textwidth}
      \includegraphics[width=\linewidth]{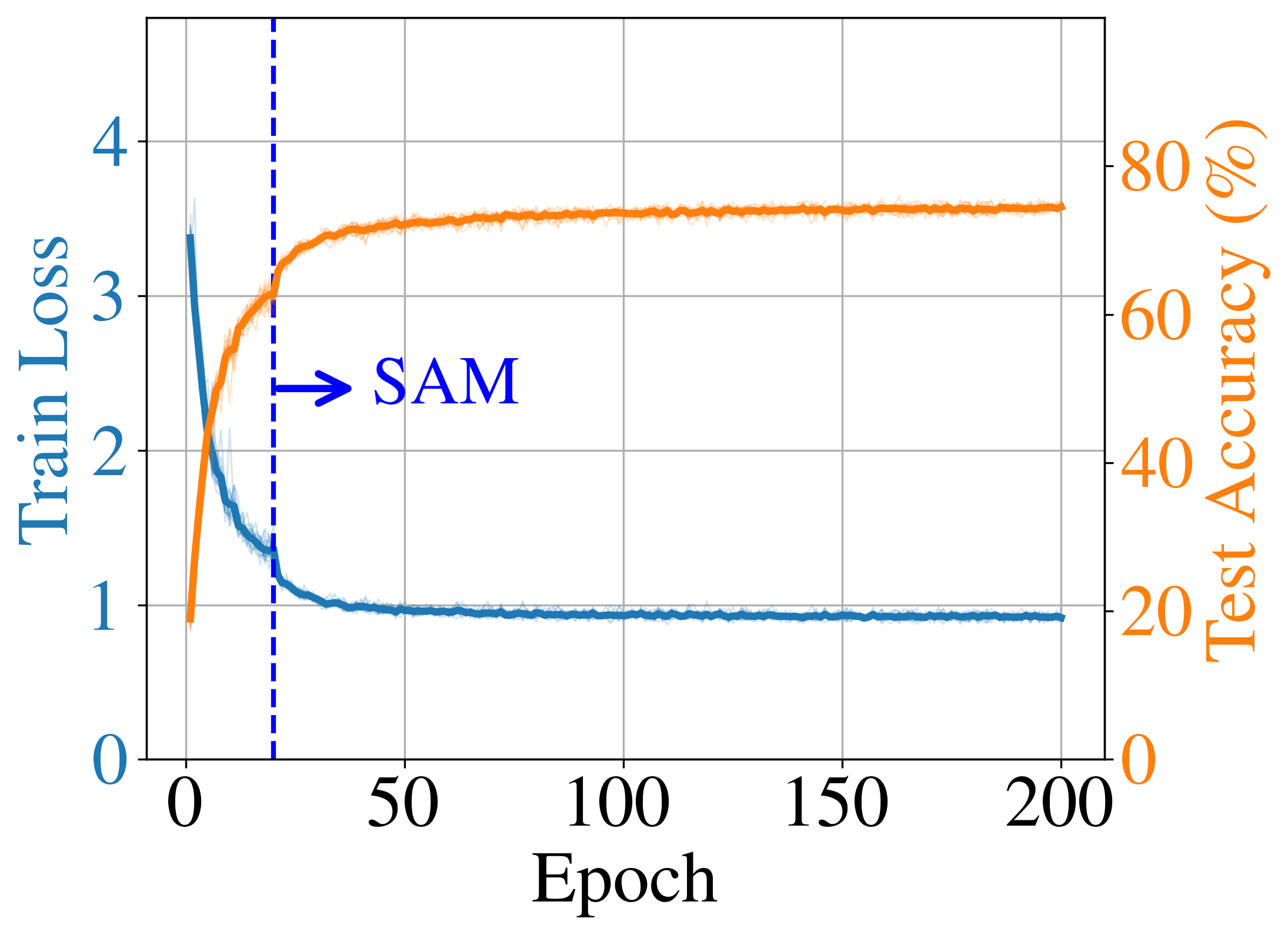}
    \end{minipage}
    \begin{minipage}[t]{0.23\textwidth}
      \includegraphics[width=\linewidth]{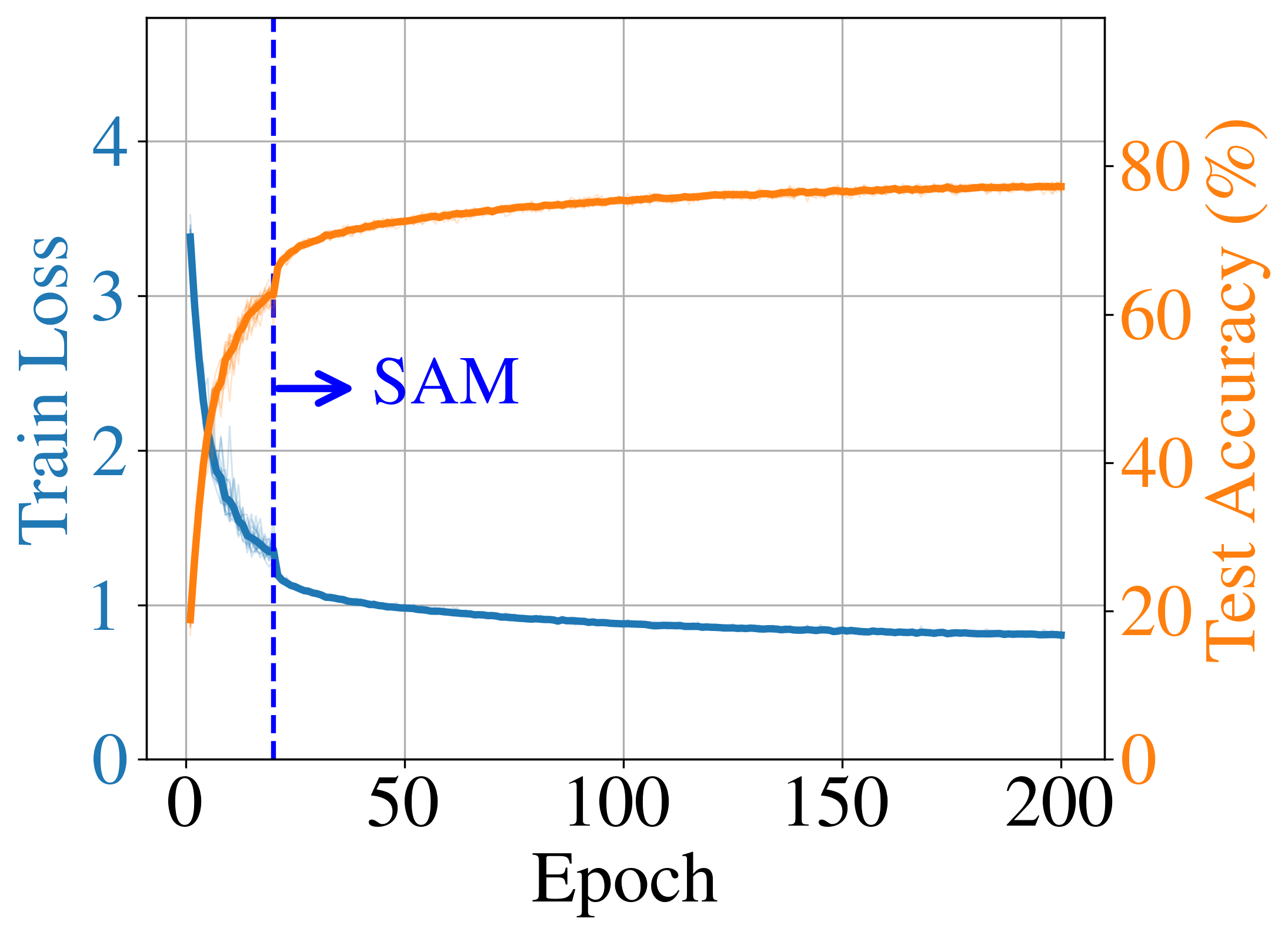}
    \end{minipage}
    \begin{minipage}[t]{0.23\textwidth}
      \includegraphics[width=\linewidth]{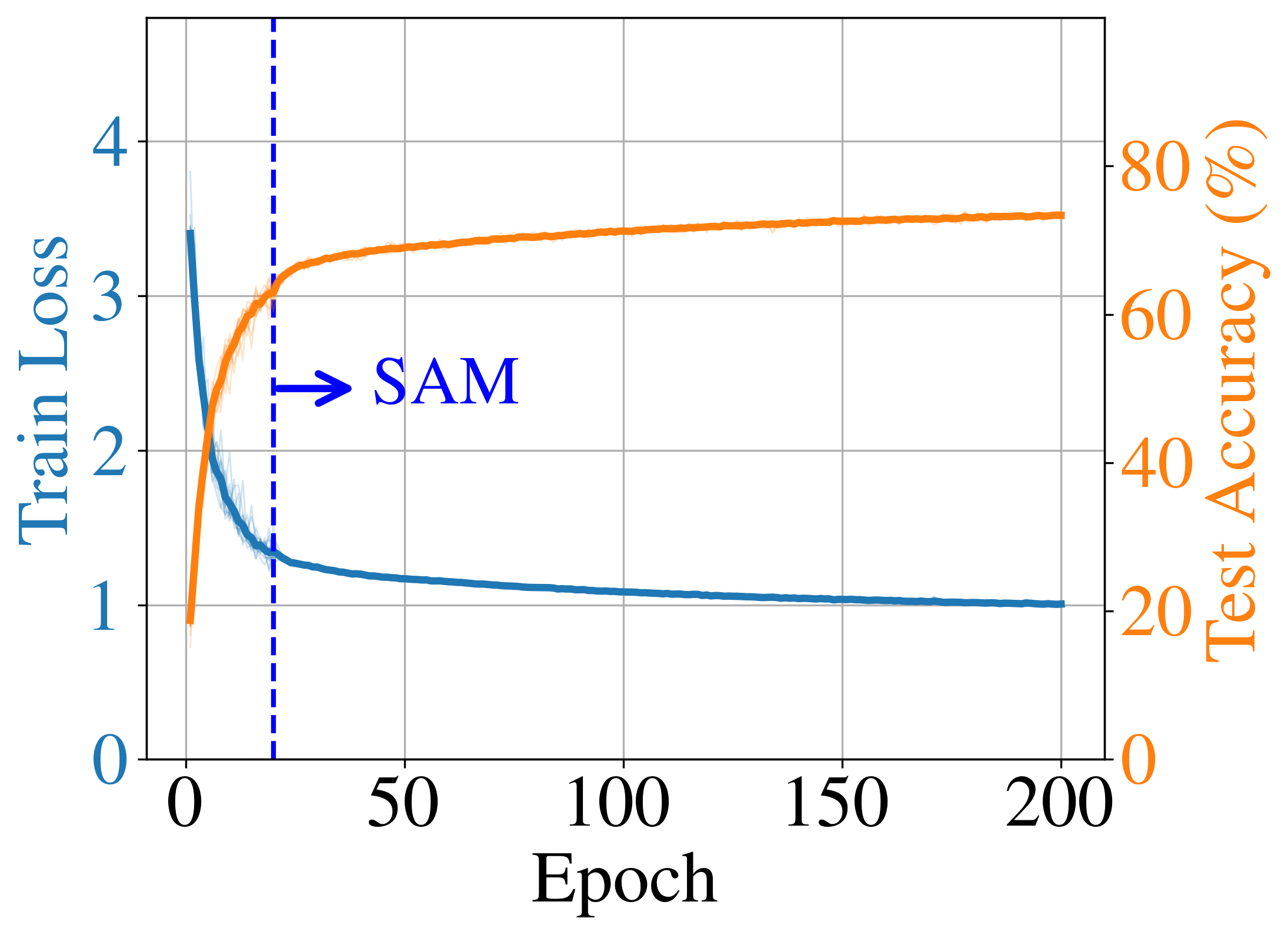}
    \end{minipage}
    \begin{minipage}[t]{0.23\textwidth}
      \includegraphics[width=\linewidth]{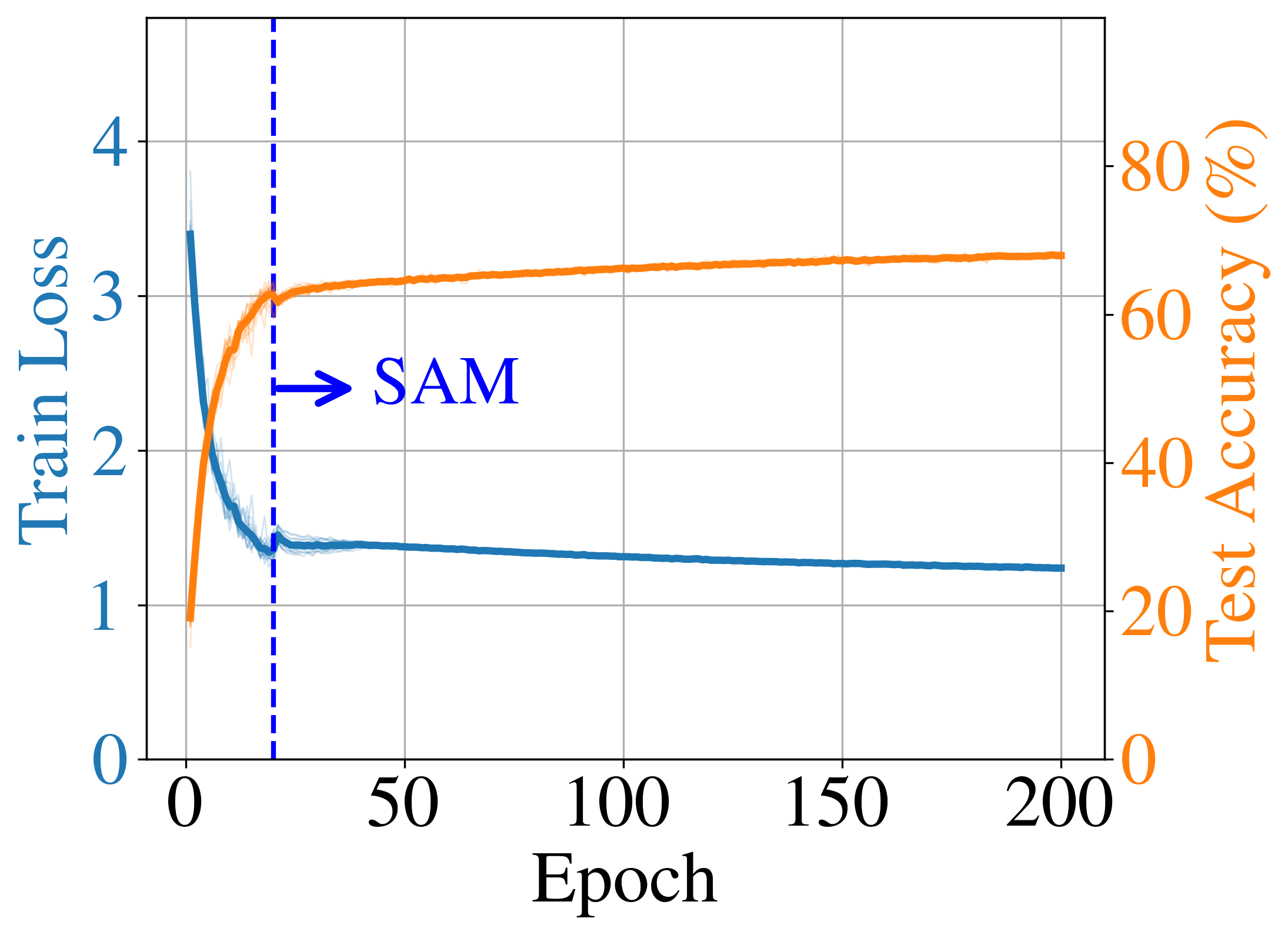}
    \end{minipage}
    \subcaption{SGD $\rightarrow$ SAM}
  \end{subfigure}

  \caption{
Training loss and test accuracy curves for SAM with stochastic gradients on CIFAR-100 using ResNet-18. 
Thin curves show individual seeds, and the bold curve shows the mean across seeds.
The top row shows SAM-only training, while the bottom row applies a switching strategy that runs stochastic gradient descent for the first 10\% of steps before switching to SAM.  
Columns correspond to  perturbation radii $\rho \in \{0.1, 0.4, 0.7, 1.0\}$.  
Final average test accuracies are  74.29\%, 76.36\%, 69.64\%, and 61.58\% for SAM-only, compared to  74.48\%, 77.22\%, 73.35\%, and 67.93\% for the switching strategy.
}
  \label{fig:sto_sam_c100_r18}
\end{figure}

\subsection{SAM with full-batch gradients on CIFAR-10/100}
\label{app:full_cifar}

This subsection reports additional trajectories for the full-batch CIFAR-10/100 experiments in Section~\ref{sub:exp1}.
We use full-batch gradients to isolate the shifted-gradient mechanism studied in our deterministic theory.

We train ResNet-18 from scratch on CIFAR-10/100 using full-batch SAM.
To compute a full-batch gradient in practice, we accumulate gradients over mini-batches across the entire training set.
The base optimizer is SGD with learning rate $10^{-3}$ and momentum $0.9$.
We use standard per-channel normalization and no data augmentation to reduce confounding effects.
We run $10{,}000$ SAM steps and sweep $\rho \in \{0.0, 0.5, 1.0, 1.5, 2.0\}$.

Figures~\ref{fig:fullbatch_sam_loss_c10}--\ref{fig:fullbatch_sam_loss_c100} plot the trajectories of the training loss $f(x_k)$
and the SAM loss $f^{\mathrm{SAM}}(x_k)=f(x_k^{+})$.
Across larger $\rho$, $f(x_k)$ often plateaus above its smallest observed values, while $f(x_k^{+})$ remains much smaller, consistent with the mismatch signatures in Section~\ref{sub:exp1}.

\subsection{SAM with stochastic gradients on CIFAR-100}\label{app:sto_SAM}

We examine how SAM behaves under standard mini-batch training on CIFAR-100 as the perturbation radius $\rho$ varies, as summarized in Figure~\ref{fig:sto_sam_acc}.
ResNet-18 is trained with standard data augmentations, including random cropping with padding, horizontal flipping, and Cutout~\citep{devries2017cutout}.
The mini-batch size is $64$, the learning rate $0.01$, momentum $0.9$, and weight decay $10^{-4}$.
Training proceeds for $200$ epochs with cosine-annealed learning rate, following the implementation details in \citet{li2024friendly}.
We compare SAM-only against a short SGD warm-start before enabling SAM: we run plain SGD for the first $10\%$ of epochs and then continue with (stochastic) SAM for the remainder.

Under this setting, Figure~\ref{fig:sto_sam_c100_r18} reports representative loss/accuracy curves at $\rho\in\{0.1,0.4,0.7,1.0\}$ over $16$ random seeds.
Each curve shows training loss and test accuracy over epochs, with bold lines denoting the mean across seeds.
Overall, as $\rho$ increases, SAM-only training becomes less stable and more sensitive to the choice of $\rho$,
whereas the SGD warm-start improves robustness across the same sweep.

\begin{figure}[t]
    \centering
    \begin{subfigure}[t]{0.24\textwidth}
        \centering
        \includegraphics[width=\linewidth]{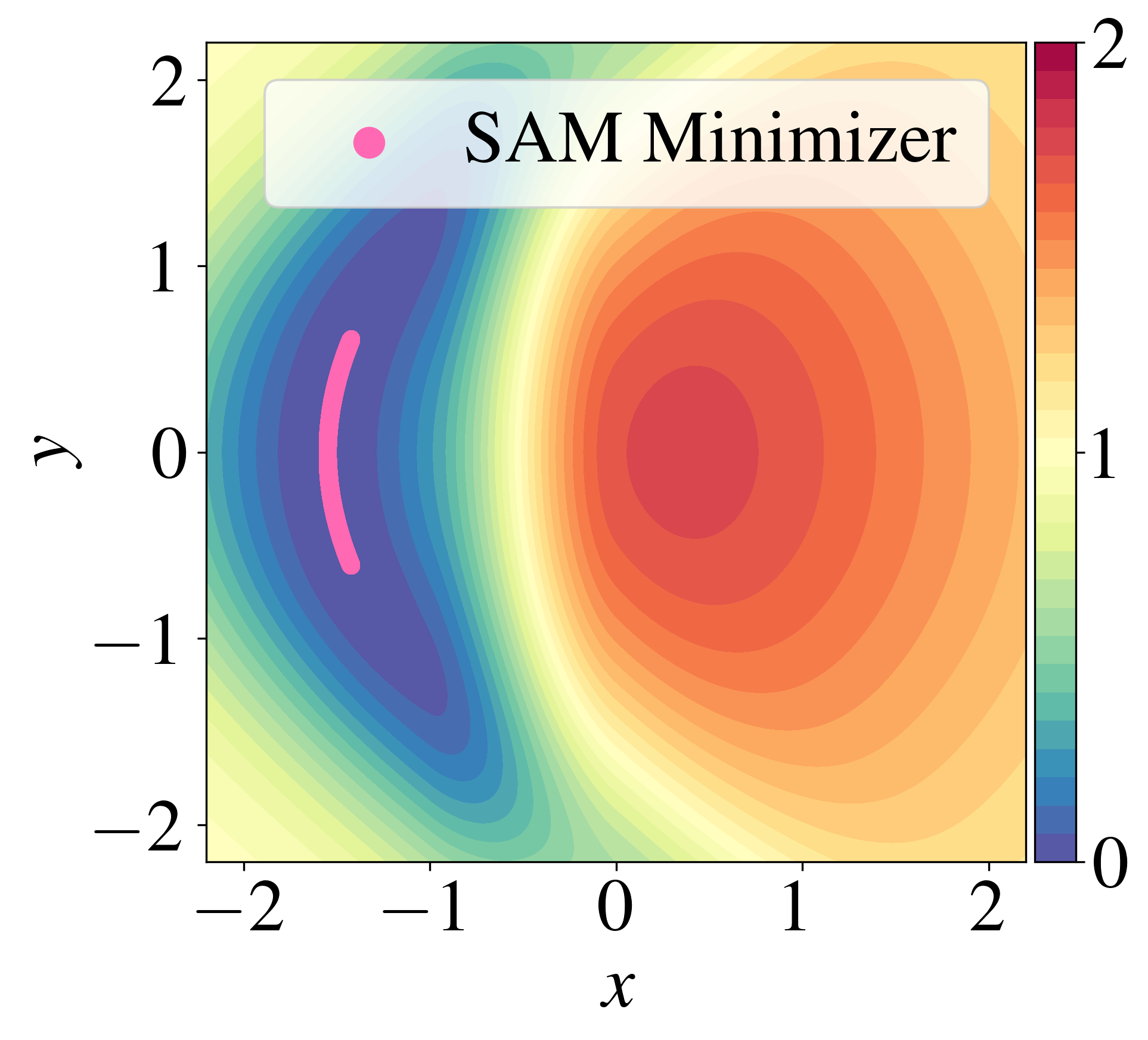}
        \caption{$\rho=0.0$}
    \end{subfigure}
    \begin{subfigure}[t]{0.24\textwidth}
        \centering
        \includegraphics[width=\linewidth]{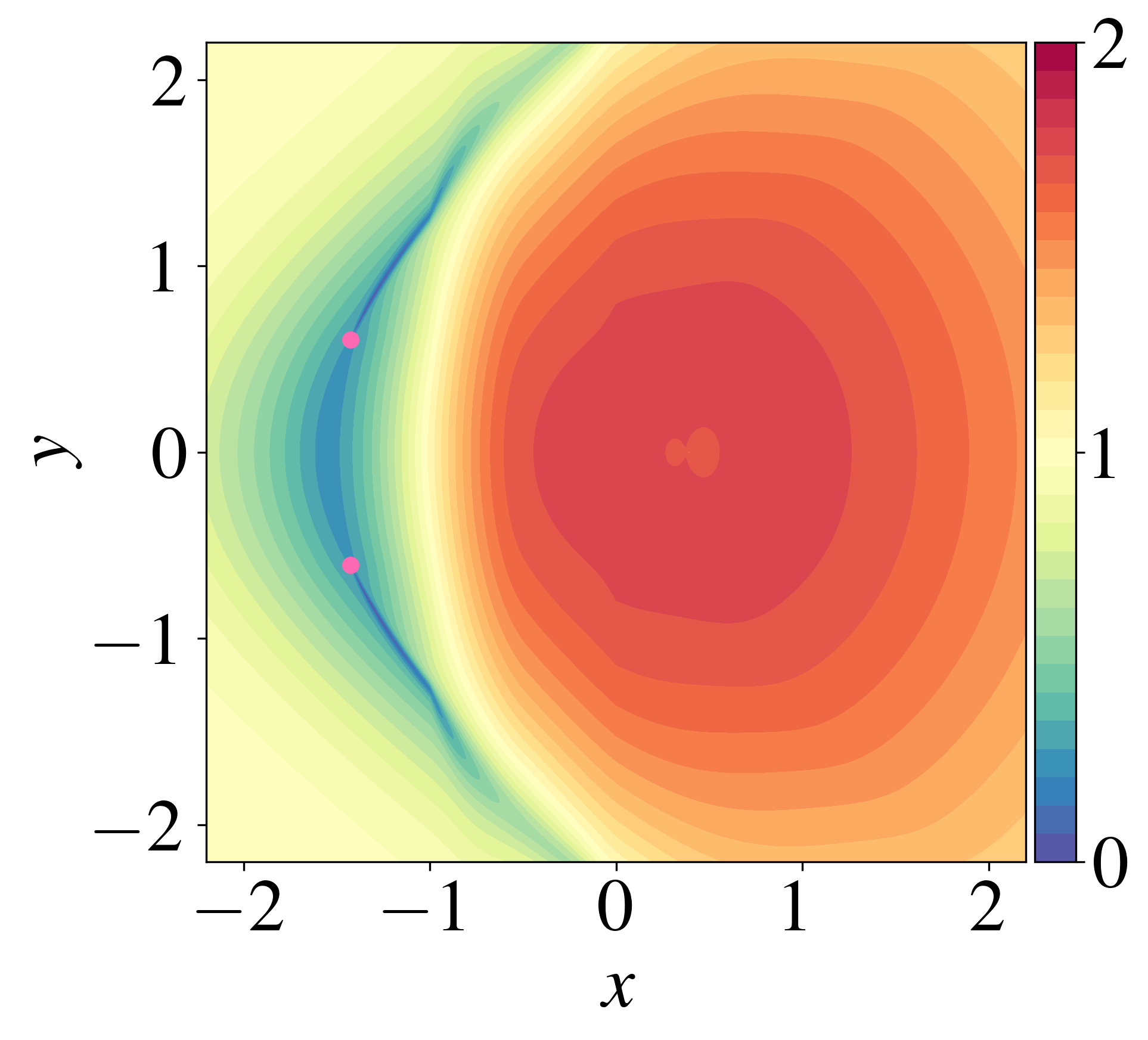}
        \caption{$\rho=0.5$}
    \end{subfigure}
    \begin{subfigure}[t]{0.24\textwidth}
        \centering
        \includegraphics[width=\linewidth]{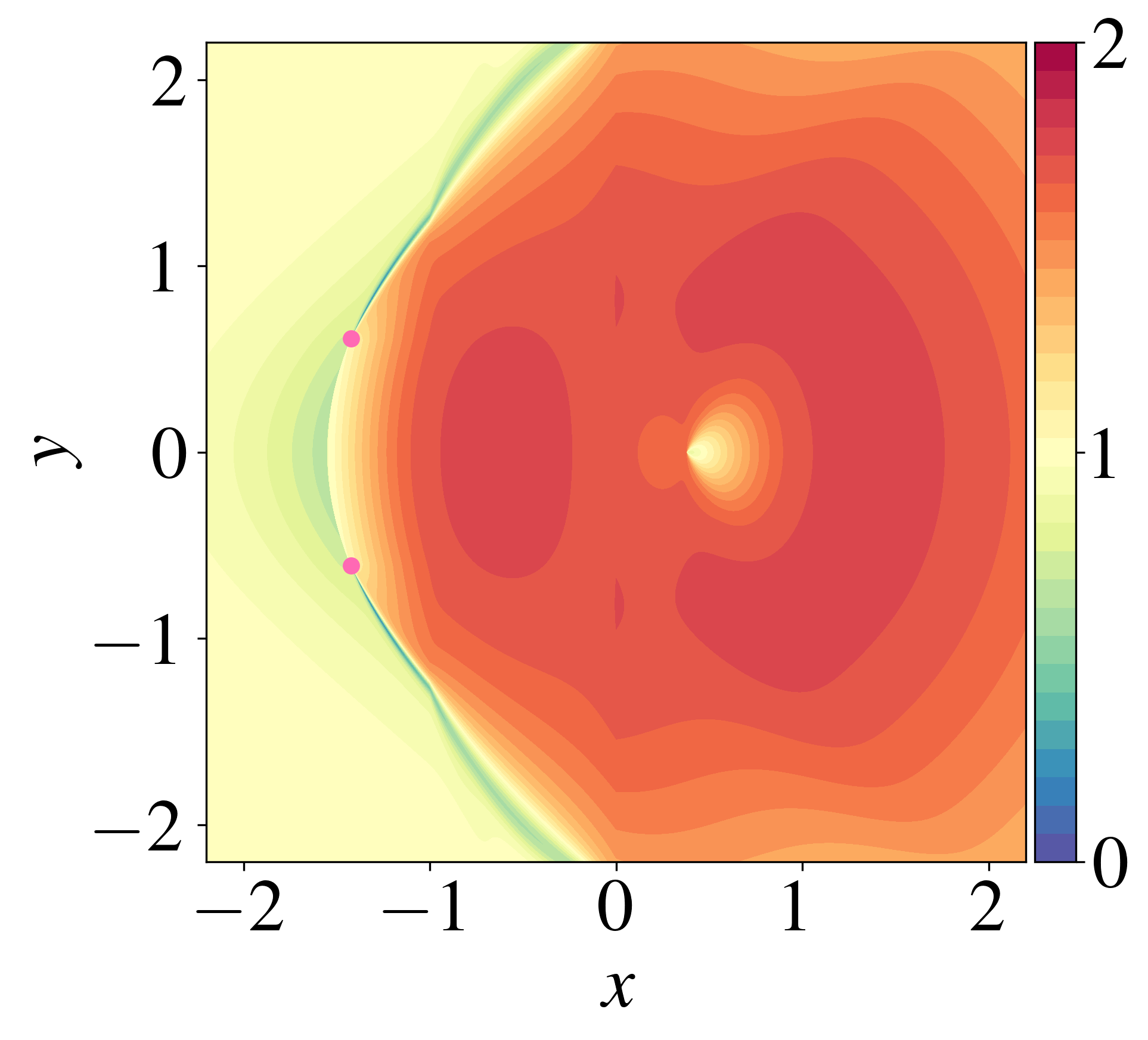}
        \caption{$\rho=1.0$}
    \end{subfigure}
    \begin{subfigure}[t]{0.24\textwidth}
        \centering
        \includegraphics[width=\linewidth]{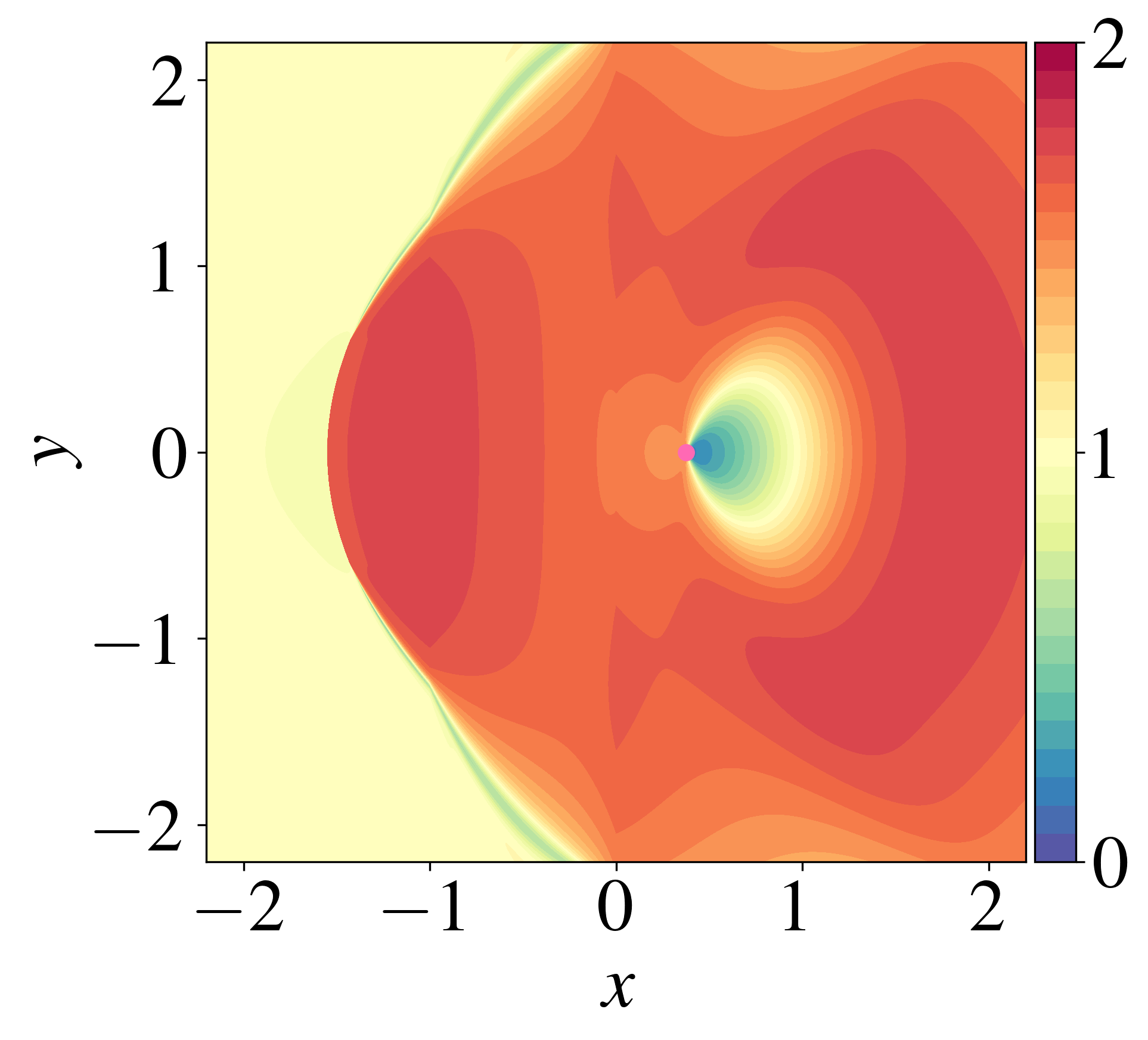}
        \caption{$\rho=1.5$}
    \end{subfigure}

    \vspace{1.0em}

    \begin{subfigure}[t]{0.24\textwidth}
        \centering
        \includegraphics[width=\linewidth]{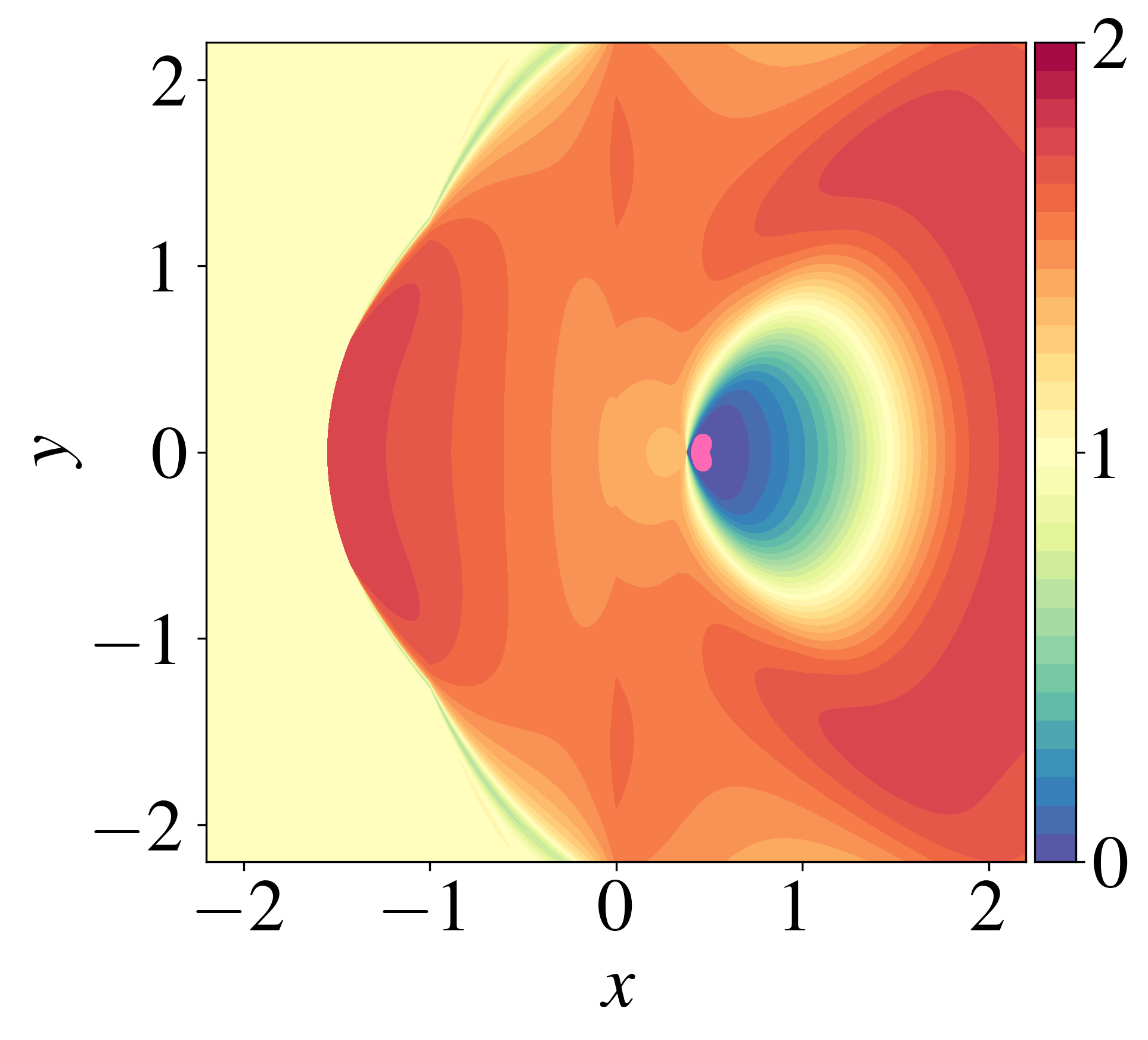}
        \caption{$\rho=2.0$}
    \end{subfigure}
    \begin{subfigure}[t]{0.24\textwidth}
        \centering
        \includegraphics[width=\linewidth]{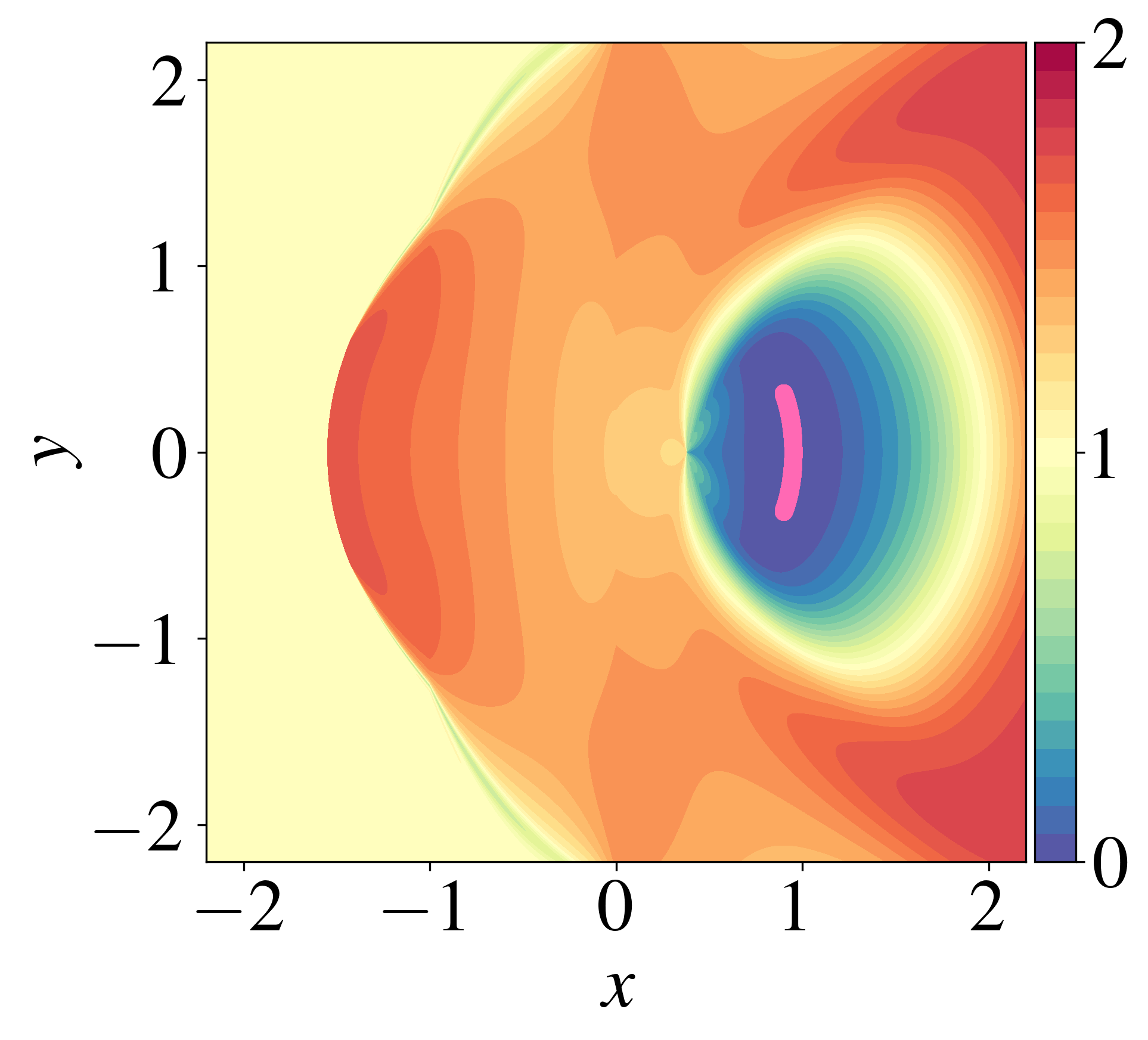}
        \caption{$\rho=2.5$}
    \end{subfigure}
    \begin{subfigure}[t]{0.24\textwidth}
        \centering
        \includegraphics[width=\linewidth]{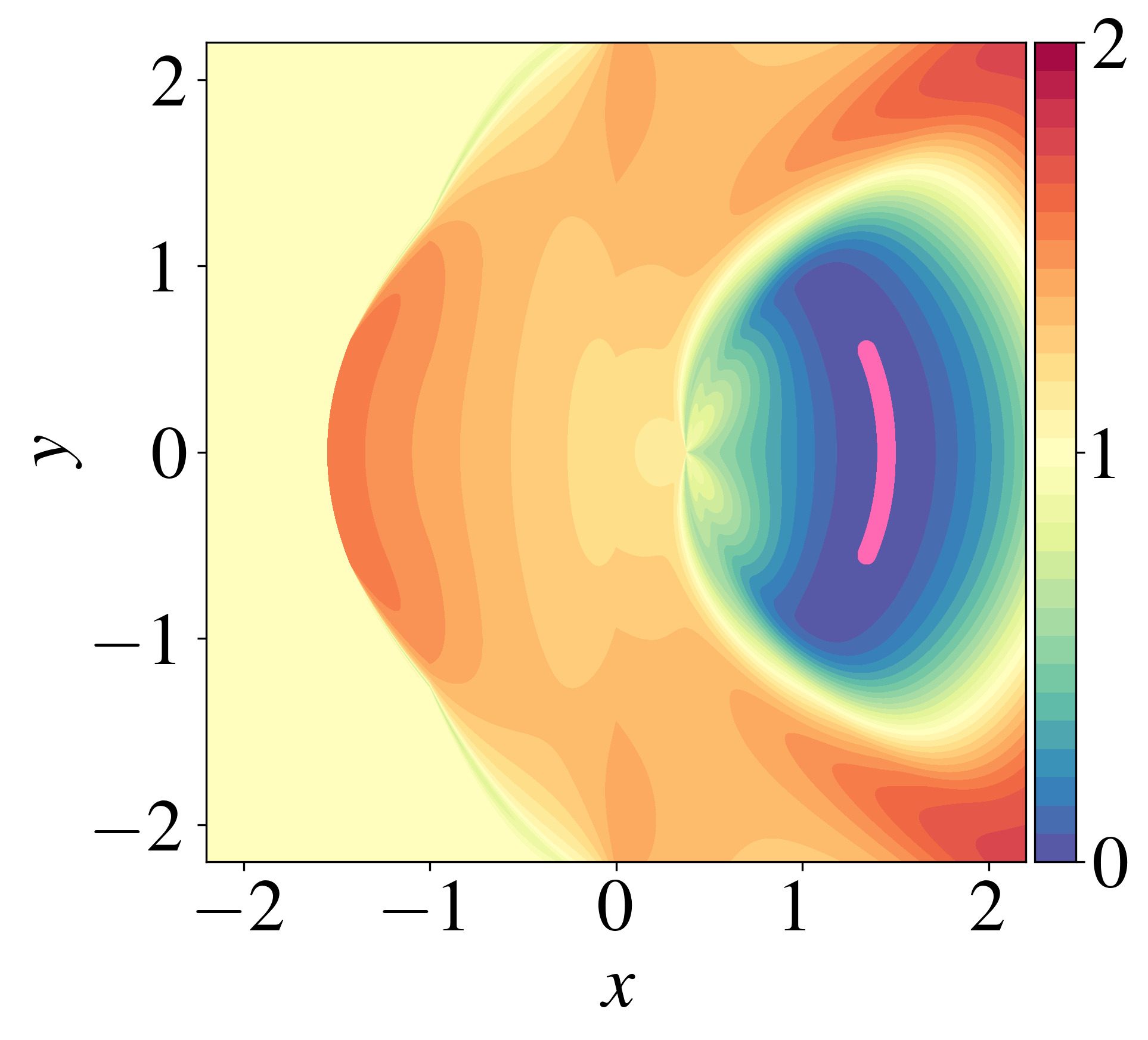}
        \caption{$\rho=3.0$}
    \end{subfigure}
    \begin{subfigure}[t]{0.24\textwidth}
        \centering
        \includegraphics[width=\linewidth]{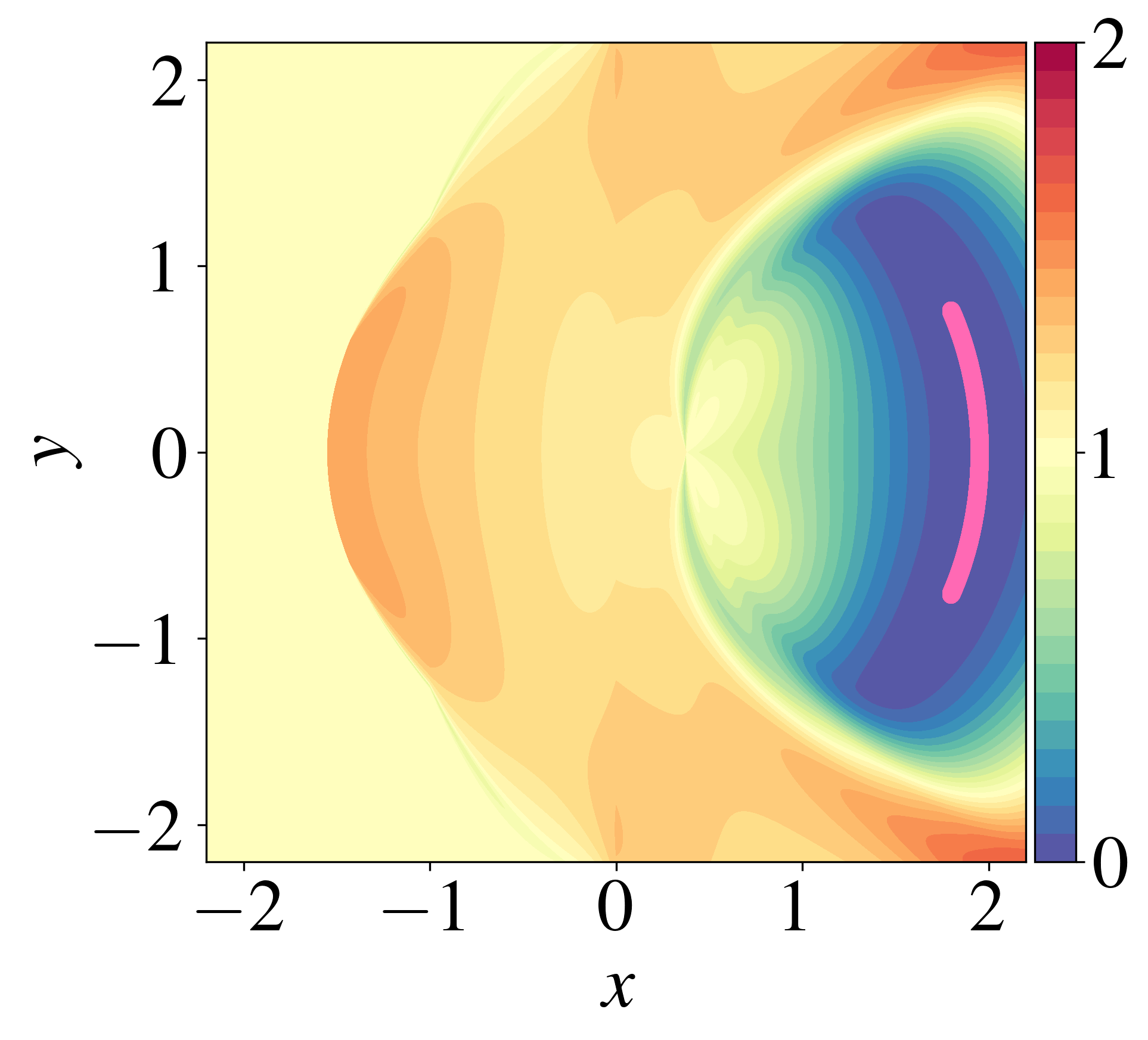}
        \caption{$\rho=3.5$}
    \end{subfigure}

    \caption{SAM loss $f^{\mathrm{SAM}}(x)$ under different perturbation radii $\rho$. The corresponding minimizers of $f^{\mathrm{SAM}}$ are shown in pink.}

    \label{fig:rho_contour}
\end{figure}

\begin{figure}[t]
    \centering
    \begin{subfigure}[t]{0.24\textwidth}
        \centering
        \includegraphics[width=\linewidth]{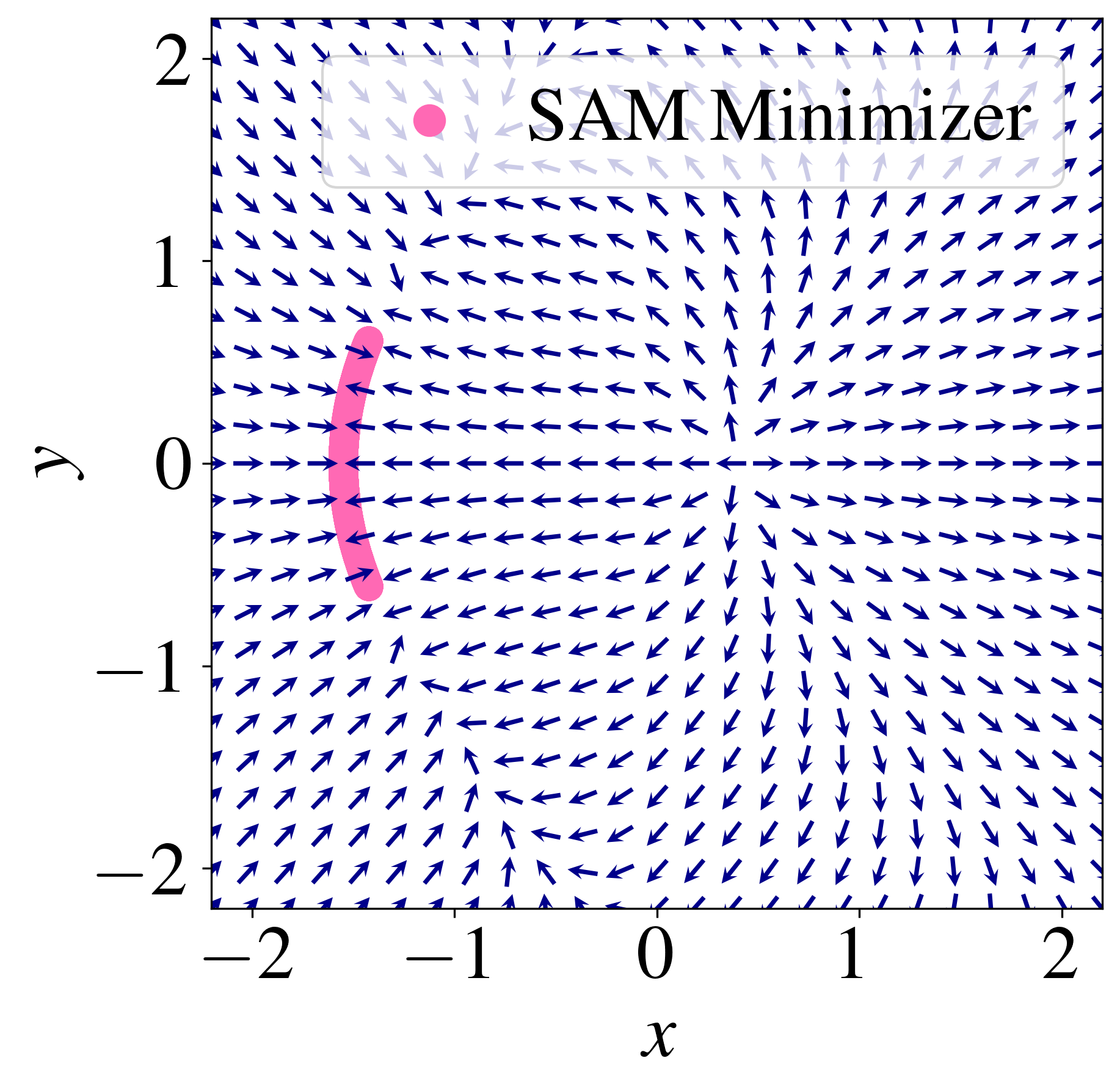}
        \caption{$\rho=0.0$}
    \end{subfigure}
    \begin{subfigure}[t]{0.24\textwidth}
        \centering
        \includegraphics[width=\linewidth]{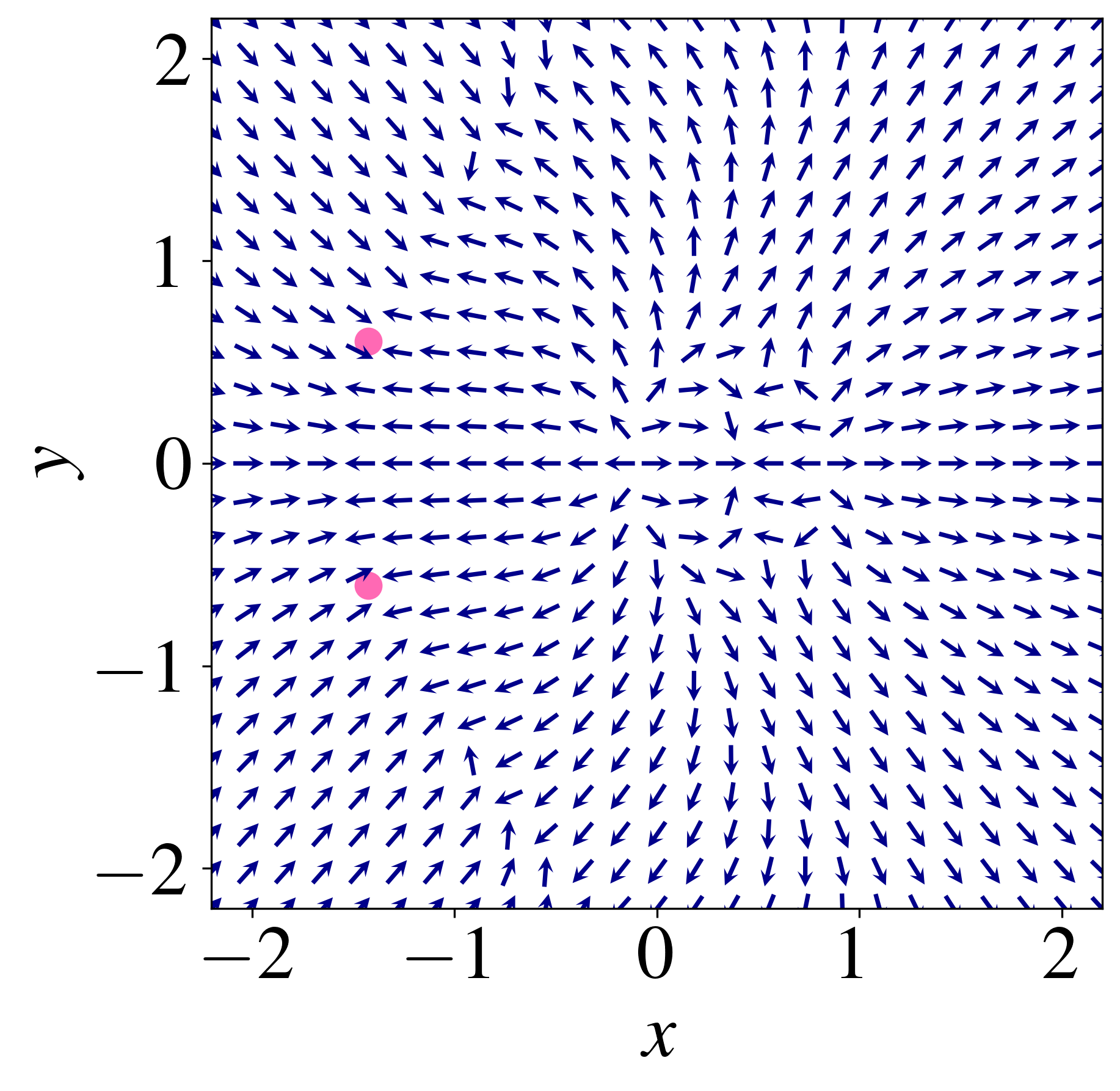}
        \caption{$\rho=0.5$}
    \end{subfigure}
    \begin{subfigure}[t]{0.24\textwidth}
        \centering
        \includegraphics[width=\linewidth]{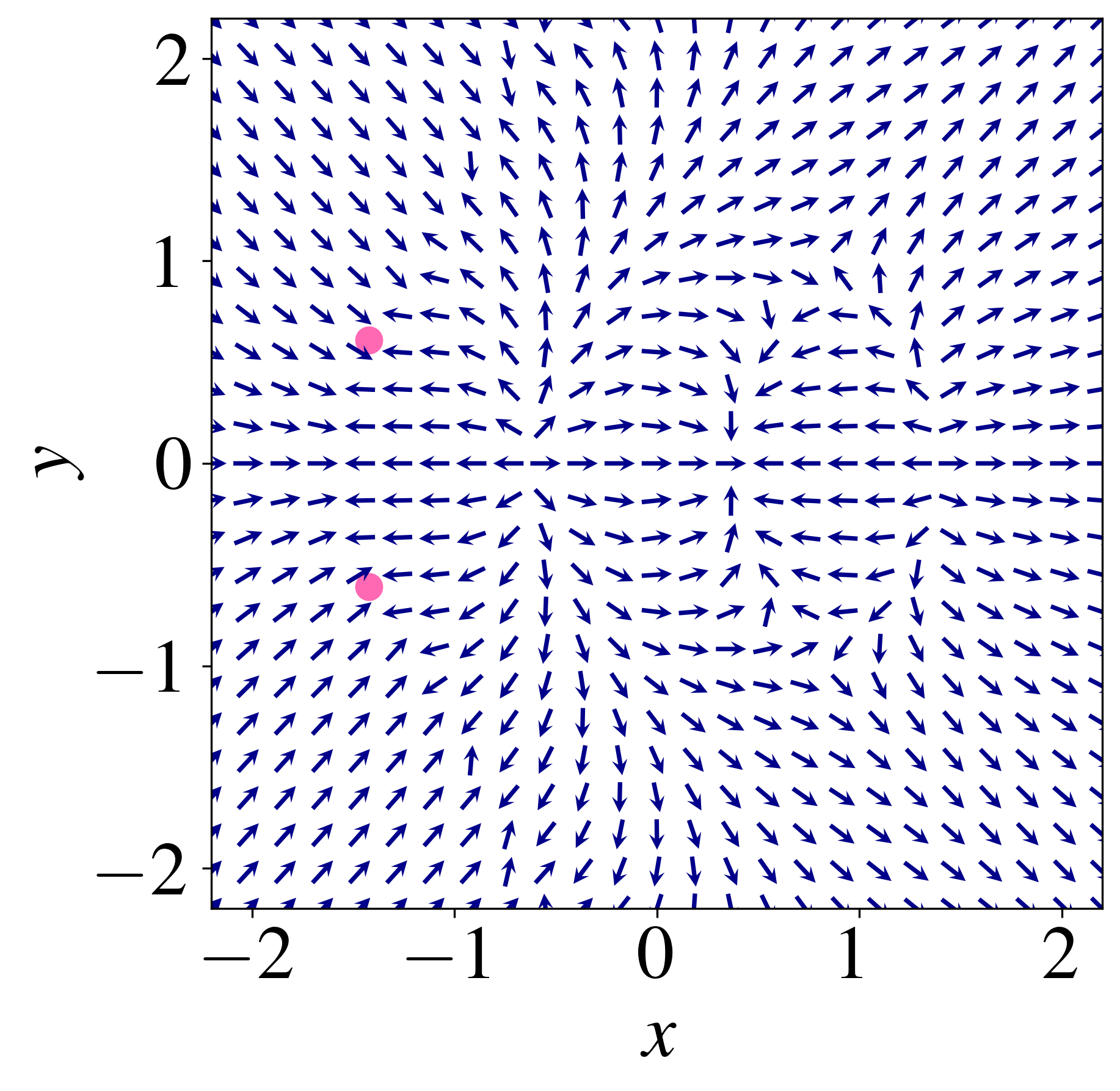}
        \caption{$\rho=1.0$}
    \end{subfigure}
    \begin{subfigure}[t]{0.24\textwidth}
        \centering
        \includegraphics[width=\linewidth]{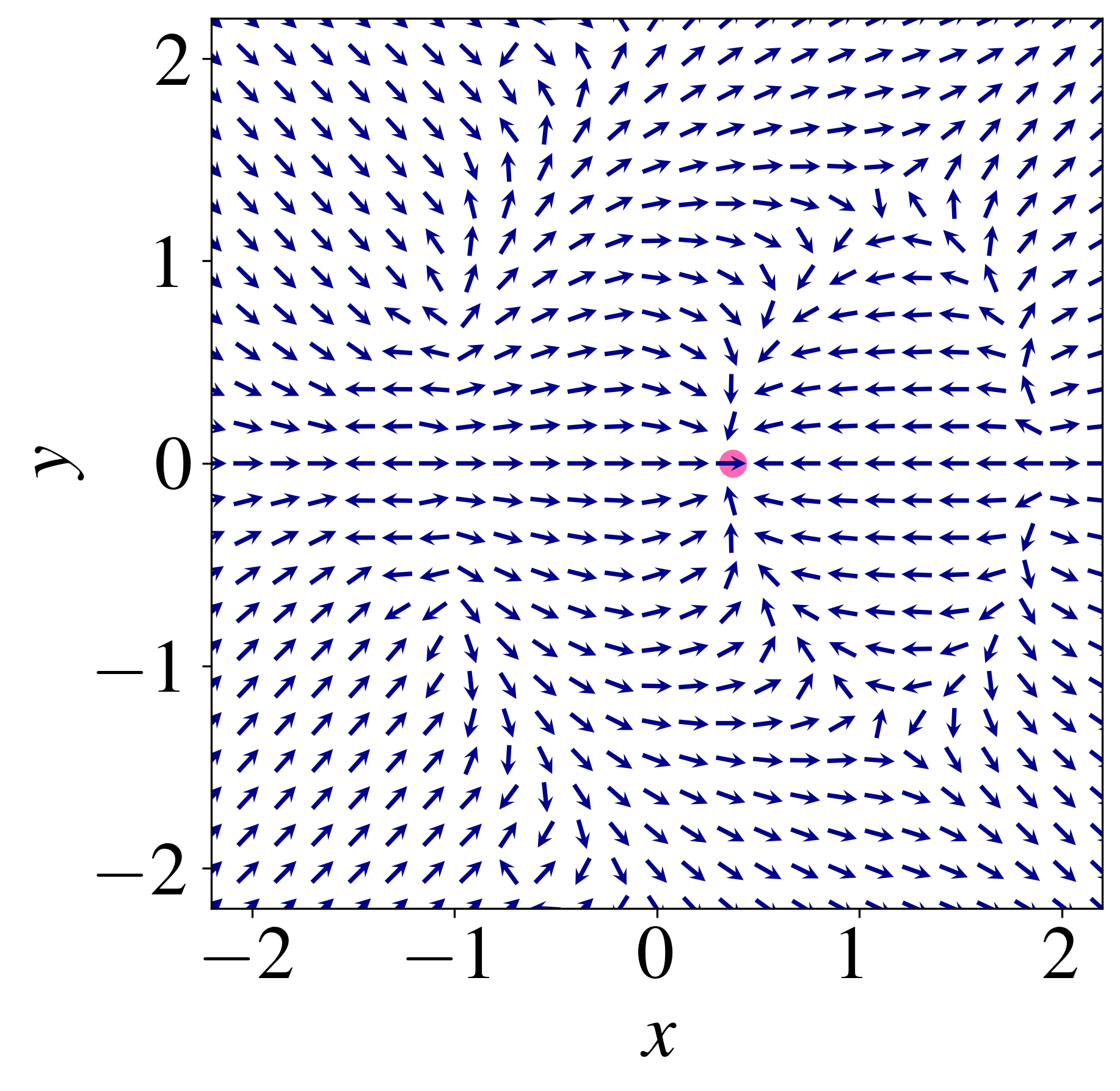}
        \caption{$\rho=1.5$}
    \end{subfigure}

    \vspace{1.0em}

    \begin{subfigure}[t]{0.24\textwidth}
        \centering
        \includegraphics[width=\linewidth]{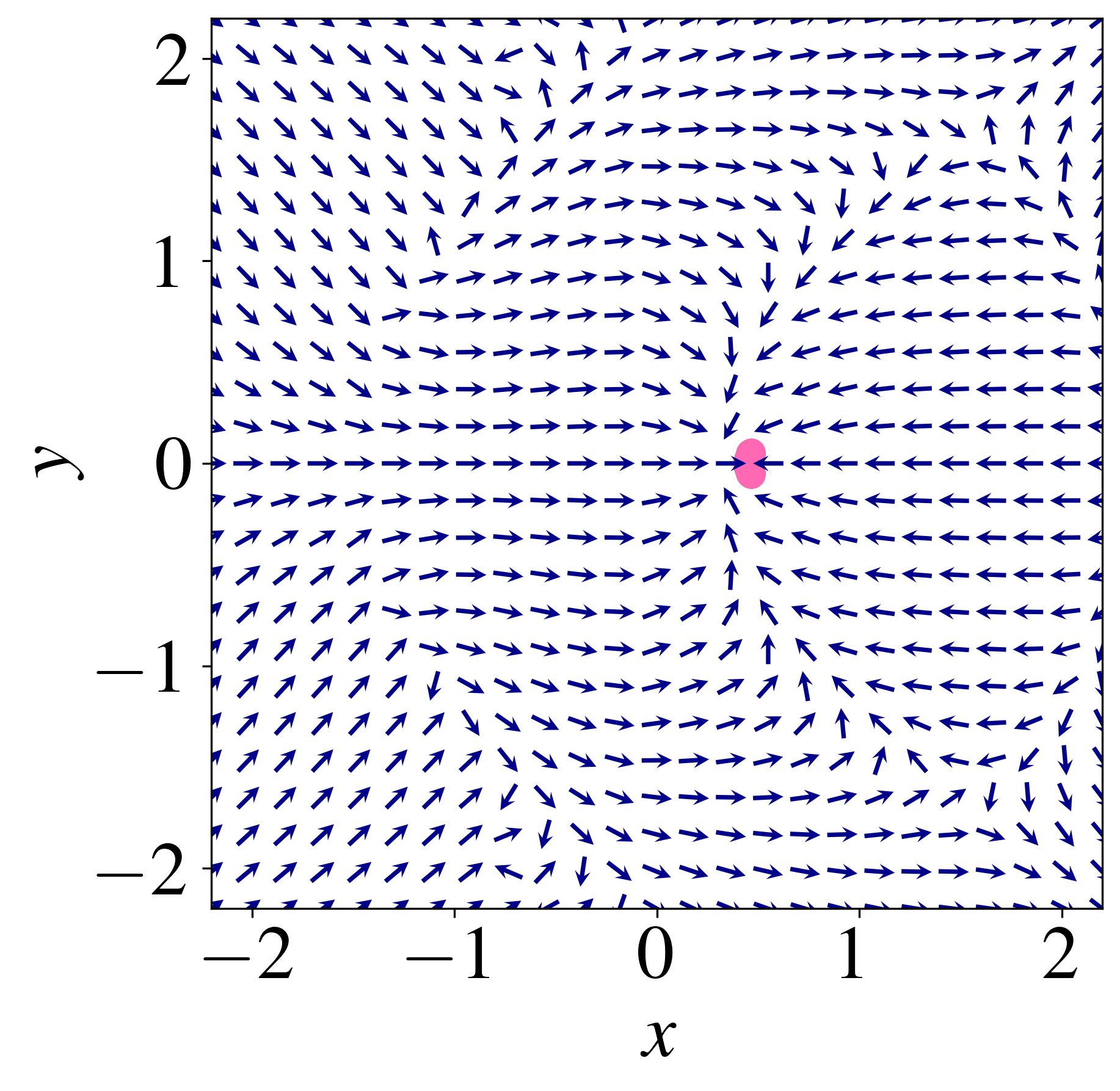}
        \caption{$\rho=2.0$}
    \end{subfigure}
    \begin{subfigure}[t]{0.24\textwidth}
        \centering
        \includegraphics[width=\linewidth]{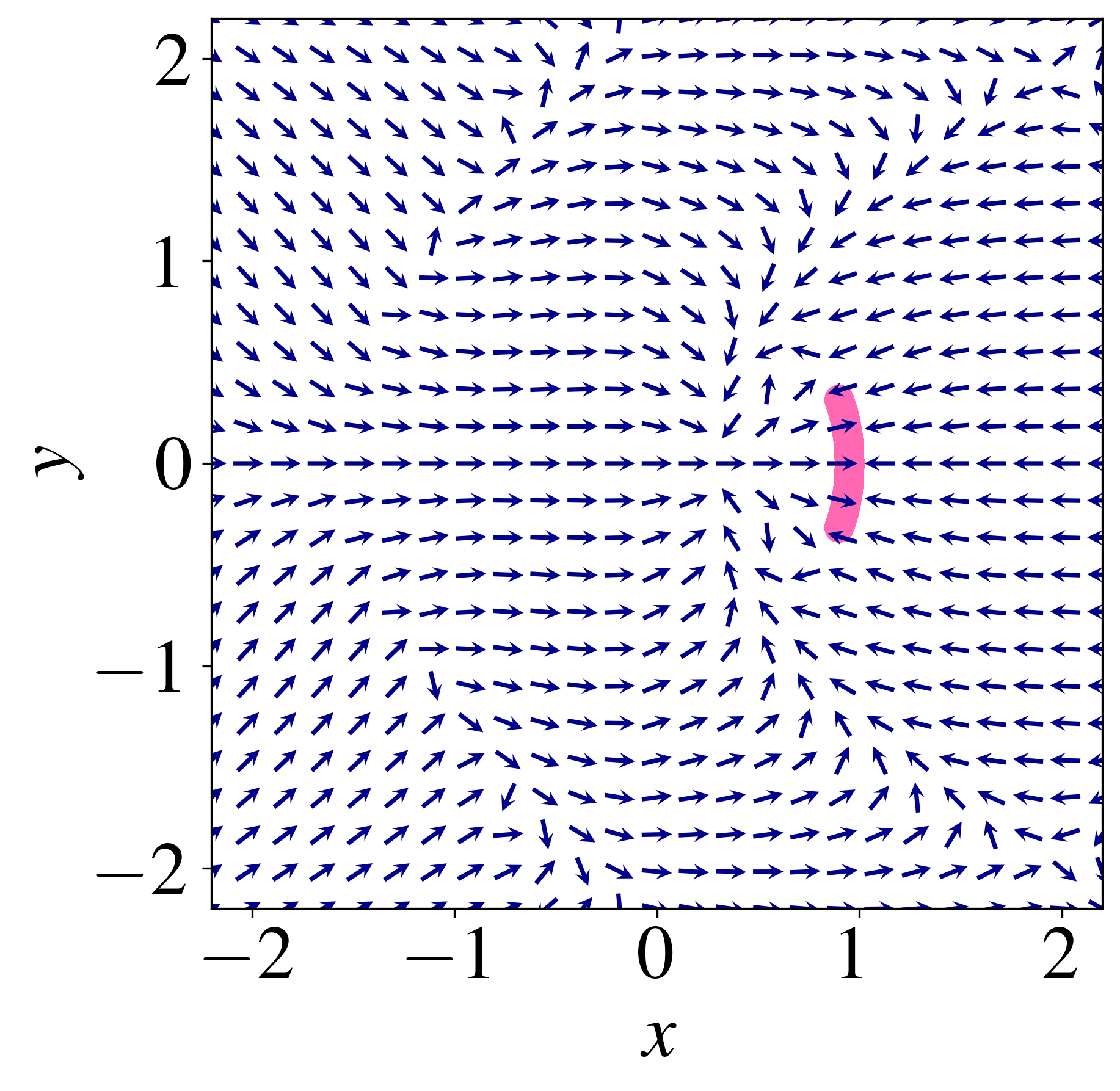}
        \caption{$\rho=2.5$}
    \end{subfigure}
    \begin{subfigure}[t]{0.24\textwidth}
        \centering
        \includegraphics[width=\linewidth]{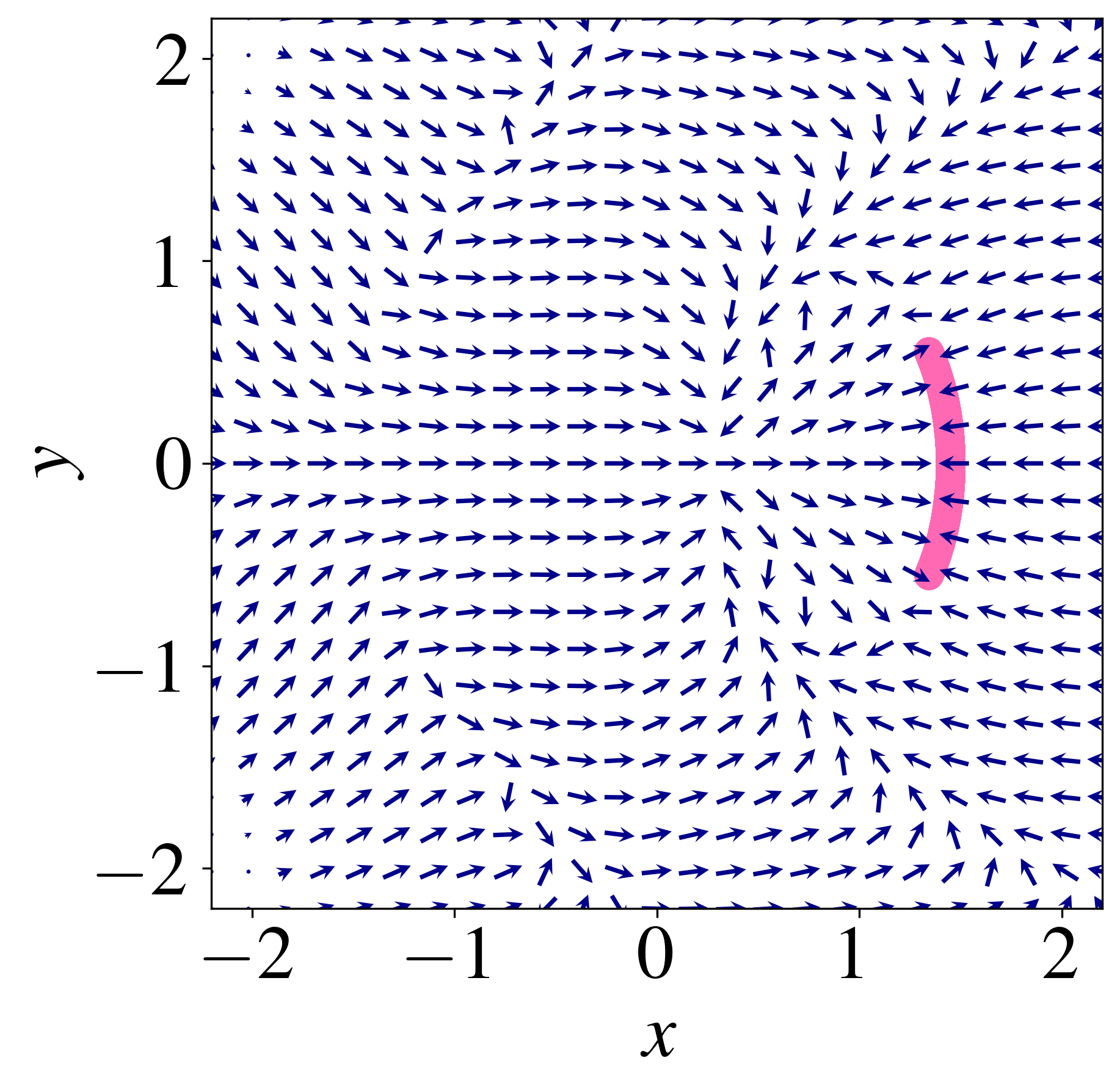}
        \caption{$\rho=3.0$}
    \end{subfigure}
    \begin{subfigure}[t]{0.24\textwidth}
        \centering
        \includegraphics[width=\linewidth]{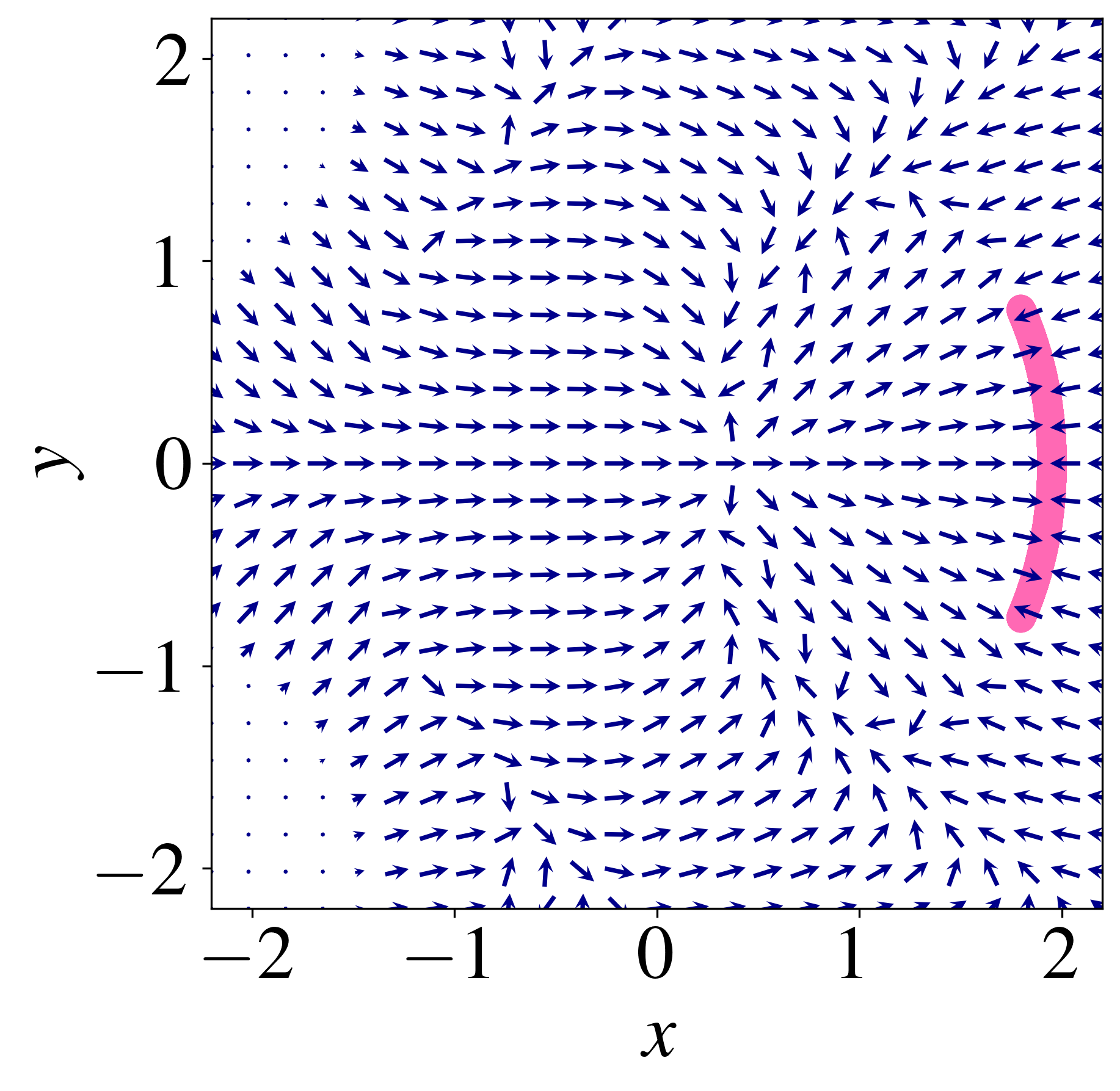}
        \caption{$\rho=3.5$}
    \end{subfigure}

    \caption{SAM gradient field $\nabla f(x+\rho\frac{\nabla f(x)}{\|\nabla f(x)\|})$ for different perturbation radii $\rho$. The corresponding minimizers of $f^{\mathrm{SAM}}$ are shown in pink.}

    \label{fig:rho_grad_fxplus}
\end{figure}

\section{Two-dimensional synthetic function for visualization}\label{app:exp1}

To visualize how the SAM perturbation radius $\rho$ affects the SAM loss $f^{\mathrm{SAM}}$,
we introduce the following two-dimensional synthetic function (originally illustrated in Figure~\ref{fig:toy}):
\begin{equation*}
f(x, y)
= 0.8\exp\left( -\frac{x^2 + y^2}{(2.5)^2} \right) \cdot W_X(x) 
 - \exp\left( -(x + 1.55 \cos(y / 1.5))^2 \right) \cdot W_Y(y) + 1,
\end{equation*}
where
\[
W_X(x) =
\begin{cases}
0, & x \leq -1, \\
0.5 \left( 1 - \cos(\pi(x + 1)) \right), & -1 < x < 0, \\
1, & x \geq 0,
\end{cases}
\]
\[
W_Y(y) =
\begin{cases}
1, & |y| \leq 0.6, \\
0.5 \left( 1 + \cos\left( \pi \cdot \frac{|y| - 0.6}{5.0} \right) \right), & 0.6 < |y| < 5.6, \\
0, & |y| \geq 5.6.
\end{cases}
\]
The function \( f(x, y) \) is continuously differentiable, since all its components are smoothly joined.  
It is designed so that its minimizer set forms a curve.
In fact, the global minimizer set of \( f \) is exactly
\[
\left\{ (x, y) \in \mathbb{R}^2 \;\middle|\; x = -1.55 \cos\left(\tfrac{y}{1.5}\right),\; |y| \leq 0.6 \right\}.
\]

Figure~\ref{fig:toy} shows the original function $f$, the SAM loss $f^{\mathrm{SAM}}$, and the SAM gradient $\nabla f(x+\rho\frac{\nabla f(x)}{\|\nabla f(x)\|})$ at perturbation radius $\rho = 2.8$.
To further examine the effect of the perturbation radius, Figures~\ref{fig:rho_contour} and~\ref{fig:rho_grad_fxplus} illustrate how the SAM loss $f^{\mathrm{SAM}}$ and the SAM gradient $\nabla f(x+\rho\frac{\nabla f(x)}{\|\nabla f(x)\|})$ evolve as $\rho$ varies over the range $0, 0.5, \ldots, 3.5$.  
In this setting, the minimizers of $f^{\mathrm{SAM}}$ are defined as the regions where $f^{\mathrm{SAM}}$ attains its minimum values; in practice, they appear either as isolated points or as continuous curve-like structures.  

An analysis of the minimizers of $f^{\mathrm{SAM}}$ as a function of the perturbation radius $\rho$ reveals two distinct regimes.
For small $\rho$, the minimizers approach the critical points of $f$.
Although $f^{\mathrm{SAM}}$ is not defined at a critical point, higher-resolution numerical experiments show convergence arbitrarily close to such points.
At $\rho = 0$, $f^{\mathrm{SAM}}$ coincides with $f$, and the minimizers exactly match those of $f$.
For $\rho = 0.5$, the minimizers remain on the original minimizer set, whereas for $\rho = 1.0$ and $\rho = 1.5$, they shift toward the maximizers of $f$.
The corresponding gradient fields indicate that these critical points act as attractors of the SAM dynamics, consistent with the theoretical analysis of~\citet{compagnoni2023sde}.

In contrast, for larger perturbation radii, the minimizers of $f^{\mathrm{SAM}}$ form a curve on the right-hand side, starting near the maximizer and drifting outward as $\rho$ increases.
This behavior is consistent with the proof of Theorem~\ref{main:thm1}, which shows that hallucinated minimizers emerge for a large $\rho$.
Theorem~\ref{thm:manifold} further establishes that these minimizers preserve the dimensionality of the original minimizer manifold.
Meanwhile, the SAM gradient field shows that these minimizers act as attractors within their neighborhood, and the conditions of Theorem~\ref{thm:attractor} are indeed satisfied.

Taken together,  this 2D example illustrates three qualitative regimes:
(i) for small $\rho$, minimizers of $f^{\mathrm{SAM}}$ track minimizers/critical regions of $f$ (with the caveat that $f^{\mathrm{SAM}}$ is undefined exactly at critical points);
(ii) as $\rho$ increases, the minimizers of $f^{\mathrm{SAM}}$ can drift toward high-loss regions near maximizers, consistent with the large-$\rho$ mechanism in Theorem~\ref{main:thm1};
and (iii) the resulting minimizers of $f^{\mathrm{SAM}}$ can form non-isolated, curve-like sets, consistent with the manifold mechanism in Theorem~\ref{thm:manifold}. 
Moreover, the SAM gradient field visually points toward these minimizer sets, suggesting local attraction in the sense of Theorem~\ref{thm:attractor} (the eigenvalue condition can be checked numerically in this low-dimensional setting).  

\clearpage
\bibliography{reference}
\bibliographystyle{plainnat}

\end{document}